%% file: ijcai23.tex
\documentclass{article}
\usepackage{ijcai23}

\input{math_commands.tex}

\usepackage{times}
\usepackage{soul}
\usepackage{url}
\usepackage[hidelinks]{hyperref}
\usepackage[utf8]{inputenc}
\usepackage[small]{caption}
\usepackage{graphicx}
\usepackage{amsmath}
\usepackage{amsthm}
\usepackage{booktabs}
\usepackage{algorithm}
\usepackage{algorithmic}
\usepackage[switch]{lineno}
\usepackage{multirow}
\usepackage{graphicx}
\usepackage{subfig}
\usepackage{xcolor}
\usepackage[scientific-notation=true]{siunitx}

\author{
Sagar Srinivas Sakhinana\thanks{Conceived, designed, implemented the research(programmed the software) and drafted the manuscript}, \{Krishna Sai Sudhir Aripirala, Shivam Gupta\}\thanks{Performed computational experiments, interpretation and visualization analysis of the results}, Venkataramana Runkana
\affiliations
TCS Research\\
\emails
\{sagar.sakhinana, k.aripirala, g.shivam4, venkat.runkana\}@tcs.com,
}

\begin{document}
\thispagestyle{myheadings}
\markboth{}{Knowledge-Based Compositional Generalization Workshop, International Joint Conferences on Artificial Intelligence(IJCAI-23).} \nonumber

\title{Multi-Source Knowledge-Based Hybrid Neural Framework for Time Series Representation Learning}


\maketitle

\vspace{-1mm}
\section{ABSTRACT} 
\vspace{-1mm}
Accurately predicting the behavior of complex dynamical systems, characterized by high-dimensional multivariate time series(MTS) in interconnected sensor networks, is crucial for informed decision-making in various applications to minimize risk. While graph forecasting networks(GFNs) are ideal for forecasting MTS data that exhibit spatio-temporal dependencies, prior works rely solely on the domain-specific knowledge of time-series variables inter-relationships to model the nonlinear dynamics, neglecting inherent relational-structural dependencies among the variables within the MTS data. In contrast, contemporary works infer relational structures from MTS data but neglect domain-specific knowledge. The proposed hybrid architecture addresses these limitations by combining both domain-specific knowledge and implicit knowledge of the relational structure underlying the MTS data using Knowledge-Based Compositional Generalization. The hybrid architecture shows promising results on multiple benchmark datasets, outperforming state-of-the-art forecasting methods. Additionally, the architecture models the time-varying uncertainty of multi-horizon forecasts.

\vspace{-3mm}
\section{INTRODUCTION} 
\vspace{-1mm}
Multivariate time series forecasting(MTSF) is a crucial task with diverse applications in various sectors, including finance, healthcare, energy, and others. MTSF facilitates strategic decision-making by predicting interrelated variables that change over time. In retail and e-commerce, it is used to forecast product demand, optimize supply chains, and manage inventory levels. Cloud providers utilize MTSF to accurately predict web traffic and efficiently scale server fleets to meet anticipated demand. Forecasting MTS data is challenging due to the complex interrelationships among multiple variables and the unique characteristics of MTS data, such as non-linearity, high-dimensionality and non-stationarity. Over the past few years, Spatial-temporal graph neural networks(STGNNs) have gained popularity for modeling intricate dependencies in MTS data, improving forecast accuracy. While explicit relationships among variables are provided by human experts through a predefined or explicit graph, implicit relationships are obtained using data-driven neural relational inference methods(\cite{kipf2018neural}). Implicit relationships, characterized by their high complexity and non-linearity, evolve over time and reveal hidden interrelations among variables unknown to human experts which are not trivial, whereas explicit relations stemming from domain expertise remain static. Existing ``human-in-the-loop" STGNNs(\cite{yu2017spatio}, \cite{li2017diffusion}, \cite{guo2020optimized}) integrate domain-specific knowledge and learn MTS data dynamics, but real-world situations often present unknown or incomplete graph structures, resulting in suboptimal forecasting. These predefined structures may also inadequately capture non-static spatio-temporal dependencies, impeding the accurate inference of latent time-conditioned relations that influence variable co-movements within the MTS data. Moreover, STGNNs neglect the significance of edges in explicit graph structure, hindering modeling of complex systems. Incorporating effective methods to represent edge information within STGNNs is necessary for accurately modeling complex dynamical systems. Additionally, STGNNs have limitations in capturing the importance of subgraphs within the larger explicit graph structure, which can be addressed by developing new methods to incorporate subgraph dynamics to model complex systems. Conversely, recent ``human-out-of-the-loop" STGNNs(\cite{shang2021discrete}, \cite{deng2021graph}, \cite{wu2020connecting}) simultaneously learn the discrete dependency graph structures and the MTS data dynamics, but neglect predefined inter-relationships from domain expertise, leading to subpar performance in graph time-series forecasting. Moreover, implicitly learning the latent graph structure from MTS data is constrained by the limitations of pairwise connections among the variables. However, the interconnected networks in complex dynamical systems could have higher-order structural relations beyond pairwise associations. Hypergraphs, a generalization of graphs, can effectively model these relations in high-dimensional MTS data. Additionally, while conventional STGNNs emphasize pointwise forecasting, they do not offer any estimates of uncertainty for the multi-horizon forecasts. We propose the Multi-Source Knowledge-Based Hybrid Neural Framework(for brevity, \textbf{MKH-Net}) to address these challenges. This framework integrates the domain-specific knowledge and data-driven knowledge using a joint-learning approach to model the complex spatio-temporal dynamics underlying the MTS data, resulting in better forecast accuracy and reliable uncertainty estimates. The proposed framework has two main components: spatial and temporal inference components. Using a space-then-time(STT, \cite{gao2022equivalence}) approach, the framework performs the spatial message-passing schemes prior to the temporal-encoding step. The spatial inference component combines ``implicit hypergraph", ``explicit subgraph", and ``dual-hypergraph" representation learning methods to learn the various aspects of the underlying structure of interrelationships among variables in MTS data, characterizing the complex sensor network-based dynamical systems. The ``implicit hypergraph" method learns the hierarchical interdependencies between variables in MTS data by modeling the discrete hypergraph relational structure and performs the hypergraph representation learning schemes to obtain latent hypernode-level representations, which accurately capture the spatio-temporal dynamics of the hypergraph-structured MTS data. The ``explicit subgraph" method extracts overlapping subgraph patches and uses subgraph message-passing schemes to learn spatio-temporal dynamics within the explicit graph-structured MTS data. The `dual-hypergraph" method captures the latent information of edges in explicit graph-structured MTS data by utilizing the Dual Hypergraph Transformation(DHT) method. This is achieved through a powerful message-passing scheme that is tailored specifically to the edges in the structured MTS data, resulting in a more accurate modeling of the underlying spatio-temporal dynamics. The proposed \textbf{MKH-Net} framework uses a gating mechanism to perform a convex combination of the multi-knowledge representations computed by the different methods, resulting in more accurate latent representations of the complex non-linear dynamics in MTS data. The \textbf{MKH-Net} framework is capable of capturing various types of dependencies that exist across different observation scales, as correlations among variables may vary in short-term versus long-term views of the MTS data. The temporal learning component models the time-evolving dynamics of interdependencies among variables in MTS data, enabling the framework to provide accurate multi-horizon forecasts and precise predictive uncertainty estimates. To put it briefly, the proposed framework offers an end-to-end methodology for learning spatio-temporal dynamics in MTS data with both explicit graph and implicit hypergraph structures. This approach utilizes multiple representation learning methods, including ``explicit subgraph", ``implicit hypergraph", and ``dual-hypergraph", to capture evolutionary and multi-scale interactions among variables in the latent representations to achieve better modeling accuracy. The framework also models time-varying uncertainty in forecasts and utilizes the learned latent representations for downstream MTSF tasks, resulting in accurate multi-horizon forecasts and reliable predictive uncertainty estimates. Additionally, the framework is designed to offer better generalization and scalability for large-scale spatio-temporal MTS forecasting tasks found in real-world applications.

\vspace{-3mm}  
\section{PROBLEM DEFINITION}
\vspace{-1mm}
Let us consider a  historical time series dataset with $n$ correlated variables, observed over $\mathrm{T}$ time steps, represented by the notation \thickmuskip=0.15\thickmuskip\resizebox{.135\textwidth}{!}{$\mathbf{X} = \big(\mathbf{x}_{1}, \ldots, \mathbf{x}_{\mathrm{T}}  \big)$}. Here, the subscript indicates the time step, while the observations for all the $n$ variables at time step $t$ are denoted by \thickmuskip=0.15\thickmuskip\resizebox{.255\textwidth}{!}{$\mathbf{x}_{t} = \big(\mathbf{x}_t^{(1)}, \mathbf{x}_t^{(2)}, \ldots, \mathbf{x}_t^{(n)}\big) \in \mathbb{R}^{(n)}$}, where the superscript refers to the variables. In the context of MTSF, we employ the rolling-window method for multi-horizon forecasting, where a look-back window is predefined at the current time step $t$ to include the prior $\tau$-steps of MTS data, to predict the next $\upsilon$-steps. Specifically, we aim to utilize a historical window of $n$-correlated variables, represented by the notation \resizebox{.14\textwidth}{!}{$\mathbf{X}_{(t - \tau : \hspace{1mm}t-1)} \in \mathbb{R}^{n \times \tau}$}, which have been observed over the previous $\tau$-steps prior to the current time step $t$, to make predictions about the future values of $n$ variables for the next $\upsilon$-steps, denoted as \resizebox{.145\textwidth}{!}{$\mathbf{X}_{(t  : t + \upsilon - 1)} \in \mathbb{R}^{n \times \upsilon}$}.  \textcolor{black}{To capture the spatio-temporal correlations among a multitude of correlated time series variables, the MTSF problem is formulated on graph and hypergraph structures}. The historical inputs are represented as continuous-time spatial-temporal graphs denoted by \resizebox{.225\textwidth}{!}{$\mathcal{G}_{t} = \big(\mathcal{V}, \mathcal{E}, \mathbf{X}_{(t - \tau : \hspace{1mm}t-1)}, \text{A}^{(0)}\big)$}, where \resizebox{.02\textwidth}{!}{$\mathcal{G}_{t}$} is composed of a set of nodes(\resizebox{.013\textwidth}{!}{$\mathcal{V}$}) which represent the variables, edges(\resizebox{.013\textwidth}{!}{$\mathcal{E}$}) that describe the connections among the variables, and a node feature matrix \resizebox{.09\textwidth}{!}{$\mathbf{X}_{(t - \tau : \hspace{1mm}t-1)}$} that changes over time. The explicit static-graph structure based on prior knowledge of time-series variables relationships is described by the adjacency matrix  \resizebox{.15\textwidth}{!}{$\text{A}^{(0)} \in \{0,1\}^{|\mathcal{V}| \times |\mathcal{V}|}$}. To further capture the complex relationships among MTS data, we consider the historical inputs as a sequence of dynamic hypergraphs, denoted by \resizebox{.25\textwidth}{!}{$\mathcal{HG}_{t} = \big(\mathcal{HV}, \mathcal{HE}, \mathbf{X}_{(t - \tau : \hspace{1mm}t-1)}, \text{I}\big)$}. Here, the hypergraph is represented by a fixed set of hypernodes(\resizebox{.03\textwidth}{!}{$\mathcal{HV}$}) and hyperedges(\resizebox{.03\textwidth}{!}{$\mathcal{HE}$}), where the hypernodes denote the variables, and the hyperedges capture the latent higher-order relationships between the hypernodes. The time-varying hypernode feature matrix is given by \resizebox{.09\textwidth}{!}{$\mathbf{X}_{(t - \tau : \hspace{1mm}t-1)}$}. The implicit hypergraph structure is learned through an embedding-based similarity metric learning approach. The incidence matrix \resizebox{.075\textwidth}{!}{$\mathbf{I} \in \mathbb{R}^{n \times m}$} describes the hypergraph structure, where $\mathbf{I}_{p, \hspace{0.5mm}q}=1$ if the hyperedge $q$ is incident with hypernode $p$, and otherwise 0. The sparsity of the hypergraph is determined by the number of hyperedges($\text{m}$) in the hypergraph. The proposed framework aims to learn a function \resizebox{.04\textwidth}{!}{$F(\theta)$} that can map MTS data, \resizebox{.09\textwidth}{!}{$\mathbf{X}_{(t - \tau : \hspace{1mm}t-1)}$}, to their respective future values, \resizebox{.09\textwidth}{!}{$\mathbf{X}_{(t : t + \upsilon - 1)}$}, given a \resizebox{.0205\textwidth}{!}{$\mathcal{G}_{t}$} and \resizebox{.0365\textwidth}{!}{$\mathcal{HG}_{t}$}. 

\vspace{-3mm}
\resizebox{0.945\linewidth}{!}{
\begin{minipage}{\linewidth}
\begin{align}
\left[\mathbf{x}_{(t - \tau)}, \cdots, \mathbf{x}_{(t-1)} ; \mathcal{G}_{t}, \mathcal{HG}_{t}\right] \stackrel{F(\theta)}{\longrightarrow}\left[\mathbf{x}_{(t + 1)}, \cdots, \mathbf{x}_{(t + \upsilon-1)}\right] \nonumber
\end{align} 
\end{minipage} 
}

\vspace{1mm}
The MTSF task formulated on the explicit graph (\resizebox{.0205\textwidth}{!}{$\mathcal{G}_{t}$}) and implicit hypergraph (\resizebox{.035\textwidth}{!}{$\mathcal{HG}_{t}$}) can be expressed as follows:

\vspace{-2mm}
\resizebox{0.945\linewidth}{!}{
\begin{minipage}{\linewidth}
\begin{align}
\min _{\theta} \mathcal{L}\big(\mathbf{X}_{(t : t + \upsilon-1)}, \hat{\mathbf{X}}_{(t : t + \upsilon-1)} ; \mathbf{X}_{(t - \tau : \hspace{1mm}t-1)}, \mathcal{G}_{t}, \mathcal{HG}_{t}\big) \nonumber
\end{align} 
\end{minipage}
}

\vspace{0mm}
Here, $\theta$ represents all the learnable parameters of the trainable function \resizebox{.04\textwidth}{!}{$F(\theta)$}. The model predictions are denoted by \resizebox{.085\textwidth}{!}{$\hat{\mathbf{X}}_{(t : t + \upsilon-1)}$}, and $\mathcal{L}$ represents the loss function. We train our learning algorithm using the mean absolute error(MAE) function, which is defined as follows:

\vspace{-2mm}
\resizebox{0.925\linewidth}{!}{
\begin{minipage}{\linewidth}
\begin{align}
\mathcal{L}_{\text{MAE}}\left(\theta\right)=\frac{1}{\upsilon}\left|\mathbf{X}_{(t : t + \upsilon-1)}-\hat{\mathbf{X}}_{(t : t + \upsilon-1)}\right| \nonumber
\end{align}
\end{minipage}
}

\vspace{-3mm}
\section{OUR APPROACH}
\begin{figure}[!ht]
\resizebox{1.0\linewidth}{!}{ 
\hspace*{-4mm}\includegraphics[keepaspectratio,height=4.5cm,trim=0.0cm 4.5cm 0cm 6.0cm,clip]{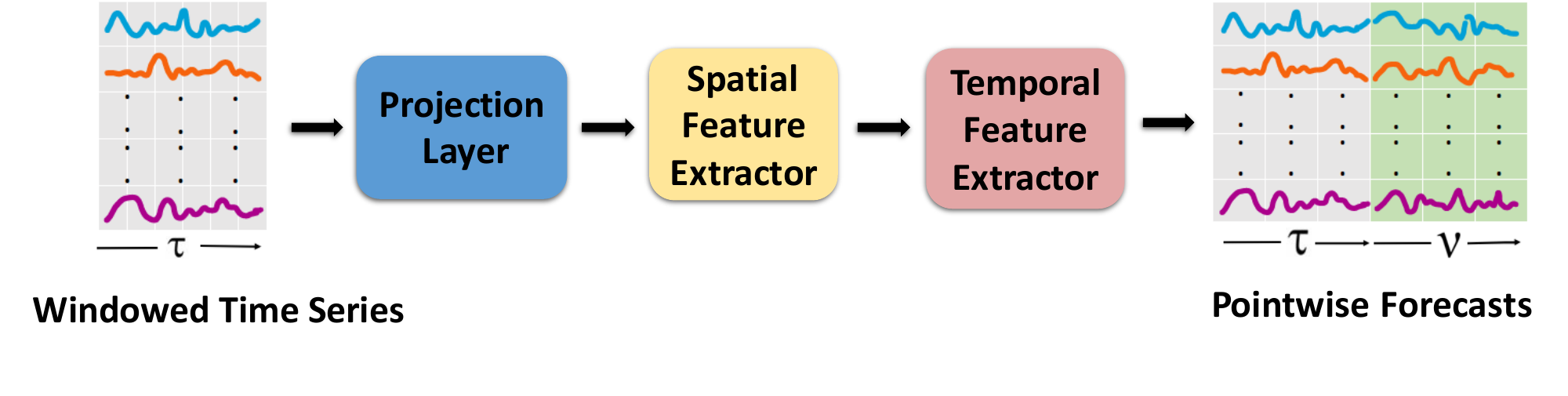} 
}
\vspace{-10mm}
\caption{Overview of \textbf{MKH-Net} framework.}
\label{fig:figure1}
\end{figure}

\vspace{-1mm}
Our framework presents a neural forecasting architecture composed of three main components: the projection layer, spatial inference, and temporal inference components, which are illustrated in Figure \ref{fig:figure1}. The spatial inference component includes three methods: ``implicit hypergraph", ``explicit subgraph", and ``dual-hypergraph" representation learning methods. The ``implicit hypergraph" method computes the dependency hypergraph structure of multiple time series variables and uses higher-order message-passing schemes to model the hypergraph-structured MTS data. This approach computes time-conditioned, optimal hypernode-level representations, capturing complex relationships between variables over time. The ``explicit subgraph" method consists of two modules, the patch extraction and the subgraph encoder. The patch extraction module extracts overlapping subgraph patches from a predefined graph, while the subgraph encoder module uses spatial graph-filtering techniques to compute time-evolving, optimal node-level representations, which effectively capture the underlying spatio-temporal dynamics of the graph-structured MTS data. The ``dual-hypergraph" method transforms edges(nodes) in an explicit graph into hypernodes(hyperedges) in a dual hypergraph, allowing hypernode-level message-passing schemes to be applied for edge representation learning of the explicit graphs. The temporal inference component of the framework combines multiple latent representations from different methods and learns their temporal dynamics. By jointly optimizing the different learning components, the framework provides accurate multi-horizon forecasts and reliable uncertainty estimates for various time-series forecasting tasks. 

\vspace{-3mm}

\vspace{-5mm}
\subsection{PROJECTION LAYER}
\vspace{-1mm}
The proposed framework includes a projection layer that utilizes gated linear networks(GLUs, \cite{dauphin2017language}) to learn non-linear representations of the input data. The input data is represented by \resizebox{.15\textwidth}{!}{$\mathbf{X}_{(t - \tau : \hspace{1mm}t-1)} \in \mathbb{R}^{n \times \tau}$}, and the projection layer uses GLUs to selectively pass information through a gating mechanism and transform the input data into a new feature matrix, \resizebox{.15\textwidth}{!}{$\bar{\mathbf{X}}_{(t  : t + \upsilon - 1)} \in \mathbb{R}^{n \times d}$}, computed as follows,

\vspace{-3mm}
\resizebox{0.935\linewidth}{!}{
\begin{minipage}{\linewidth}
\begin{align}
\centering
\bar{\mathbf{X}}_{(t  : t + \upsilon - 1)}   = \big( \sigma(\text{W}_{0}\mathbf{X}_{(t - \tau : \hspace{1mm}t-1)}) \otimes \text{W}_{1}\mathbf{X}_{(t - \tau : \hspace{1mm}t-1)}\big)\text{W}_{2} \nonumber
\end{align}
\end{minipage}
}

\vspace{1mm}
where $\text{W}_{0}, \text{W}_{1}, \text{W}_{2} \in \mathbb{R}^{\tau \times d}$ denotes the trainable weight matrices, and an element-wise multiplication operation denoted by $\otimes$. The non-linear activation function, $\sigma$, is applied to enhance the representation learning process.

\vspace{-1mm}
\subsection{SPATIAL-INFERENCE}
\vspace{0mm}
Figure \ref{fig:figure5} shows the spatial inference component of the framework, which comprises of three distinct methods. Further details are discussed in the following sections.

\vspace{-1mm}
\subsubsection{Hypergraph inference and representation learning}
\vspace{1mm}
The ``implicit hypergraph" method is composed of two modules: hypergraph inference(HgI) and hypergraph representation learning(HgRL). HgI module uses a similarity metric learning method to capture the hierarchical interdependence relations among time-series variables and compute a discrete hypergraph topology for a hypergraph-structured representation of MTS data. The differentiable embeddings, $\mathbf{z_{i}}, \mathbf{z_{j}} \in \mathbb{R}^{(d)}$, where $1 \leq i \leq n$ and $1 \leq j \leq m$, represent the hypernodes and hyperedges of the hypergraph, respectively, and capture their global-contextual behavioral patterns in a $d$-dimensional vector space. These embeddings enable the HgI module to effectively model the dynamic relationships among the variables over time, making it a powerful tool for learning task-relevant relational hypergraph structures from complex MTS data. We compute the pairwise similarity between any pair $\mathbf{z_{i}}$ and $\mathbf{z_{j}}$ as follows,

\vspace{-3mm}
\resizebox{0.90\linewidth}{!}{
\begin{minipage}{\linewidth}
\begin{align}
\text{P}_{i,j} = \sigma \big([\text{S}_{i,j} || 1- \text{S}_{i,j}]\big); \text{S}_{i,j} = \frac{\mathbf{z^{T}_{i}} \mathbf{z_{j}} + 1}{2\left\|\mathbf{z_{i}}\right\| \cdot\left\|\mathbf{z_{j}}\right\|}  \nonumber
\end{align}
\end{minipage}
}

\vspace{0mm}
where \resizebox{.00985\textwidth}{!}{$\Vert$} denotes vector concatenation. The sigmoid activation function maps the pairwise scores to the range [0,1]. The hyperedge probability over hypernodes of the hypergraph is denoted by \resizebox{.1\textwidth}{!}{$\text{P}^{(k)}_{i,j} \in \mathbb{R}^{nm \times 2}$}, where $k \in \{0,1\}$. The scalar value of \resizebox{.08\textwidth}{!}{$\text{P}^{(k)}_{i,j} \in [0,1]$} encodes the relation between an arbitrary pair of hypernodes and hyperedges $(i,j)$. \resizebox{.03\textwidth}{!}{$\text{P}^{(0)}_{i,j}$} represents the probability of a hypernode $i$ connected to hyperedge $j$, while \resizebox{.03\textwidth}{!}{$\text{P}^{(1)}_{i,j}$} represents the probability that the hypernode $i$ is not connected to the hyperedge $j$. We utilize the Gumbel-softmax trick, as presented in \cite{jang2016categorical}, which allows for accurate and efficient sampling of discrete hypergraph structures from the hyperedge probability distribution $\text{P}_{i,j}$. This technique enables the HgI module to effectively capture intricate relationships among variables in MTS data. The connectivity pattern of the sampled hypergraph structure is represented by an incidence matrix $\mathbf{I} \in \mathbb{R}^{n \times m}$, which encapsulates the relationships between hypernodes and hyperedges in the hypergraph. The Gumbel-softmax trick enables learning of the hypergraph structure in an end-to-end differentiable manner, facilitating the application of gradient-based optimization methods during model training within an inductive-learning approach. The incidence matrix is computed as,

\vspace{-2mm}
\resizebox{0.95\linewidth}{!}{
\begin{minipage}{\linewidth}
\begin{align}
\mathbf{I}_{i,j} =\exp \big(\big(g^{(k)}_{i,j} + \text{P}^{(k)}_{i,j}\big) / \gamma\big)\big/{\sum \exp \big(\big(g^{(k)}_{i,j} + \text{P}^{(k)}_{i,j}\big) / \gamma\big)} \nonumber
\end{align}
\end{minipage}
}

\vspace{1mm}
Where, the temperature parameter($\gamma$) of the Gumbel-Softmax distribution is set to 0.05. The random noise sampled from the Gumbel distribution is denoted by \resizebox{.315\textwidth}{!}{$g^{(k)}_{i j} \sim \operatorname{Gumbel}(0,1) = \log (-\log (\text{U}(0,1))$}, where $\text{U}$ denotes the uniform distribution with a range of 0 to 1.
The learned hypergraph is then regularized to be sparse by optimizing the probabilistic hypergraph distribution parameters, which drops the redundant hyperedges over hypernodes. The forecasting task provides indirect supervisory information, which helps to reveal the higher-order structure or hypergraph relation structure in the observed MTS data. We utilize a sequence of dynamic hypergraphs to represent the MTS data, where each hypergraph is denoted by \resizebox{.24\textwidth}{!}{$\mathcal{HG}_{t} = \big(\mathcal{HV}, \mathcal{HE}, \mathbf{X}_{(t - \tau : \hspace{1mm}t-1)}, \text{I}\big)$}. A hypergraph representation learning(HgRL) module is employed to compute optimal hypernode-level representations that capture the spatio-temporal dynamics within the hypergraph-structured MTS data. These representations are then used for performing inference on the downstream multi-horizon forecasting task. The HgRL module is a neural network architecture that utilizes both Hypergraph Attention Network(HgAT) and Hypergraph Transformer(HgT) to accomplish this task. HgT employs multi-head self-attention mechanisms to learn latent hypergraph representations, $\mathbf{h^{\prime}_{i}}^{(t)}$, without prior knowledge about the hypergraph structure. On the other hand, HgAT performs higher-order message-passing schemes on the hypergraph topology to compute the latent hypernode representations, $\mathbf{h}^{(t)}_{i}$. The combination of HgT and HgAT provides HgRL with a powerful backbone for capturing complex relationships and dependencies among variables within the hypergraph-structured MTS data in the differentiable latent hypergraph representations. Further implementation details and an in-depth explanation are available in the appendix. A gating mechanism  is implemented to regulate the information flow from  $\mathbf{h^{\prime}_{i}}^{(t)}$ and $\mathbf{h}^{(t)}_{i}$, which produces a weighted combination of representations \resizebox{.05\textwidth}{!}{$\mathbf{h}^{(t)}_{i, \text{IMP}}$}. The gating mechanism is described by,

\vspace{-3mm}
\resizebox{0.925\linewidth}{!}{
\hspace{0mm}\begin{minipage}{\linewidth}
\begin{align}
 g^{\prime} &= \sigma \big( f^{\prime}_s(\mathbf{h^{\prime}_{i}}^{(t)}) + f^{\prime}_g(\mathbf{h}^{(t)}_{i}) \big) \nonumber \\
\mathbf{h^{(t)}_{i, IMP}}  &= \sigma \big( g^{\prime}(\mathbf{h^{\prime}_{i}}^{(t)})) + (1-g^{\prime})(\mathbf{h}^{(t)}_{i}) \big) \nonumber 
\end{align} 
\end{minipage}
} 

\vspace{1mm}
where $f^{\prime}_s$ and $f^{\prime}_g$ are linear projections. Fusing representations can be useful for modeling the multi-scale interactions underlying spatio-temporal hypergraph data and can help mitigate overfitting. By incorporating the most relevant information, the proposed framework captures time-evolving underlying patterns in MTS data, resulting in more accurate and robust forecasts. In brief, the hypergraph learning module optimizes the discrete hypergraph structure using a similarity metric learning technique and formulates the posterior forecasting task as message-passing schemes with hypergraph neural networks to learn optimal hypergraph representations, resulting in accurate and expressive representations of MTS data, thereby improving forecast accuracy.

\vspace{-4mm}
\begin{figure}[!ht]
\hspace*{-12mm}
\includegraphics[keepaspectratio,height=4.5cm, width=10cm,trim=0cm 6cm 4cm 7.5cm,clip]{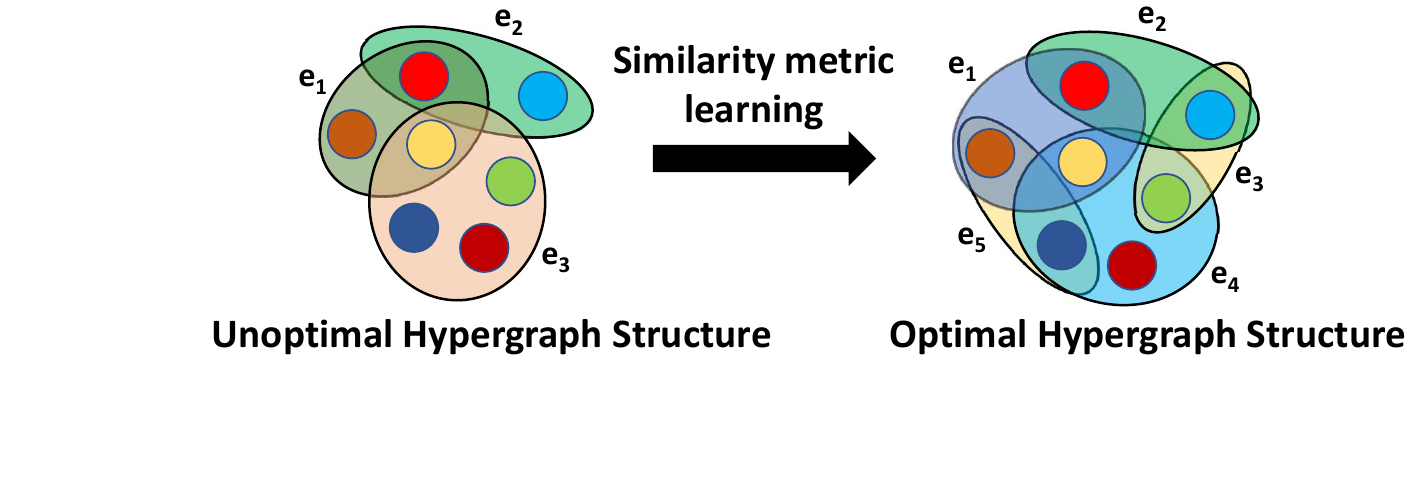} 
\vspace{-14mm}
\caption{The hypergraph inference(HgI) module learns complex hierarchical structural-dynamic dependencies among multiple time series variables in a sparse discrete hypergraph structure using similarity metric learning. In this structure, hypernodes represent variables, and hyperedges represent higher-order relations among an arbitrary number of hypernodes. The HgI module is fully differentiable, allowing for efficient optimization using gradient-based algorithms. Simultaneously, the hypergraph representation learning(HgRL) module encodes the structural spatio-temporal inductive biases for modeling the nonlinear dynamics of interconnected sensor networks. In short, the HgRL module learns a latent representations of the continuous-time spatio-temporal hypergraph structures to capture the underlying patterns and trends in the MTS data. For illustration purpose, in the above figure filled circles denote the hypernodes and filled eclipses($e$) denotes the hyperedges.}
\label{fig:figure2}
\end{figure}

\vspace{-5mm}
\subsubsection{Subgraph Representation Learning}
\begin{figure}[!ht]  
\hspace*{-1mm}
\includegraphics[keepaspectratio,height=5cm,trim=0cm 0.15cm 0cm 0.45cm,clip]{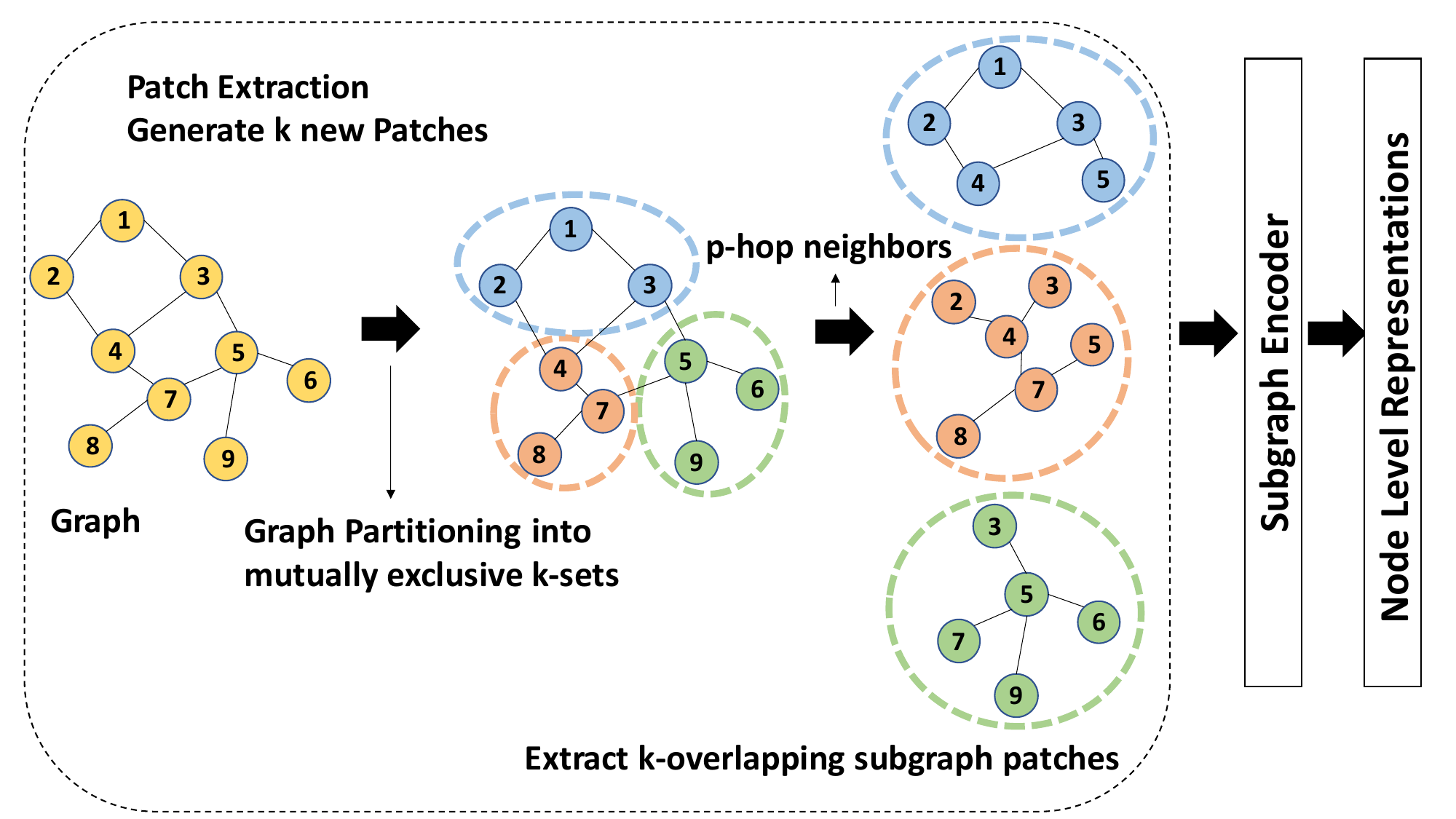} 
\vspace{-5mm}
\caption{The Subgraph representation learning(SgRL) method comprises of two modules: the patch extraction and the subgraph encoder. The patch extraction module partitions spatio-temporal graphs into mutually exclusive k-sets and expands them to include their p-hop neighborhood, creating k-overlapping subgraph patches. The subgraph encoder module processes these subgraph patches to compute node-level representations, which are then mean-pooled across all subgraph patches to obtain the final node representations. For illustration purpose, in the figure above, an arbitrary graph is partitioned into k(3)-mutually exclusive patches. Each patch is then expanded to include its p(1)-hop neighborhood, resulting in k(3)-overlapping subgraph patches.}
\label{fig:figure3}
\end{figure}

\vspace{1mm}
We represent the MTS data as continuous-time spatio-temporal graphs, utilizing domain-specific knowledge, where each graph is denoted by \resizebox{.215\textwidth}{!}{$\mathcal{G}_{t} = \big(\mathcal{V}, \mathcal{E}, \mathbf{X}_{(t - \tau : \hspace{1mm}t-1)}, \text{A}^{(0)}\big)$}. Subgraphs, which are substructures within a larger graph exhibit higher-order connectivity patterns, provide a powerful mechanism for capturing the complex interactions among time series variables in the graph-structured MTS data. As a result, subgraphs are more relevant for learning the complex spatio-temporal dependencies in the MTS data. By extracting higher-order connectivity patterns, subgraphs enable the learning of more accurate and interpretable latent representations, leading to improved performance on downstream MTSF task. The subgraph representation learning(SgRL) method involves two sequential modules: patch extraction and subgraph encoder, which collaboratively extract and encode subgraphs, resulting in expressive node-level representations that capture both local neighborhood and larger-scale structural information from the original explicit graph. Figure \ref{fig:figure3} depicts the SgRL method. The patch extraction module partitions the explicit graph at each time step into overlapping patches, also known as subgraph patches. Let's consider an explicit graph $\mathcal{G}_{t}=(\mathcal{V},\mathcal{E})$, an integer $k$ and a positive integer $p$. We aim to partition an explicit graph $\mathcal{G}_{t}$ with node set $\mathcal{V}$ and edge set $\mathcal{E}$ into $k$ subgraph patches, denoted by \resizebox{.15\textwidth}{!}{$\mathcal{G}^{(1)}_{t}, \mathcal{G}^{(2)}_{t}, \ldots, \mathcal{G}^{(k)}_{t}$}, where each subgraph patch consists of  $\frac{|\mathcal{V}|}{k}$ nodes in chronological order and their $p$-hop neighbors in the original graph $\mathcal{G}_{t}$. The task involves selecting the nodes and edges of each subgraph to ensure that they are mutually exclusive and form a connected subgraph with their $p$-hop neighbors. In short, to achieve the partition of explicit graph $\mathcal{G}_t=(\mathcal{V}, \mathcal{E})$ into $k$ subgraphs with $p$-hop neighbors, we can perform the following steps:

\begin{itemize}
\item Divide the set of nodes $\mathcal{V}$ into non-overlapping $k$ partitions \resizebox{.15\textwidth}{!}{$\mathcal{V}^{(1)}, \mathcal{V}^{(2)}, \ldots, \mathcal{V}^{(k)}$}, each containing $\frac{|\mathcal{V}|}{k}$ nodes of the graph in chronological order. \resizebox{.15\textwidth}{!}{$\mathcal{V}=\mathcal{V}^{(1)} \cup \ldots \cup \mathcal{V}^{(k)}$} and \resizebox{.15\textwidth}{!}{$\mathcal{V}^{(i)} \cap \mathcal{V}^{(j)}=\emptyset$, $\forall i \neq j$}.
\item For each partition $\mathcal{V}^{(i)}$, find the set of $p$-hop neighbors $\mathcal{N}_p(u), u \in \mathcal{V}^{(i)}$ in the original graph $\mathcal{G}_t$, where $\mathcal{N}_p(u)$ defines the $p$-hop neighborhood of node $u$. Basically, we expand each partition to their $p$-hop neighbourhood in order to preserve the structural information between multiple partitions and utilize pair-wise graph connections: \resizebox{.215\textwidth}{!}{$\mathcal{V}^{(i)} \leftarrow \mathcal{V}^{(i)} \cup\left\{\mathcal{N}_{p}(u) \mid u \in \mathcal{V}^{(i)}\right\}$}.
\item Define the subgraph patch \resizebox{.185\textwidth}{!}{$\mathcal{G}^{(i)}_t=(\mathcal{V}^{(i)}\cup \mathcal{N}_p(u), \mathcal{E}^{(i)})$}, where $u \in \mathcal{V}^{(i)}$, $\mathcal{E}^{(i)}$ is the set of edges in $\mathcal{E}$ that have both endpoints in \resizebox{.15\textwidth}{!}{$\mathcal{V}^{(i)}\cup \mathcal{N}_p(u), u \in \mathcal{V}^{(i)}$}.
\end{itemize}

Repeat steps 2-3 for all partitions $\mathcal{V}^{(i)}$ to obtain the set of subgraph patches \resizebox{.15\textwidth}{!}{${\mathcal{G}^{(1)}_t, \mathcal{G}^{(2)}_t, \ldots, \mathcal{G}^{(k)}_t}$}. The subgraphs, $\mathcal{G}^{(i)}_{t}$ exhibit a diverse range of topological structures with varying numbers of nodes, edges, and connectivity, rendering them non-uniform in size. The subgraph encoder, which is applicable to arbitrary subgraph patches, captures the structural relationships between multiple-time series variables and generates fixed-length node-level vector representations of the subgraph patches. Subgraph encoding is particularly useful for large spatio-temporal graphs that are impractical to process as a whole, as it enables computation of node-level representations for subsets of the graph, rather than the entire graph. This enables capturing the essential structural information of spatio-temporal graphs while maintaining low computational complexity and memory requirements. The subgraph encoder uses a $p$-subtree GNN extractor to generate node-level representations for each subgraph patch \resizebox{.03\textwidth}{!}{$\mathcal{G}^{(i)}_{t}$}. The encoder extracts local structural and feature information by applying a GNN model(\cite{kipf2016semi}) to the subgraph with node feature vector, \resizebox{.05\textwidth}{!}{$\bar{\mathbf{x}}^{(i)}_{(u, \hspace{1mm}t)}$} for a given node $u \in \mathcal{V}^{(i)}$. The output node representation, denoted by $\mathbf{h}^{(i, t)}_{u}$ at $u$, serves as the subgraph representation at $u$, where the superscript $i$ denotes the subgraph patch. The $p$-subtree GNN extractor is a technique used in graph neural networks(GNNs) for feature extraction. For a node $u$ in a patch, the $p$-subtree GNN extractor recursively constructs all possible subtrees of radius p around it. Each subtree is treated as a separate subgraph, and a GNN is applied to each subgraph to aggregate features from the nodes within the subtree. The resulting feature vector represents the p-hop neighborhood of node $u$. The connectivity pattern for a given subgraph patch $\mathcal{G}^{(i)}_{t}$ is described by the patch adjacency matrix \resizebox{.145\textwidth}{!}{$\text{A}^{(i)} \in \{0,1\}^{|\mathcal{V}_{i}| \times |\mathcal{V}_{i}|}$}. To extract features from the subgraph, a GNN model with $p$ layers, denoted as \resizebox{.07\textwidth}{!}{$\mathrm{GNN}_{\mathcal{G}^{(i)}_{t}}^{(p)}$}, is applied to the subgraph patch $\mathcal{G}^{(i)}_{t}$ with node feature vector \resizebox{.05\textwidth}{!}{$\bar{\mathbf{x}}^{(i)}_{(u, \hspace{1mm} t)}$} and patch adjacency matrix \resizebox{.035\textwidth}{!}{$\text{A}^{(i)}$}, resulting in the output node representation, $\mathbf{h}^{(i, \hspace{1mm} t)}_{u}$ at node $u$. In concise form, we can express the subgraph representation learning function as:

\vspace{-4mm}
\resizebox{0.975\linewidth}{!}{
\hspace{1mm}\begin{minipage}{\linewidth}
\begin{align}
\mathbf{h}^{(i, t)}_{u} = \operatorname{GNN}_{\mathcal{G}^{(i)}_{t}}^{(p)}(u) \nonumber
\end{align}
\end{minipage}
} 

\vspace{0mm}
The $p$-subtree GNN extractor can represent the $p$-subtree structure rooted at node $u$. Since node $u$ may appear in multiple subgraph patches, we calculate the mean of its node representations across all subgraph patches \resizebox{.065\textwidth}{!}{$\mathcal{G}^{(i)}_t, i \in k$}, to produce a fixed-size vector representation \resizebox{.055\textwidth}{!}{$\mathbf{h}^{(t)}_{u\hspace{-0.5mm},\hspace{0.5mm}\text{SUB}}$} of each node $u$ in the original input graph \resizebox{.0175\textwidth}{!}{$\mathcal{G}_t$}. 

\vspace{-3mm}

\vspace{-5mm}
\subsubsection{Dual Hypergraph Representation Learning} 
\vspace{1mm}
Spatio-temporal Graph Neural Networks(STGNNs) have shown to be effective in modeling graph-structured data by integrating both node and edge features. Nevertheless, STGNNs have primarily focused on nodes and their connectivity, neglecting the significant role of edges in graph structure. Even with explicit edge representation, STGNNs face challenges in capturing critical edge information, resulting in limitations to their success. Furthermore, STGNNs utilize edge features as auxiliary information to enhance node-level representations, leading to suboptimal edge information capture. Edges capture the interaction, dependency, or similarity between nodes, which is essential in modeling the complex sensor network-based dynamical systems. By incorporating the latent edge information, STGNNs can more accurately represent the structure and dynamics of these complex systems, leading to better predictions and decision-making. Thus, more effective methods are necessary to represent edge information within STGNNs to achieve comprehensive and accurate spatio-temporal graph modeling. To overcome this challenge, a simple yet powerful message-passing scheme tailored specifically to edges has been proposed. This approach provides optimal edge representation, effectively addressing the limitations of previous STGNN approaches. \textcolor{black}{The proposed solution to address the limited edge representation in spatio-temporal graph modeling approaches involves a Dual Hypergraph Transformation(DHT) method that transforms edges into hypernodes and nodes into hyperedges, resulting in dual hypergraphs that can capture higher-order interactions among hypernodes. This graph-to-hypergraph transformation is influenced by hypergraph duality(\cite{berge1973graphs}, \cite{scheinerman2011fractional}). By using the DHT method to represent edges as hypernodes in a hypergraph, any existing hypergraph message-passing schemes designed for hypernode-level representation learning can be applied for learning the representation of the edges in the spatio-temporal graphs.} This hypergraph-based approach is particularly effective, as it enables more comprehensive and accurate graph modeling of spatio-temporal graphs by better representing the crucial edges information.

\vspace{-4mm}
\begin{figure}[!ht]
\resizebox{1.2\linewidth}{!}{
\hspace*{-5mm}    
\includegraphics[keepaspectratio,height=5cm, width=13cm,trim=0.5cm 5cm 0cm 3.5cm,clip]{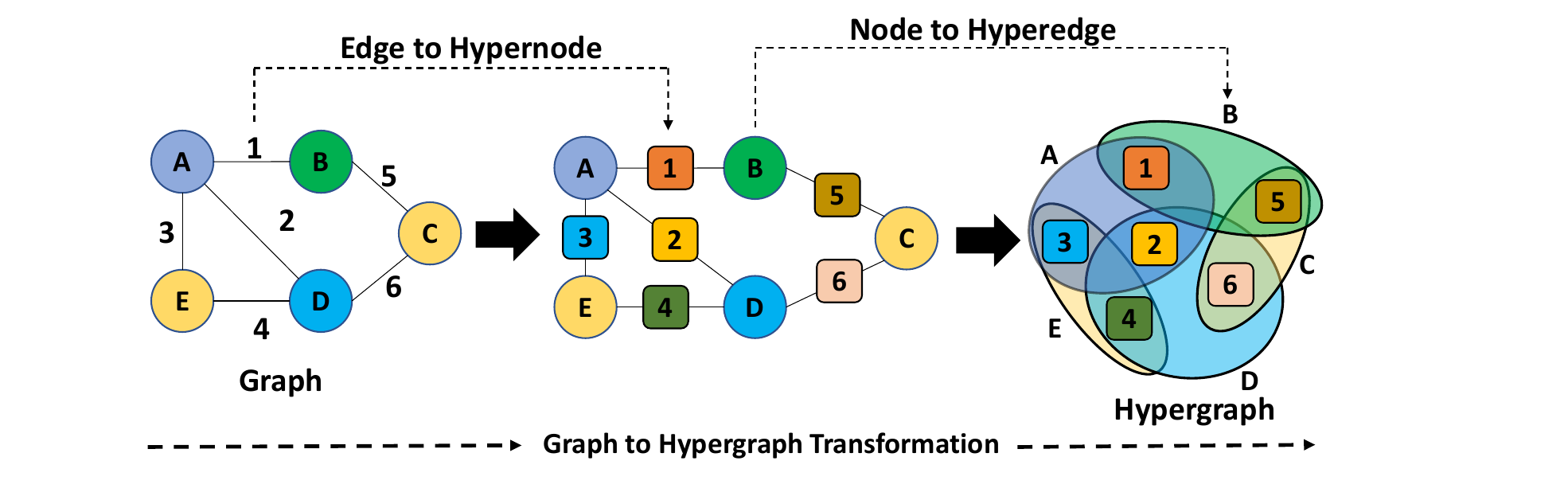} 
}
\vspace{-8mm}
\caption{The Dual Hypergraph Transformation(DHT) is an effective technique for transforming and analyzing complex spatio-temporal graph structures. It involves the graph-to-hypergraph transformation, which alters the roles of nodes and edges while preserving the shared connectivity pattern and original graph information. This powerful approach is illustrated in the figure above, where numbers(letters) represent the edges(nodes) in the original predefined graph, and hypernodes(hyperedges) in the dual hypergraph.}
\label{fig:figure4}
\end{figure}

\vspace{-2mm}
Let \resizebox{.25\textwidth}{!}{$\mathcal{G}_{t} = \big(\mathbf{X}_{(t : t + \upsilon-1)}, \text{I}^{(0)}, \mathbf{E}_{(t : t + \upsilon-1)}\big)$} denote the spatial-temporal graphs. The connections between the nodes and edges are obtained from the prior knowledge of time-series relationships, and they are described by the incidence matrix, \resizebox{.135\textwidth}{!}{$\text{I}^{(0)} \in \{0,1\}^{|\mathcal{V}| \times |\mathcal{E}|}$}.  \resizebox{.295\textwidth}{!}{$\mathbf{X}_{(t : t + \upsilon-1)} \in \mathbb{R}^{|\mathcal{V}| \times d}$,  $\mathbf{E}_{(t : t + \upsilon-1)} \in \mathbb{R}^{|\mathcal{E}| \times d}$} denote the node features and empty edge features, respectively. The Dual Hypergraph Transformation(DHT) is a method of transforming spatio-temporal graphs to obtain new dual hypergraphs. This is achieved by interchanging the roles of nodes and edges in spatio-temporal graphs, allowing for the use of hypernode-based message-passing methods to learn the edges representations while maintaining the original spatio-temporal graphs information. To achieve this transformation, the incidence matrix of the original spatio-temporal graph $\mathcal{G}_{t}$ is transposed to obtain the incidence matrix for the new dual spatio-temporal hypergraph $\mathcal{G}^{*}_{t}$. In addition to the structural transformation through the incidence matrix, the DHT interchanges node and edge features across $\mathcal{G}_{t}$ and $\mathcal{G}^{*}_{t}$. In brief, the transformation maps a spatio-temporal graph triplet representation to its corresponding dual spatio-temporal hypergraph representation while preserving the information in the original graph features, as follows:

\vspace{-3mm}
\resizebox{0.875\linewidth}{!}{
\hspace{-5mm}\begin{minipage}{\linewidth}
\begin{multline}
\boldsymbol{DHT}: \hspace{0.5mm}\mathcal{G}_{t} = \big(\mathbf{X}_{(t - \tau : \hspace{1mm}t-1)}, \text{I}^{(0)}, \mathbf{E}_{(t - \tau : \hspace{1mm}t-1)}\big)  \mapsto \\ \mathcal{G}^{*}_{t}= \big(\mathbf{E}_{(t - \tau : \hspace{1mm}t-1)}, {\text{I}^{(0)}}^{T}, \mathbf{X}_{(t - \tau : \hspace{1mm}t-1)}\big) \nonumber \
\end{multline}
\end{minipage}
}

\vspace{0.5mm}
Here, $\mathbf{E}_{(t - \tau : \hspace{1mm}t-1)}$, ${\text{I}^{(0)}}^{T}$, and $\mathbf{X}_{(t - \tau : \hspace{1mm}t-1)}$ represent the hypernode feature matrix, the incidence matrix, and the hyperedge feature matrix of the  dual hypergraph, $\mathcal{G}^{*}_{t}$, respectively. This precise and efficient approach to spatio-temporal graph transformation can capture different aspects of the underlying relational structure of interconnected dynamical systems and improves downstream forecasting accuracy. We represent the MTS data as the hyperedge-attributed dual hypergraphs($\mathcal{G}^{*}_{t}$), and the dual hypergraph-structured MTS data are processed by a hypergraph representation learning(HgRL) module. The HgRL computes optimal hyperedge-level representations \resizebox{.055\textwidth}{!}{$\mathbf{h}^{(t)}_{u\hspace{-0.5mm},\hspace{0.5mm}\text{DHT}}$} that capture the spatio-temporal dynamics within the dual hypergraph-structured MTS data. These representations are further utilized for downstream forecasting tasks, allowing for accurate and reliable multi-horizon forecasts.

\vspace{-2mm}
\subsection{TEMPORAL-INFERENCE}
\vspace{-1mm}
The mixture-of-experts(MOE) mechanism in deep learning combines the predictions of multiple subnetworks or experts, such as ``implicit hypergraph", ``explicit subgraph", and ``dual-hypergraph" representation learning methods, through a gating mechanism that calculates a weighted sum of their predictions based on the input. The objectives of training are to identify the optimal distribution of weights for the gating function and to train the experts using the specified weights. \textcolor{black}{In the context of cooperative game theory, the mixture of experts can be viewed as a cooperative game where the experts work together to optimize the system's performance by accurately predicting the output given an input, rather than maximizing their individual payoffs. The gating mechanism can be trained to optimize the weights or probabilities assigned to each agent's expertise, based on their individual performance and the overall performance of the system, resulting in a globally optimal solution}. To obtain fused representations in the MOE mechanism, the multiple experts predictions are combined using the weights calculated by the gating mechanism as follows,

\vspace{-3mm}
\resizebox{0.925\linewidth}{!}{
\hspace{-2.0mm}\begin{minipage}{\linewidth}
\begin{align}
g^{\prime\prime} &= \sigma \big( f^{\prime\prime}_s(\mathbf{h^{(t)}_{i, IMP}} + \mathbf{h^{(t)}_{i, SUB}}) + f^{\prime\prime}_g(\mathbf{h^{(t)}_{i, DHT}}) \big) \nonumber \\
\mathbf{h^{(t)}_{i}}  &= \sigma \big( g^{\prime\prime}(\mathbf{h^{(t)}_{i, IMP}} + \mathbf{h^{(t)}_{i, SUB}})) + (1-g^{\prime\prime})(\mathbf{h^{(t)}_{i, DHT}}) \big) \nonumber 
\end{align}
\end{minipage}
} 

\vspace{1mm}
The ``implicit hypergraph", ``explicit subgraph", and ``dual-hypergraph" methods compute the representations \resizebox{.20\textwidth}{!}{$\mathbf{h^{(t)}_{i, IMP}}$, $\mathbf{h^{(t)}_{i, SUB}}, \mathbf{h^{(t)}_{i, DHT}}$}, respectively, where \resizebox{.02\textwidth}{!}{$f^{\prime\prime}_s$} and \resizebox{.02\textwidth}{!}{$f^{\prime\prime}_g$} are linear projections. The fused representations are input to a subsequent temporal feature extractor that models the nonlinear temporal dynamics of inter-series dependencies among variables in the spatio-temporal MTS data using a stack of $1 \times 1$ convolutions. Finally, the extractor predicts the pointwise forecasts, \resizebox{.085\textwidth}{!}{$\hat{\mathbf{X}}_{(t  : t + \upsilon-1)}$}. Our proposed framework uses a spatial-then-time modeling approach to learn the multiple structure representations and dynamics in MTS data. By first encoding the spatial information of the relational structure, including explicit graph, implicit hypergraph, and dual hypergraph our approach captures complex  dependencies among the variables. By incorporating the temporal inference component, the framework analyzes the evolution of these dependencies over time to improve interpretability and generalization. This approach is beneficial for real-world applications involving complex spatial-temporal dependencies that are challenging to model using traditional methods. Additionally, our framework variant(\textbf{w/Unc- MKH-Net}) provides accurate and reliable uncertainty estimates of multi-horizon forecasts by minimizing the negative Gaussian log likelihood. Our proposed methods(\textbf{MKH-Net}, \textbf{w/Unc- MKH-Net}) enable simultaneous modeling of latent interdependencies and analyzing their evolution over time in sensor network-based dynamical systems in an end-to-end manner.

\vspace{-3.5mm}
\begin{figure}[!ht]
\hspace*{-7mm}
\resizebox{1.325\linewidth}{!}{  
\includegraphics[keepaspectratio,height=7cm, width=15cm,trim=2cm 2.1cm 2cm 2.5cm,clip]{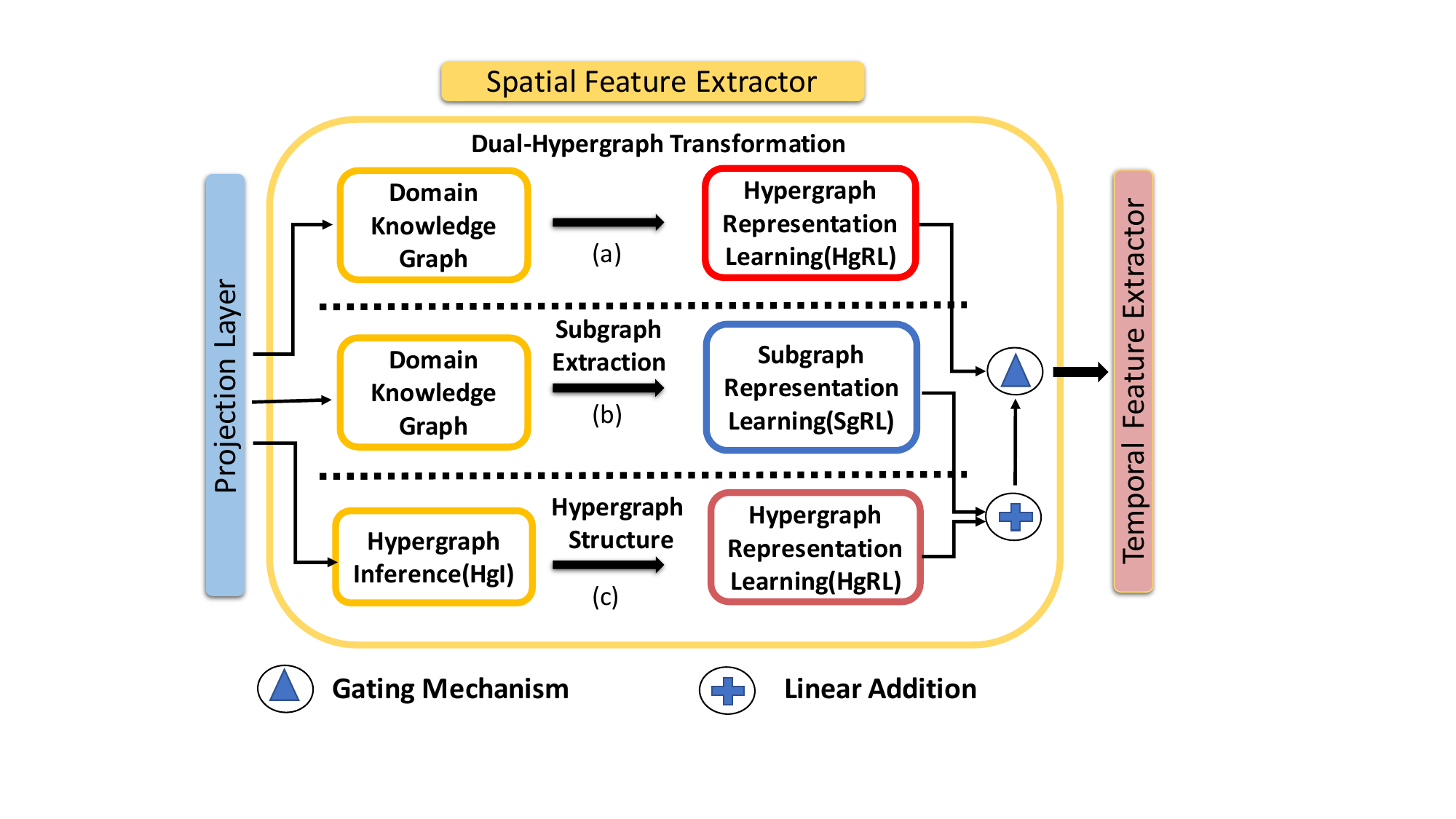} 
\vspace{-8mm}
}
\vspace{-6mm}
\caption{Usage of multiple spatial feature extractors has significant benefits by providing a comprehensive and diverse representation of MTS data and reducing the risk of overfitting to a specific aspect of the data by incentivizing each extractor to specialize in a different aspect. This collaborative mechanism design optimizes forecasting error by leveraging the strengths of multiple extractors and fusing their representations to enhance the accuracy and robustness of framework predictions, resulting in more reliable outcomes. In the above figure, subfigures (a), (b), and (c) illustrate the ``dual-hypergraph", ``explicit subgraph", and ``implicit hypergraph" representation learning methods, respectively.}
\label{fig:figure5}
\end{figure}

\begin{table*}[!ht]
\centering
\setlength{\tabcolsep}{3pt}
\renewcommand\arraystretch{1.15}
 \resizebox{\textwidth}{!}{
\begin{tabular}{c|c|ccc|c|ccc|c|ccc|c|ccc|c|ccc}
\hline
\textbf{Model} & \multirow{23}{*}{\textbf{\rotatebox[origin=c]{90}{PeMSD3}}} & \textbf{MAE} & \textbf{RMSE} & \textbf{MAPE} & \multirow{23}{*}{\textbf{\rotatebox[origin=c]{90}{PeMSD4}}} & \textbf{MAE} & \textbf{RMSE} & \textbf{MAPE} & \multirow{23}{*}{\textbf{\rotatebox[origin=c]{90}{PeMSD7}}} & \textbf{MAE} & \textbf{RMSE} & \textbf{MAPE} & \multirow{23}{*}{\textbf{\rotatebox[origin=c]{90}{PeMSD8}}} & \textbf{MAE} & \textbf{RMSE} & \textbf{MAPE} & \multirow{23}{*}{\textbf{\rotatebox[origin=c]{90}{PeMSD7(M)}}} & \textbf{MAE} & \textbf{RMSE} & \textbf{MAPE} \\ \cline{1-1} \cline{3-5} \cline{7-9} \cline{11-13} \cline{15-17} \cline{19-21}
HA &  & 31.58 & 52.39 & 33.78 &  & 38.03 & 59.24 & 27.88 &  & 45.12 & 65.64 & 24.51 &  & 34.86 & 59.24 & 27.88 &  & 4.59 & 8.63 & 14.35 \\
ARIMA &  & 35.41 & 47.59 & 33.78 &  & 33.73 & 48.80 & 24.18 &  & 38.17 & 59.27 & 19.46 &  & 31.09 & 44.32 & 22.73 &  & 7.27 & 13.2 & 15.38 \\
VAR &  & 23.65 & 38.26 & 24.51 &  & 24.54 & 38.61 & 17.24 &  & 50.22 & 75.63 & 32.22 &  & 19.19 & 29.81 & 13.10 &  & 4.25 & 7.61 & 10.28 \\
FC-LSTM &  & 21.33 & 35.11 & 23.33 &  & 26.77 & 40.65 & 18.23 &  & 29.98 & 45.94 & 13.20 &  & 23.09 & 35.17 & 14.99 &  & 4.16 & 7.51 & 10.10 \\
TCN &  & 19.32 & 33.55 & 19.93 &  & 23.22 & 37.26 & 15.59 &  & 32.72 & 42.23 & 14.26 &  & 22.72 & 35.79 & 14.03 &  & 4.36 & 7.20 & 9.71 \\
TCN(w/o causal) &  & 18.87 & 32.24 & 18.63 &  & 22.81 & 36.87 & 14.31 &  & 30.53 & 41.02 & 13.88 &  & 21.42 & 34.03 & 13.09 &  & 4.43 & 7.53 & 9.44 \\
GRU-ED &  & 19.12 & 32.85 & 19.31 &  & 23.68 & 39.27 & 16.44 &  & 27.66 & 43.49 & 12.20 &  & 22.00 & 36.22 & 13.33 &  & 4.78 & 9.05 & 12.66 \\
DSANet &  & 21.29 & 34.55 & 23.21 &  & 22.79 & 35.77 & 16.03 &  & 31.36 & 49.11 & 14.43 &  & 17.14 & 26.96 & 11.32 &  & 3.52 & 6.98 & 8.78 \\
STGCN &  & 17.55 & 30.42 & 17.34 &  & 21.16 & 34.89 & 13.83 &  & 25.33 & 39.34 & 11.21 &  & 17.5 & 27.09 & 11.29 &  & 3.86 & 6.79 & 10.06 \\
DCRNN &  & 17.99 & 30.31 & 18.34 &  & 21.22 & 33.44 & 14.17 &  & 25.22 & 38.61 & 11.82 &  & 16.82 & 26.36 & 10.92 &  & 3.83 & 7.18 & 9.81 \\
GraphWaveNet &  & 19.12 & 32.77 & 18.89 &  & 24.89 & 39.66 & 17.29 &  & 26.39 & 41.5 & 11.97 &  & 18.28 & 30.05 & 12.15 &  & 3.19 & 6.24 & 8.02 \\
ASTGCN(r) &  & 17.34 & 29.56 & 17.21 &  & 22.93 & 35.22 & 16.56 &  & 24.01 & 37.87 & 10.73 &  & 18.25 & 28.06 & 11.64 &  & 3.14 & 6.18 & 8.12 \\
MSTGCN &  & 19.54 & 31.93 & 23.86 &  & 23.96 & 37.21 & 14.33 &  & 29.00 & 43.73 & 14.30 &  & 19.00 & 29.15 & 12.38 &  & 3.54 & 6.14 & 9.00 \\
STG2Seq &  & 19.03 & 29.83 & 21.55 &  & 25.20 & 38.48 & 18.77 &  & 32.77 & 47.16 & 20.16 &  & 20.17 & 30.71 & 17.32 &  & 3.48 & 6.51 & 8.95 \\
LSGCN &  & 17.94 & 29.85 & 16.98 &  & 21.53 & 33.86 & 13.18 &  & 27.31 & 41.46 & 11.98 &  & 17.73 & 26.76 & 11.20 &  & 3.05 & 5.98 & 7.62 \\
STSGCN &  & 17.48 & 29.21 & 16.78 &  & 21.19 & 33.65 & 13.9 &  & 24.26 & 39.03 & 10.21 &  & 17.13 & 26.8 & 10.96 &  & 3.01 & 5.93 & 7.55 \\
AGCRN &  & 15.98 & 28.25 & 15.23 &  & 19.83 & 32.26 & 12.97 &  & 22.37 & 36.55 & 9.12 &  & 15.95 & 25.22 & 10.09 &  & 2.79 & 5.54 & 7.02 \\
STFGNN &  & 16.77 & 28.34 & 16.30 &  & 20.48 & 32.51 & 16.77 &  & 23.46 & 36.6 & 9.21 &  & 16.94 & 26.25 & 10.60 &  & 2.90 & 5.79 & 7.23 \\
STGODE &  & 16.50 & 27.84 & 16.69 &  & 20.84 & 32.82 & 13.77 &  & 22.59 & 37.54 & 10.14 &  & 16.81 & 25.97 & 10.62 &  & 2.97 & 5.66 & 7.36 \\
Z-GCNETs &  & 16.64 & 28.15 & 16.39 &  & 19.50 & 31.61 & 12.78 &  & 21.77 & 35.17 & 9.25 &  & 15.76 & 25.11 & 10.01 &  & 2.75 & 5.62 & 6.89 \\ \cline{1-1} \cline{3-5} \cline{7-9} \cline{11-13} \cline{15-17} \cline{19-21} 
\textbf{MKH-Net} &  & \textbf{13.177} & \textbf{19.937} & \textbf{11.679} &  & \textbf{19.194} & \textbf{28.925} & \textbf{12.259} &  & \textbf{21.084} & \textbf{35.087} & \textbf{8.768} &  & \textbf{13.605} & \textbf{21.241} & \textbf{7.762} &  & \textbf{2.42} & \textbf{5.16} & \textbf{6.67} \\
\textbf{w/Unc- MKH-Net} &  & 13.376 & 20.154 & 11.817 &  & 19.494 & 28.879 & 12.566 &  & 21.348 & 35.226 & 8.934 &  & 12.871 & 19.735 & 7.721 &  & - & - & - \\ \hline
\end{tabular}
}
\vspace{-2mm}
\caption{The forecast errors were estimated based on model predictions on benchmark datasets for horizon@12. The symbol ``-" indicates an Out Of Memory(OOM) error.}
\label{tab:results1}
\end{table*}

\vspace{-7mm}
\section{DATASETS}
\vspace{-1mm}
Our study involves conducting experiments to evaluate the efficacy of two novel models(\textbf{MKH-Net}, \textbf{w/Unc-MKH-Net}) on large-scale spatial-temporal datasets. These datasets, comprising PeMSD3, PeMSD4, PeMSD7, PeMSD7(M), and PeMSD8, contain real-world traffic information from the Caltrans Performance Measurement System (PeMS, \cite{chen2001freeway}), which records real-time traffic flow measurements. To ensure fairness and consistency with prior research, we conducted a preprocessing step on all benchmark datasets.  This involved aggregating the 30-second interval data into 5-minute averages, as per the method proposed by \cite{choi2022graph}. Moreover, we made use of publicly accessible datasets for traffic flow prediction(METR-LA and PEMS-BAY) as presented by \cite{LiYS018}. To ensure alignment with our methodology, we converted the datasets into 5-minute interval averages, resulting in 288 observations daily. By adopting this approach, we can effectively demonstrate the potential and benefits of our proposed methodology for analyzing and modeling complex spatio-temporal MTS data, surpassing existing methods. For further information on the benchmark datasets utilized, please refer to the appendix.

\vspace{-3mm}
\section{EXPERIMENTAL RESULTS}
\vspace{-1mm}
Table \ref{tab:results1} presents a thorough comparison of the proposed models(\textbf{MKH-Net} and \textbf{w/Unc-MKH-Net}), against various baseline models on the MTSF task across five distinct benchmark datasets(PeMSD3, PeMSD4, PeMSD7, PeMSD7M, and PeMSD8). We evaluated the models performance using forecast errors for a widely recognized benchmark of 12($\tau$)-step-prior to 12($\upsilon$)-step-ahead forecasting task. Our evaluation employed a multi-metric approach in multi-horizon prediction tasks for a comprehensive evaluation of the models performance compared to baseline models. To ensure a comprehensive and robust evaluation of the model's performance, various performance metrics, such as mean absolute error(MAE), root mean squared error(RMSE), and mean absolute percentage error(MAPE), were employed in the assessment. The results of the baseline models from \cite{choi2022graph} are reported in this current study. Our experimental results demonstrate that the proposed models(\textbf{MKH-Net}, \textbf{w/Unc-MKH-Net}) consistently outperform baseline models with lower forecast errors across the various benchmark datasets. The proposed model(\textbf{MKH-Net}) achieved a significant reduction of 28.39$\%$, 8.50$\%$, 0.24$\%$, 15.41$\%$, and 6.86$\%$ in the RMSE metric compared to the next-best baseline models on the PeMSD3, PeMSD4, PeMSD7, PeMSD8, and PeMSD7(M) datasets, respectively. In addition to pointwise forecasts, the \textbf{w/Unc-MKH-Net} model(\textbf{MKH-Net} integrated with local uncertainty estimation) predicts time-varying estimates of uncertainty in  model predictions. Despite its slightly inferior performance to the \textbf{MKH-Net} model, it still outperforms several robust baselines in the literature, as evidenced by the reduced prediction error. The empirical results highlight the effectiveness of the proposed neural forecasting architecture in capturing the complex and nonlinear spatio-temporal dynamics present in MTS data, resulting in improved forecasts. Further information on the experimental methodology, ablation studies, and additional experimental results can be found in the appendix. The appendix also includes a detailed analysis of the \textbf{MKH-Net} capability to handle missing data, and provides a more in-depth analysis of the \textbf{w/Unc-MKH-Net} ability to estimate uncertainty. Moreover, the appendix provides comprehensive visualizations of model predictions with uncertainty estimates compared to the ground truth and offers additional information on existing works and a brief overview of baselines. Lastly, the appendix addresses the subject of compositional generalization in the context of graph time series forecasting. 

\vspace{-3mm}
\section{CONCLUSION}
\vspace{-1mm}
Our proposed framework integrates implicit hypergraph, explicit subgraph, and dual-hypergraph representation learning methods to provide a thorough understanding of the spatio-temporal dynamics in MTS data, enabling accurate multi-horizon forecasting. Our approach has been validated by experimental results on real-world datasets, demonstrating its effectiveness through improved multi-horizon forecasts and reliable uncertainty estimations. 

\vspace{-3mm}
\bibliographystyle{named}
\bibliography{ijcai23}

\clearpage
\newpage

\section{APPENDIX}

\subsection{HYPERGRAPH ATTENTION NETWORK(HgAT)}
The Hypergraph Attention Network(HgAT) operator extends attention-based convolution operations to spatio-temporal hypergraphs, allowing for more flexibility and expressiveness in modeling complex relationships among time series variables in large MTS datasets. By incorporating both local and global attention-based convolution operations, the HgAT operator
efficiently learns hypergraph representations that captures the complex spatio-temporal dynamics within the hypergraph-structured MTS data, making it a powerful and flexible tool for spatio-temporal data analysis and modeling. This hypergraph encoder performs inference on hypergraph-structured MTS data, denoted by \resizebox{.235\textwidth}{!}{$\mathcal{HG}_{t} = \big(\mathcal{HV}, \mathcal{HE}, \bar{\mathbf{X}}_{(t : t + \upsilon-1)}, \text{I}\big)$}, to compute the hypernode representation matrix, \resizebox{.14\textwidth}{!}{$\mathbf{H}_{(t : t + \upsilon-1)} \in \mathbb{R}^{n \times d}$}, wherein each row denotes the hypernode representations, \resizebox{.085\textwidth}{!}{$\mathbf{h}^{(t)}_{i} \in \mathbb{R}^{(d)}$} encapsulating the intricate dynamics of the hypergraph-structured MTS data. Here, \resizebox{.035\textwidth}{!}{$\mathcal{HG}_{t}$} at time step t, is characterized by the incidence matrix, \resizebox{.075\textwidth}{!}{$\mathbf{I} \in \mathbb{R}^{n \times m}$}, and feature matrix, \resizebox{.14\textwidth}{!}{$\bar{\mathbf{X}}_{(t : t + \upsilon-1)} \in \mathbb{R}^{n \times d}$}. The HgAT operator consistently captures the time-evolving dependencies of multiple variables in hypernode representations by encoding both structural and feature attributes of spatio-temporal hypergraphs, thereby modeling the intricate interdependencies and relationships among the variables. Let $\mathcal{N}_{j, i}$ represents a subset of hypernodes $i$ associated with a specific hyperedge $j$. The intra-edge neighborhood of a hypernode i, described as $\mathcal{N}_{j,i} \backslash i$, comprises a localized group of semantically-correlated time series variables, where the hyperedge $j$ captures the higher-order relationships among the incident hypernodes. Meanwhile, the inter-edge neighborhood of a hypernode $i$, denoted as $\mathcal{N}_{i, j}$, encompasses the set of hyperedges $j$ connected with hypernode $i$, further enriching the relationships between the hypernodes and hyperedges. The HgAT operator utilizes the relational inductive bias encoded within the hypergraph's connectivity, and performs the intra-edge and inter-edge neighborhood aggregation schemes to explicitly model the spatio-temporal correlations among time series variables. The intra-edge neighborhood aggregation delves into the interrelations between a particular hypernode $i$ and its neighboring hypernodes $\mathcal{N}_{j,i} \backslash i$ connected by a specific hyperedge $j$, while the inter-edge neighborhood aggregation considers the relationships between a specific hypernode $i$ and all other hyperedges $\mathcal{N}_{i, j}$ incident with it. We perform the attention-based intra-edge neighborhood aggregation to learn the latent hyperedge representations, which helps us better understand the hypergraph-structured MTS data, described as follows.

\vspace{-1mm}
\resizebox{0.9\linewidth}{!}{
\begin{minipage}{\linewidth}
\begin{equation}
\mathbf{h}^{(t, \ell)}_{j} =  \sum_{z=1}^{\mathcal{Z}} \sigma \big( \hspace{-0.25mm}  \sum_{i \hspace{0.5mm}\in \hspace{0.5mm}{\mathcal{N}_{j, i}}} \hspace{-1mm}  \alpha^{(t, \ell, z)}_{j, i} \mathbf{W}^{(z)}_{0}\mathbf{h}^{(t, \ell-1, z)}_{i} \big) \nonumber
\end{equation}
\end{minipage}
}

where the hyperedge representations are denoted by $\mathbf{h}_{j} \in \mathbb{R}^{(d)}$, with the layer denoted by $\ell$. Each hypernode is initially represented by its corresponding feature vector, $\mathbf{h}^{(t, 0, z)}_{i} = \hspace{1mm}\bar{\mathbf{x}}^{(t)}_{i}$, where $\bar{\mathbf{x}}^{(t)}_{i} \in \mathbb{R}^{(d)}$ denotes the $i^{th}$ row of the feature matrix \resizebox{.14\textwidth}{!}{$\bar{\mathbf{X}}_{(t : t + \upsilon-1)} \in \mathbb{R}^{n \times d}$}. The symbol $\sigma$ represents the sigmoid function. The HgAT operator generates multiple representations of the input data, $\mathbf{h}^{(t, \ell-1, z)}_{j}$ represented by superscript $z$ each with its own set of parameters. These representations are then combined by summation, akin to the multi-head self-attention mechanism(\cite{vaswani2017attention}). This approach enables the HgAT operator to capture various underlying aspects of the hypergraph-structured MTS data. The attention coefficient $\alpha_{j, i}$ is computed by determining the relative importance of the hypernode $i$ that is incident with hyperedge $j$ and is computed by,

\vspace{-1mm}
\resizebox{0.925\linewidth}{!}{
\hspace{0.0cm}\begin{minipage}{\linewidth}
\begin{align}
e^{(t, \ell, z)}_{j, i} &= \operatorname{ReLU}\big(\text{W}^{(z)}_{0} \mathbf{h}^{(t, \ell-1, z)}_{i}\big) \nonumber \\
\alpha^{(t, \ell, z)}_{j, i} &= \frac{\exp \big(e^{(t, \ell, z)}_{j, i}\big)}{{\textstyle \sum_{k \hspace{0.5mm}\in \hspace{0.5mm}{\mathcal{N}_{j, i}}} \exp \big(e^{(t, \ell, z)}_{j, k}\big)}}  \nonumber
\end{align}
\end{minipage}
}

\vspace{1mm}
where $e_{j, i}$ denotes the unnormalized attention score. The HgAT method utilizes an attention-based inter-edge neighborhood aggregation scheme, which allows for the capture of complex dependencies and relationships between hyperedges and hypernodes. This aggregation scheme computes expressive hypernode representations by summing over ReLU activations of linear transformations of previous layer hypernode representations and weighted hyperedge representations as described below,

\vspace{-2mm}
\resizebox{0.9\linewidth}{!}{
\hspace{-0.01cm}\begin{minipage}{\linewidth}
\begin{equation}
\mathbf{h}^{(t, \ell)}_{i}=\sum_{z=1}^{\mathcal{Z}} \operatorname{ReLU}\big(\text{W}^{(z)}_{0}\mathbf{h}^{(t, \ell-1, z)}_{i} + \sum_{j \in \mathcal{N}_{i, j}} \beta^{(t, \ell, z)}_{i, j} \text{W}^{(z)}_{1} \mathbf{h}^{(t, \ell, z)}_{j}\big) \nonumber
\end{equation}
\end{minipage}
} 

\vspace{1mm}
Where \resizebox{.14\textwidth}{!}{$\text{W}^{(z)}_{0}, \text{W}^{(z)}_{1} \in \mathbb{R}^{d \times d}$} denotes the trainable weight matrices. The $\operatorname{ReLU}$ activation function is utilized to introduce non-linearity for updating the hypernode-level representations. The normalized attention scores $\beta_{i, j}$ determine the importance of each hyperedge $j$ that is incident with hypernode $i$, enabling the HgAT operator to focus on the most relevant hyperedges, computed by,

\vspace{-1mm}
\resizebox{0.945\linewidth}{!}{
\hspace{-0.01cm}\begin{minipage}{\linewidth}
\begin{align}
\phi^{(t, \ell, z)}_{i, j} &= \operatorname{ReLU}\big(\text{W}^{(z)}_{3} \cdot \big(\text{W}^{(z)}_{2}\mathbf{h}^{(t, \ell-1, z)}_{i} \oplus \text{W}^{(z)}_{2} \mathbf{h}^{(t, \ell, z)}_{j}\big)\big) \nonumber \\
\beta^{(t, \ell, z)}_{i, j} &=  \frac{\exp (\phi^{(t, \ell, z)}_{i, j})}{{\textstyle \sum_{k \hspace{0.5mm}\in \hspace{0.5mm}{\mathcal{N}_{i, j}}} \exp (\phi^{(t,\ell, z)}_{i, k})}}\nonumber
\end{align}
\end{minipage}
} 

\vspace{1mm}
Where \resizebox{.09\textwidth}{!}{$\text{W}^{(z)}_{2} \in \mathbb{R}^{d \times d}$} and \resizebox{.0825\textwidth}{!}{$\text{W}^{(z)}_{3} \in \mathbb{R}^{2d}$} are trainable weight matrix and vector, respectively. $\oplus$ denotes the concatenation operator. The unnormalized attention score is denoted by $\phi_{i, j}$. We utilize batch normalization and dropout techniques to enhance the generalization performance of the HgAT operator and mitigate overfitting, thereby increasing its reliability and accuracy for downstream MTSF task. A gating mechanism is used to selectively combine features from $\bar{\mathbf{x}}^{(t)}_{i}$ and $\mathbf{h}^{(t, \ell)}_{i}$, which is regulated through a differentiable approach and are described below,

\vspace{-2mm}
\resizebox{0.965\linewidth}{!}{
\hspace{0cm}\begin{minipage}{\linewidth}
\begin{align}
g^{(t)}  &= \sigma \big( f_s(\mathbf{h}^{(t, \ell)}_{i}) + f_g(\bar{\mathbf{x}}^{(t)}_{i}) \big)  \nonumber \\
\mathbf{h}^{(t, \ell)}_{i}  &= \sigma \big( g^{(t)}(\mathbf{h}^{(t, \ell)}_{i}) + (1-g^{(t)})(\bar{\mathbf{x}}^{(t)}_{i}) \big) \nonumber
\end{align}
\end{minipage}
} 

\vspace{1mm}
where $f_s$ and $f_g$ represent linear projections. This design choice allows the HgAT operator to capture the relationships between the different time-series variables and their changes over time, which leads to improved forecast accuracy. In brief, the HgAT operator is an effective tool for encoding and analyzing spatio-temporal hypergraphs. 

\subsection{HYPERGRAPH TRANSFORMER(HgT)}
The proposed Hypergraph Transformer (HgT) operator is an extension of transformer networks (\cite{vaswani2017attention}), designed to handle arbitrary sparse hypergraph structures with full attention as a desired structural inductive bias. The HgT operator allows the model to attend to all hypernodes in the hypergraph, enabling the learning of fine-grained interrelations unconstrained by domain-specific hierarchical structural information underlying the multivariate time series data. This approach facilitates the learning of optimal hypergraph representations by allowing the model to span large receptive fields for global reasoning of the complex dependencies within hypergraph-structured MTS data. Unlike existing methods, such as stacking multiple neural network layers with residual connections (\cite{fey2019just}, \cite{xu2018representation}), virtual hypernode mechanisms (\cite{gilmer2017neural}, \cite{ishiguro2019graph}, \cite{pham2017graph}), or hierarchical pooling schemes (\cite{rampavsek2021hierarchical}, \cite{gao2019graph}, and \cite{lee2019self}), the HgT operator does not rely on structural priors to model the long-range correlations in the hypergraph-structured MTS data. The permutation-invariant HgT module leverages global contextual information to model pairwise relations between all hypernodes in hypergraph-structured MTS data. As a result, the HgT module serves as a drop-in replacement for existing methods in modeling hierarchical dependencies and relationships among time-series variables in spatio-temporal hypergraphs, resulting in more robust and generalizable representations for a variety of downstream tasks. The transformer encoder(\cite{vaswani2017attention}) comprises alternating layers of multiheaded self-attention(MSA) and multi-layer perceptron(MLP) blocks that capture both local and global contextual information. Layer normalization (LN(\cite{ba2016layer})) and residual connections are applied after each block to improve performance and regularize the HgT operator. Inspired by ResNets(\cite{he2016deep}), the transformer encoder is designed to address vanishing gradients and over-smoothing issues by incorporating skip-connections through an initial connection strategy, allowing the HgT operator to learn complex and deep representations of the hypergraph-structured MTS data.

\vspace{-3mm}
\resizebox{0.95\linewidth}{!}{
\hspace{0.0cm}\begin{minipage}{\linewidth}
\begin{align}
\mathbf{h^{\prime}_{i}}^{(t, \ell)} &= \hspace{1mm}\operatorname{MSA}\big(\operatorname{LN}\big(\mathbf{h}^{(t, \ell-1)}_{i}  \big)\big) + \mathbf{h}^{(t, \ell-1)}_{i}  \nonumber \\
\mathbf{h^{\prime}_{i}}^{(t, \ell)} &= \hspace{1mm}\operatorname{MLP}\big(\operatorname{LN}\big(\mathbf{h^{\prime}_{i}}^{(t, \ell)}\big)\big)+ \bar{\mathbf{x}}^{(t)}_{i}  \nonumber
\end{align}
\end{minipage}
} 

\vspace{2mm}
The proposed method initializes each hypernode's representation to its corresponding feature vector, $\mathbf{h}^{(t, 0)}_{v_{i}} = \hspace{1mm}\bar{\mathbf{x}}^{(t)}_{i} \in \hspace{0.5mm}\mathbb{R}^{(d)}$. The HgT method is an effective approach for summarizing hypergraph-structured MTS data, overcoming the limitations of the representational capacity in HgAT operator. The HgT operator captures task-specific interrelations between hypernodes beyond the original sparse structure and distills long-range information in downstream layers to learn expressive, task-specific hypergraph representations to improve forecast accuracy.

\vspace{-1mm}
\subsection{ADDITIONAL RESULTS}
\begin{figure}[ht!]
\centering
\begin{minipage}[!h]{0.825\columnwidth}
  \includegraphics[width=\linewidth]{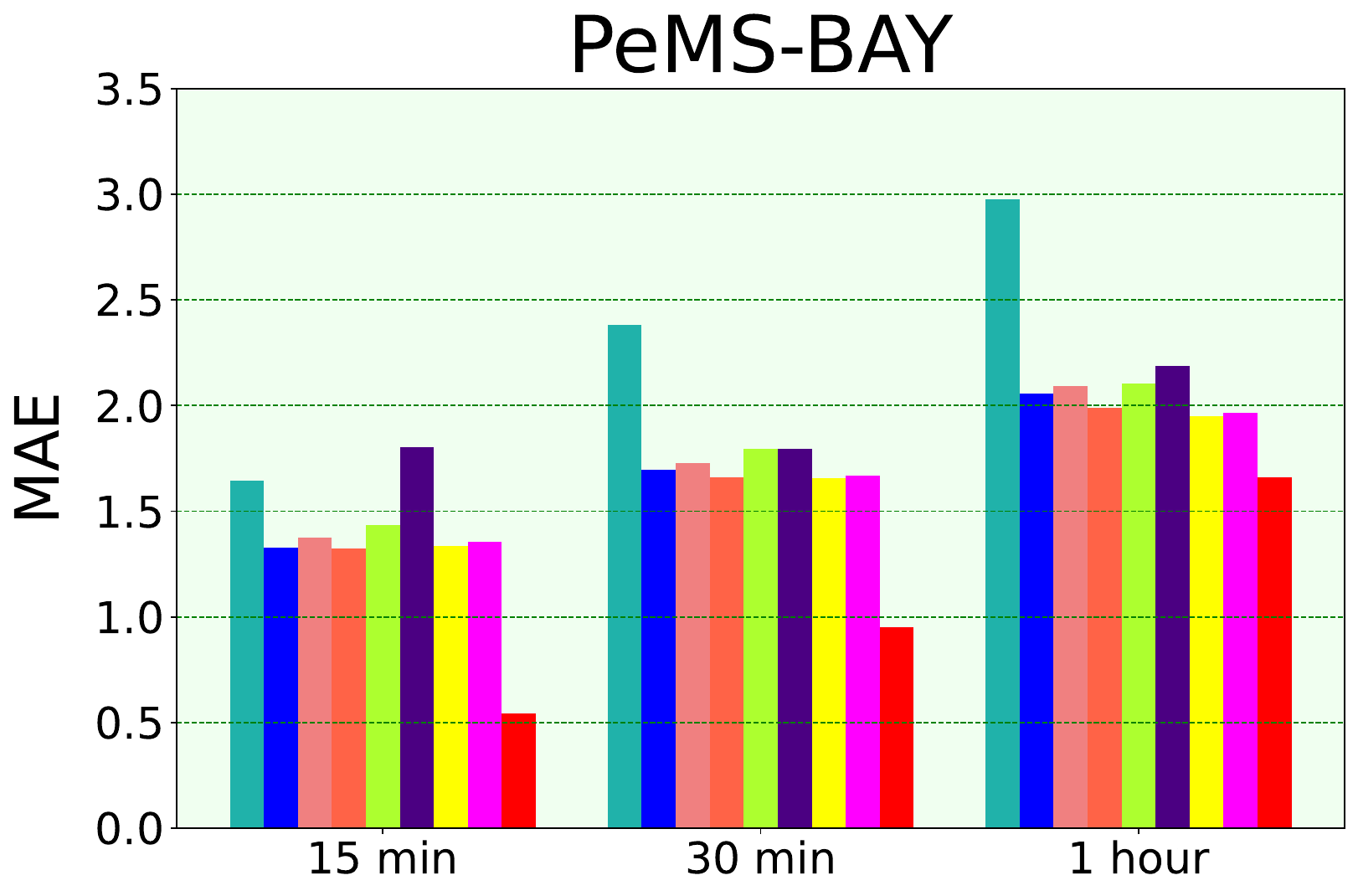}
\end{minipage}\vspace{1mm} 
\begin{minipage}[!h]{0.925\columnwidth}
\hspace*{4mm}\includegraphics[width=\linewidth]{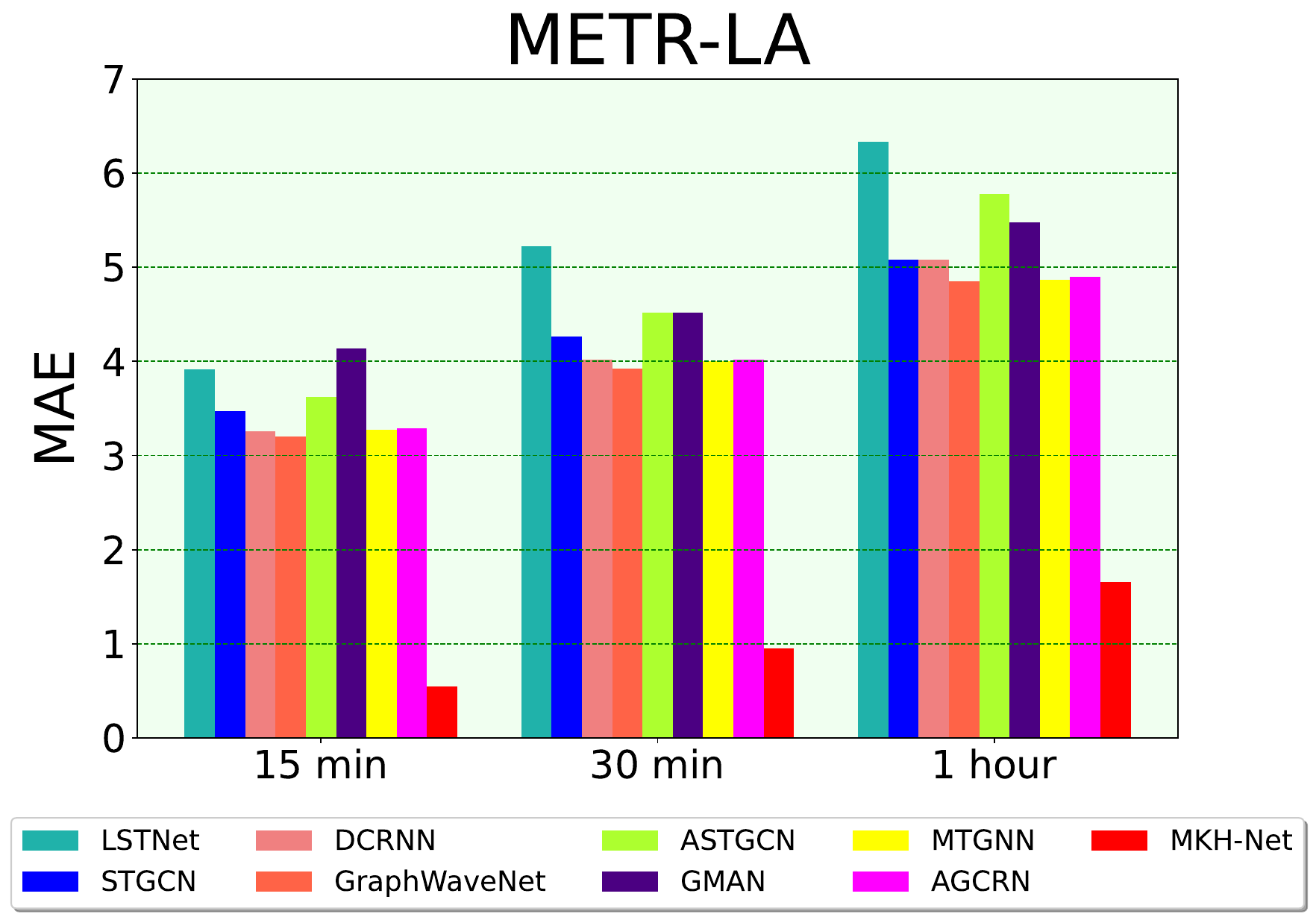}
\end{minipage}
\vspace{1mm}
\caption{The figure compares the pointwise forecast error of the proposed model(\textbf{MKH-Net}), with that of the baseline models for multiple prediction horizons on both the METR-LA and PeMS-BAY benchmark datasets. The results show that proposed model consistently outperformed the baseline models by a significant margin, demonstrating its robustness and effectiveness on MTSF task.}
\label{fig-prediction1}
\end{figure}

Figure \ref{fig-prediction1} shows how different models(\textbf{MKH-Net} and some baseline models) perform on two datasets(METR-LA and PEMS-BAY) for the MTSF task. The models performance is evaluated using a range of metrics, including MAE, RMSE, and MAPE, with corresponding forecast errors reported for 3-, 6-, and 12-steps-ahead forecast horizons. A lower forecast error signifies superior performance, underscoring the models effectiveness on the MTSF task. The results for the baseline models are reported from a previous study by \cite{jiang2021dl}. The experimental study found that our proposed model, namely \textbf{MKH-Net}, outperformed the baseline models in the aforementioned evaluation metrics across different forecast horizons. Specifically, the results showed that \textbf{MKH-Net} model consistently achieved higher accuracy and lower error rates than the baseline models. On the PEMS-BAY dataset, the proposed model showed much lower forecast errors compared to the next-to-best baseline model. Specifically, the forecast errors for the \textbf{MKH-Net} model were 32.92$\%$, 22.45$\%$, and 4.16$\%$ lower than those of the next-to-best baseline model for forecasting 3-, 6-, and 12-steps-ahead, respectively, as measured by the MAPE metric. Similarly, on the METR-LA dataset, the proposed model also outperformed the next-best baseline models.  The \textbf{MKH-Net} model showed improvements of 45.16$\%$, 26.53$\%$, and 5.47$\%$ in terms of lower forecast errors over the next-best baseline model for forecasting 3-, 6-, and 12-steps-ahead, respectively, as measured by the MAPE metric.

\vspace{-3mm}

\vspace{-5mm}
\subsection{ABLATION STUDY}
The \textbf{MKH-Net} framework, serving as the baseline for our ablation study, seamlessly integrates spatial and temporal inference components to capture the intricate inter and intra-time-series correlations, thus effectively modeling the nonlinear dynamics of complex interconnected sensor networks. Further, the spatial inference component of the framework consists of three main methods, namely the ``explicit subgraph", the ``implicit hypergraph", and the ``dual-hypergraph" representation learning methods. We undertake an extensive ablation study to assess the influence of each learning component within the \textbf{MKH-Net} framework on the MTSF task. The impact of individual components on the overall framework performance can be observed by selectively removing or modifying them, providing valuable insight into their unique contributions towards the framework's effectiveness.
This in-depth analysis offers valuable insights and  identifies the critical components that contribute to better framework performance. We systematically eliminated these components to generate various ablated variants and scrutinized the impact of each component on the framework's overall performance for MTSF task by comparing the results against the baseline. 
This approach enabled a comprehensive understanding of each component's contribution to the framework's effectiveness on the MTSF task. Our ablation study enabled us to comprehend the relationship between various ablated variants and the baseline, contributing to an improved understanding of the various mechanisms underlying their generalization performance. We provide the detailed information on each ablated variant that was created by systematically removing specific components as follows,

\begin{itemize}
    \item ``\textbf{w/o Spatial}": A variant of \textbf{MKH-Net} framework that excluded the spatial inference component. This ablated variant performance demonstrates the importance of utilizing message-passing schemes for learning on predefined graphs, implicit hypergraph, and dual-hypergraph relational structures to better capture the inter-series correlations underlying the MTS data.
    \item ``\textbf{w/o Temporal}": A variant of \textbf{MKH-Net} that excluded the temporal inference component. This ablated variant verifies the effectiveness of incorporating temporal inference component to model the time-varying inter-series dependencies in complex systems.        
    \item ``\textbf{w/o Explicit Subgraph}": A variant of \textbf{MKH-Net} framework without the explicit subgraph representation learning method for modeling the spatio-temporal dynamics of complex interconnected systems. The results emphasized the critical role of learning on the subgraphs.     
    \item  ``\textbf{w/o Implicit Hypergraph}": A variant of \textbf{MKH-Net} framework without the implicit hypergraph representation learning method. This ablated variant reinforced the importance of learning the implicit hypergraph structure underlying MTS data and its optimal representations, highlighting the critical role of this module in the overall framework performance.
    \item  ``\textbf{w/o Dual Hypergraph}": A variant of \textbf{MKH-Net} framework without the dual-hypergraph representation learning method. This ablated variant signifies the importance of performing dual-hypergraph transformation to reveal hidden patterns and relationships in complex MTS data for better capturing the spatio-temporal dynamics within the MTS data.     
\end{itemize}

\vspace{-1mm}
Tables \ref{tab:ablation1}, \ref{tab:ablation2}, \ref{tab:ablation3} and \ref{tab:ablation4} presents the findings from ablation studies conducted on the benchmark datasets. We utilized different forecast accuracy measures—including Mean Absolute Error(MAE), Root Mean Squared Error(RMSE), and Mean Absolute Percentage Error(MAPE)—to ensure a holistic comprehension of the ablated variants performance vis-à-vis the baseline. For multistep-ahead forecasting, the accuracy of pointwise forecasts was compared to the observed data(ground-truth) over the prediction interval, and the results were expressed in terms of the aforementioned forecast accuracy metrics. To provide further clarification, we included the relative percentage difference between the ablated variant and the baseline performance in parentheses. We ensured the accuracy of our findings by conducting multiple(quintuple) experiments and reporting the average results. Additionally, we evaluated the ablated variants ability to handle long-term predictions compared to the baseline by setting the forecast horizon to 12. Tables \ref{tab:ablation1}, \ref{tab:ablation2}, \ref{tab:ablation3} and \ref{tab:ablation4} clearly shows that the variant models exhibit reduced forecast accuracy and significantly underperform compared to the baseline. Upon examination, it becomes evident that the spatial inference component within the \textbf{MKH-Net} framework holds greater significance than the temporal inference component for attaining state-of-the-art performance on the benchmark datasets. For the PeMSD8 dataset, the ``w/o Spatial" variant shows a substantial decline in performance relative to the baseline, evidenced by a marked increase of 8.65$\%$ in RMSE, 10.93$\%$ in MAE, and 24.16$\%$ in MAPE. In contrast, the ``w/o Temporal" variant exhibits a marginally inferior performance compared to the baseline, with a modest increase of 6.26$\%$ in RMSE, 8.36$\%$ in MAE, and 11.34$\%$ in MAPE. Analogously, we see similar patterns on the PeMSD4 dataset. The ``w/o Spatial" variant, perform much worse than the benchmark, with an increase of 11.64$\%$ in RMSE, 15.23$\%$ in MAE, and 5.78$\%$ in MAPE. Conversely, the ``w/o Temporal" variant demonstrates a slight decrement in performance compared to the baseline, with a negligible increase of 5.93$\%$ in RMSE, 7.27$\%$ in MAE, and 6.02$\%$ in MAPE. A higher increase in the ablated variants error metrics, juxtaposed against the baseline, substantiates the relative significance of the mechanisms underpinning the omitted components of the baseline. In conclusion, the predominant backbone contributing to the improved performance of the \textbf{MKH-Net} framework on multi-horizon forecasting is the spatial inference component, which is responsible for capturing the intricate interdependencies among multiple time series variables and learning the dynamics of interacting systems. The importance of the spatial inference component is demonstrated by the significant drop in performance when it is excluded compared to the baseline, highlighting its indispensable nature. The spatial inference component is composed of three essential methods, namely the ``explicit graph", ``implicit hypergraph", and ``dual hypergraph" representation learning methods, which form the foundation of our neural forecast architecture. The ``\textbf{w/o Explicit Subgraph}" variant performed worse than the baseline, showing an increase of 2.22$\%$, 1.23$\%$, and 5.97$\%$ on PeMSD4, and 5.14$\%$, 4.26$\%$, and 5.76$\%$ on PeMSD8, with respect to the RMSE, MAE, and MAPE metrics, respectively. The results support the rationale of learning on the subgraph structures to achieve better performance in multi-horizon forecasting tasks. The ``\text{w/o Implicit HyperGraph}" variant performance drops compared to the baseline with a marginal increase of 2.32$\%$, 3.28$\%$, and 1.02$\%$ on PeMSD4; 4.39$\%$, 5.17$\%$, 3.80$\%$ on PeMSD8 w.r.t. RMSE, MAE, and MAPE metrics, respectively. Incorporating the hypergraph inference and representation learning method into the framework was found to be crucial, as the results demonstrated a notable improvement in forecast accuracy. The variant ``\text{w/o Dual HyperGraph}" performed worse than the baseline on several metrics, showing an increase of up to 2.49$\%$, 3.45$\%$, and 1.72$\%$ on PeMSD4; 3.91$\%$, 7.48$\%$, 2.19$\%$ on PeMSD8 w.r.t. RMSE, MAE, and MAPE metrics, respectively. The results supports learning on dual-hypergraph transformation that help the framework learn more expressive representations, leading to better performance in multi-horizon forecasting tasks. Tables \ref{tab:ablation1}, \ref{tab:ablation2}, \ref{tab:ablation3} and \ref{tab:ablation4} shows additional results from the ablation study on benchmark datasets. The results demonstrate that the proposed \textbf{MKH-Net} framework generalizes well, even with complex patterns across a broad spectrum of datasets, and scales well on large-scale graph datasets. Overall, the ablation studies support the hypothesis of joint optimization of spatial-temporal learning components to achieve improved performance in multi-horizon time series forecasting tasks.

\begin{table*}[!ht]
\centering
\renewcommand{\arraystretch}{1.2}
\resizebox{1.05\textwidth}{!}{
\begin{tabular}{c|c|ccc|c|ccc}
\hline
\textbf{Method} & \multirow{7}{*}{\textbf{\rotatebox[origin=c]{90}{PeMSD3}}} & \textbf{RMSE} & \textbf{MAE} & \textbf{MAPE} & \multirow{7}{*}{\textbf{\rotatebox[origin=c]{90}{PeMSD4}}} & \textbf{RMSE} & \textbf{MAE} & \textbf{MAPE}  \\ \cline{1-1} \cline{3-5} \cline{7-9}
\textbf{MKH-Net} &  & \textbf{19.937} & \textbf{13.177} & \textbf{11.679} &  & \textbf{28.925} & \textbf{19.194} & \textbf{12.259}  \\ \cline{1-1} \cline{3-5} \cline{7-9} 
``\textbf{w/o Spatial}"  &  & 22.840$(\color{black}14.56\%\uparrow)$ & 14.813$(\color{black}12.42\%\uparrow)$ & 12.491$(\color{black}6.95\%\uparrow)$ &  & 32.293$(\color{black}11.64\%\uparrow)$ & 22.117$(\color{black}15.23\%\uparrow)$ & 12.967$(\color{black}5.78\%\uparrow)$ \\ 
``\textbf{w/o Temporal}" &  & 21.813$(\color{black}9.40\%\uparrow)$ & 13.958$(\color{black}5.93\%\uparrow)$ & 11.739$(\color{black}0.51\%\uparrow)$ &  & 30.641$(\color{black}5.93\%\uparrow)$ & 20.589$(\color{black}7.27\%\uparrow)$ & 12.997$(\color{black}6.02\%\uparrow)$ \\
``\textbf{w/o Explicit Subgraph}" &  & 20.502$(\color{black}2.83\%\uparrow)$ & 13.742$(\color{black}4.29\%\uparrow)$ & 11.979$(\color{black}2.57\%\uparrow)$ &  & 29.568$(\color{black}2.22\%\uparrow)$ & 19.430$(\color{black}1.23\%\uparrow)$ & 12.991$(\color{black}5.97\%\uparrow)$ \\
``\textbf{w/o Dual Hypergraph}" &  & 20.383$(\color{black}2.24\%\uparrow)$ & 13.190$(\color{black}0.10\%\uparrow)$ & 11.890$(\color{black}1.80\%\uparrow)$ &  & 29.646$(\color{black}2.49\%\uparrow)$ & 19.857$(\color{black}3.45\%\uparrow)$ & 12.470$(\color{black}1.72\%\uparrow)$ \\
``\textbf{w/o Implicit Hypergraph}" &  & 20.190$(\color{black}1.27\%\uparrow)$ & 13.160$(\color{black}0.13\%\downarrow)$ & 11.741$(\color{black}0.53\%\uparrow)$ &  & 29.596$(\color{black}2.32\%\uparrow)$ & 19.824$(\color{black}3.28\%\uparrow)$ & 12.384$(\color{black}1.02\%\uparrow)$ \\ \hline
\end{tabular}
}
\vspace{-2mm}
\caption{The table presents the results of an ablation study on multi-horizon forecasting using the PeMSD3 and PeMSD4 benchmark datasets.}
\label{tab:ablation1}
\end{table*}

\begin{table*}[!ht]
\centering
\renewcommand{\arraystretch}{1.2}
\resizebox{1.05\textwidth}{!}{
\begin{tabular}{c|c|ccc|c|ccc}
\hline
\textbf{Method} & \multirow{7}{*}{\textbf{\rotatebox[origin=c]{90}{PeMSD7}}} & \textbf{RMSE} & \textbf{MAE} & \textbf{MAPE} & \multirow{7}{*}{\textbf{\rotatebox[origin=c]{90}{PeMSD8}}} & \textbf{RMSE} & \textbf{MAE} & \textbf{MAPE}  \\ \cline{1-1} \cline{3-5} \cline{7-9} 
\textbf{MKH-Net} &  & \textbf{35.087} & \textbf{21.984} & \textbf{9.368} &  & \textbf{21.241} & \textbf{13.605} & \textbf{7.762} \\ \cline{1-1} \cline{3-5} \cline{7-9}
``\textbf{w/o Spatial}" &  &  37.684$(\color{black}7.40\%\uparrow)$ & 23.189$(\color{black}5.48\%\uparrow)$ & 10.186$(\color{black}8.73\%\uparrow)$ &  & 23.079$(\color{black}8.65\%\uparrow)$ & 15.092$(\color{black}10.93\%\uparrow)$ & 9.637$(\color{black}24.16{\%\uparrow})$ \\
``\textbf{w/o Temporal}" &  & 36.898$(\color{black}5.16\%\uparrow)$ & 22.637$(\color{black}2.97\%\uparrow)$ & 9.897$(\color{black}5.65\%\uparrow)$ &  & 22.570$(\color{black}6.26\%\uparrow)$ & 14.742$(\color{black}8.36\%\uparrow)$ & 8.642$(\color{black}11.34\%\uparrow)$ \\
``\textbf{w/o Explicit Subgraph}"&  &  35.547$(\color{black}1.31{\%\uparrow})$ & 22.243$(\color{black}1.18{\%\uparrow})$ & 9.739$(\color{black}3.96{\%\uparrow})$ &  & 22.332$(\color{black}5.14{\%\uparrow})$ & 14.185$(\color{black}4.26{\%\uparrow})$ & 8.209$(\color{black}5.76\%\uparrow)$ \\
``\textbf{w/o Dual Hypergraph}" &  & 35.769$(\color{black}1.94{\%\uparrow})$ & 22.135$(\color{black}0.69{\%\uparrow})$ & 9.875$(\color{black}{5.41\%\uparrow})$ &  & 22.072$(\color{black}3.91{\%\uparrow})$ & 14.622$(\color{black}7.48{\%\uparrow})$ & 7.932$(\color{black}2.19{\%\uparrow})$ \\
``\textbf{w/o Implicit Hypergraph}" &  & 35.284$(\color{black}0.56{\%\uparrow})$ & 22.208$(\color{black}1.02{\%\uparrow})$ & 9.551$(\color{black}1.95{\%\uparrow})$ &  & 22.174$(\color{black}4.39\%\uparrow)$ & 14.309$(\color{black}5.17{\%\uparrow})$ & 8.057$(\color{black}3.80{\%\uparrow})$ \\ \hline
\end{tabular}
}
\vspace{-2mm}
\caption{The table presents the results of an ablation study on multi-horizon forecasting using the PeMSD7 and PeMSD8 benchmark datasets.}
\label{tab:ablation2}
\end{table*}

\vspace{-2mm}
\begin{table*}[!ht]
\centering
\renewcommand{\arraystretch}{1.2}
\resizebox{1.05\textwidth}{!}{
\begin{tabular}{c|c|ccc|c|ccc}
\hline
\textbf{Method} & \multirow{7}{*}{\textbf{\rotatebox[origin=c]{90}{PeMSD7(M)}}} & \textbf{RMSE} & \textbf{MAE} & \textbf{MAPE} & \multirow{7}{*}{\textbf{\rotatebox[origin=c]{90}{METR-LA}}} & \textbf{RMSE} & \textbf{MAE} & \textbf{MAPE}\\ \cline{1-1} \cline{3-5} \cline{7-9} 
\textbf{MKH-Net} &  & \textbf{6.162} & \textbf{3.428} & \textbf{7.097} &  & \textbf{7.495} & \textbf{4.594} & \textbf{8.578}  \\ \cline{1-1} \cline{3-5} \cline{7-9}
``\textbf{w/o Spatial}" &  & 7.887$(\color{black}27.99\%\uparrow)$ & 3.998$(\color{black}16.63{\%\uparrow})$ & 8.134$(\color{black}14.61{\%\uparrow})$ &  & 8.646$(\color{black}15.36{\%\uparrow})$ & 5.190$(\color{black}12.03{\%\uparrow})$ & 9.473$(\color{black}10.43{\%\uparrow})$ \\
``\textbf{w/o Temporal}" &  & 7.044$(\color{black}14.31{\%\uparrow})$ & 3.647$(\color{black}6.39{\%\uparrow})$ & 7.964$(\color{black}12.22{\%\uparrow})$ &  & 8.040$(\color{black}7.27{\%\uparrow})$ & 4.913$(\color{black}6.94{\%\uparrow})$ & 8.793$(\color{black}2.51{\%\uparrow})$ \\
``\textbf{w/o Explicit Subgraph}" &  & 6.736$(\color{black}9.32{\%\uparrow})$ & 3.541$(\color{black}3.30{\%\uparrow})$ & 7.424$(\color{black}4.61{\%\uparrow})$ &  & 7.928$(\color{black}5.78{\%\uparrow})$ & 4.972$(\color{black}8.23{\%\uparrow})$ & 8.827$(\color{black}{2.90\%\uparrow})$ \\
``\textbf{w/o Dual Hypergraph}" &  & 6.891$(\color{black}11.83{\%\uparrow})$ & 3.692$(\color{black}7.70{\%\uparrow})$ & 7.615$(\color{black}7.30{\%\uparrow})$ &  & 7.694$(\color{black}2.66{\%\uparrow})$ & 4.704$(\color{black}2.40{\%\uparrow})$ & 8.616$(\color{black}{0.44\%\uparrow})$ \\
``\textbf{w/o Implicit Hypergraph}" &  & 6.576$(\color{black}6.72{\%\uparrow})$ & 3.615$(\color{black}5.46{\%\uparrow})$ & 7.459$(\color{black}5.10{\%\uparrow})$ &  & 7.782$(\color{black}3.83{\%\uparrow})$ & 4.689$(\color{black}2.07{\%\uparrow})$ & 8.696$(\color{black}1.37{\%\uparrow})$\\ \hline
\end{tabular}
}
\vspace{-2mm}
\caption{The table presents the results of an ablation study on multi-horizon forecasting using the PeMSD7(M), METR-LA benchmark datasets.}
\label{tab:ablation3}
\end{table*}

\vspace{-2mm}
\begin{table*}[!ht]
\centering
\renewcommand{\arraystretch}{1.2}
\resizebox{0.6\textwidth}{!}{
\begin{tabular}{c|c|ccc}
\hline
\textbf{Method} & \multirow{7}{*}{\textbf{\rotatebox[origin=c]{90}{PeMSD7(M)}}} & \textbf{RMSE} & \textbf{MAE} & \textbf{MAPE}  \\ \cline{1-1} \cline{3-5}
\textbf{MKH-Net} &  & \textbf{3.042} & \textbf{1.638} & \textbf{3.035} \\ \cline{1-1} \cline{3-5}
``\textbf{w/o Spatial}" &  & 3.539$(\color{black}16.34{\%\uparrow})$ & 1.779$(\color{black}8.61{\%\uparrow})$ & 3.963$(\color{black}30.58{\%\uparrow})$ \\
``\textbf{w/o Temporal}" &  &  3.266$(\color{black}7.36{\%\uparrow})$ & 1.685$(\color{black}2.43{\%\uparrow})$ & 3.541$(\color{black}16.67{\%\uparrow})$ \\
``\textbf{w/o Explicit Subgraph}" &  & 3.193$(\color{black}4.96{\%\uparrow})$ & 1.697$(\color{black}3.60{\%\uparrow})$ & 3.297$(\color{black}8.63{7\%\uparrow})$ \\
``\textbf{w/o Dual Hypergraph}" &  & 3.134$(\color{black}3.02{\%\uparrow})$ & 1.686$(\color{black}2.93{\%\uparrow})$ & 3.189$(\color{black}5.07{\%\uparrow})$ \\
``\textbf{w/o Implicit Hypergraph}" &  & 3.114$(\color{black}2.37{\%\uparrow})$ & 1.671$(\color{black}2.01{\%\uparrow})$ & 3.195$(\color{black}5.27{\%\uparrow})$ \\ \hline
\end{tabular}
}
\vspace{-2mm}
\caption{The table presents the results of an ablation study on multi-horizon forecasting using the PeMS-BAY benchmark dataset.}
\label{tab:ablation4}
\end{table*}

\vspace{3mm}
\subsection{POINTWISE PREDICTION ERROR FOR MULTI-HORIZON FORECASTING}
\begin{figure*}[!ht]
\centering
\subfloat[MAE on PeMSD3]{
 \includegraphics[width=45mm]{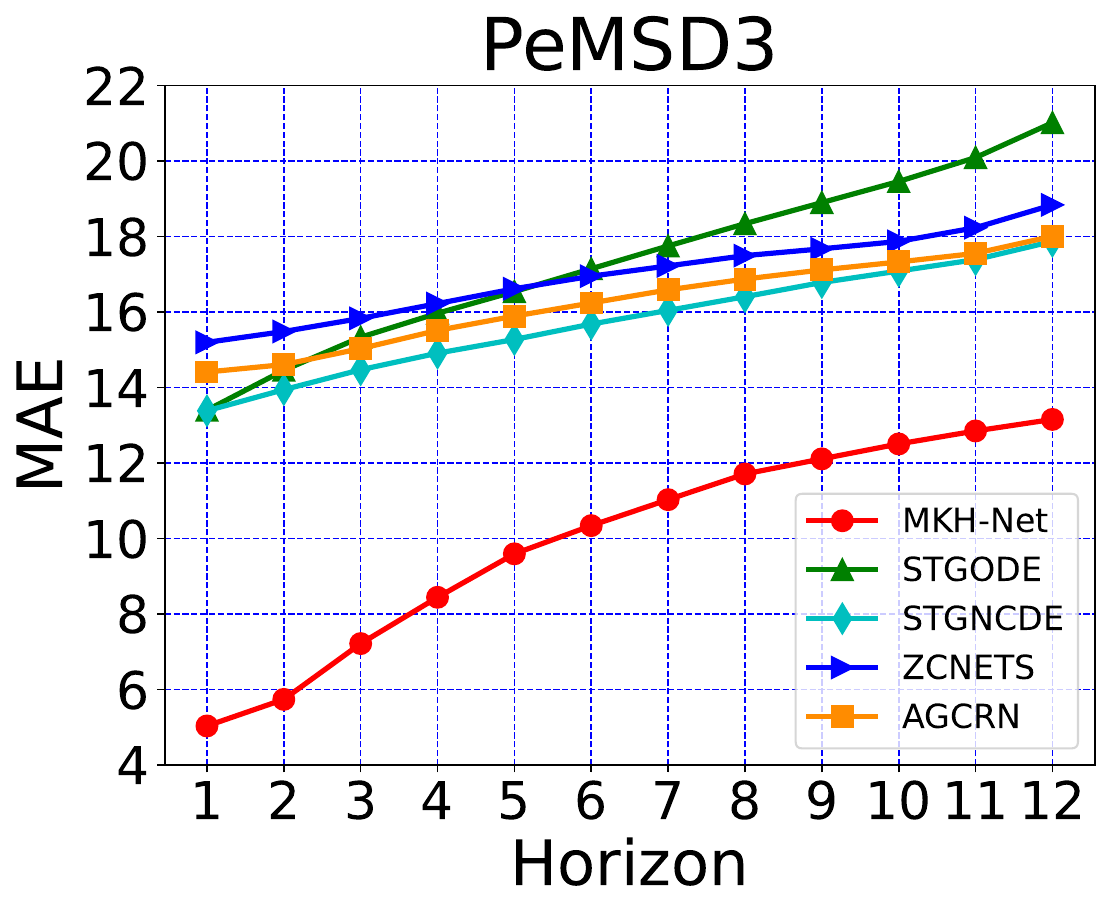} 
}
\subfloat[MAPE on PeMSD3]{
 \includegraphics[width=45mm]{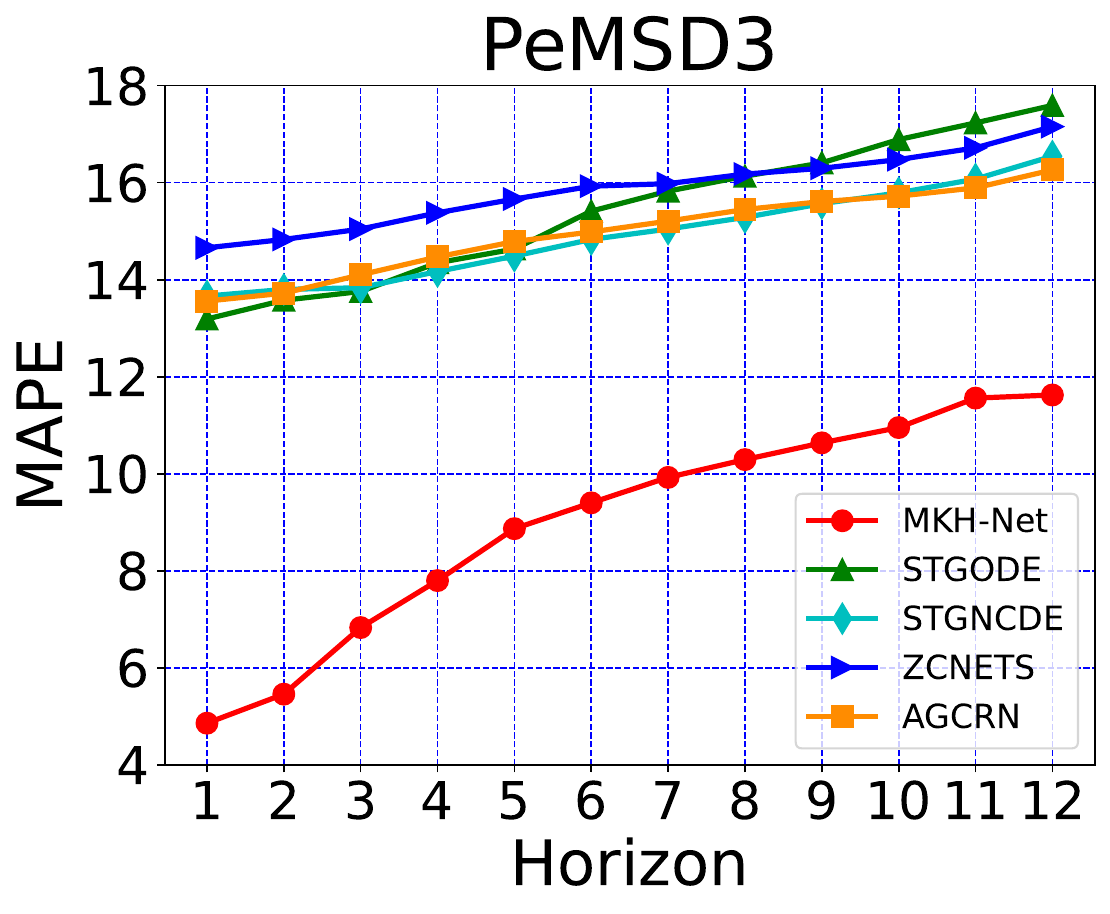}
}
\subfloat[RMSE on PeMSD3]{
 \includegraphics[width=45mm]{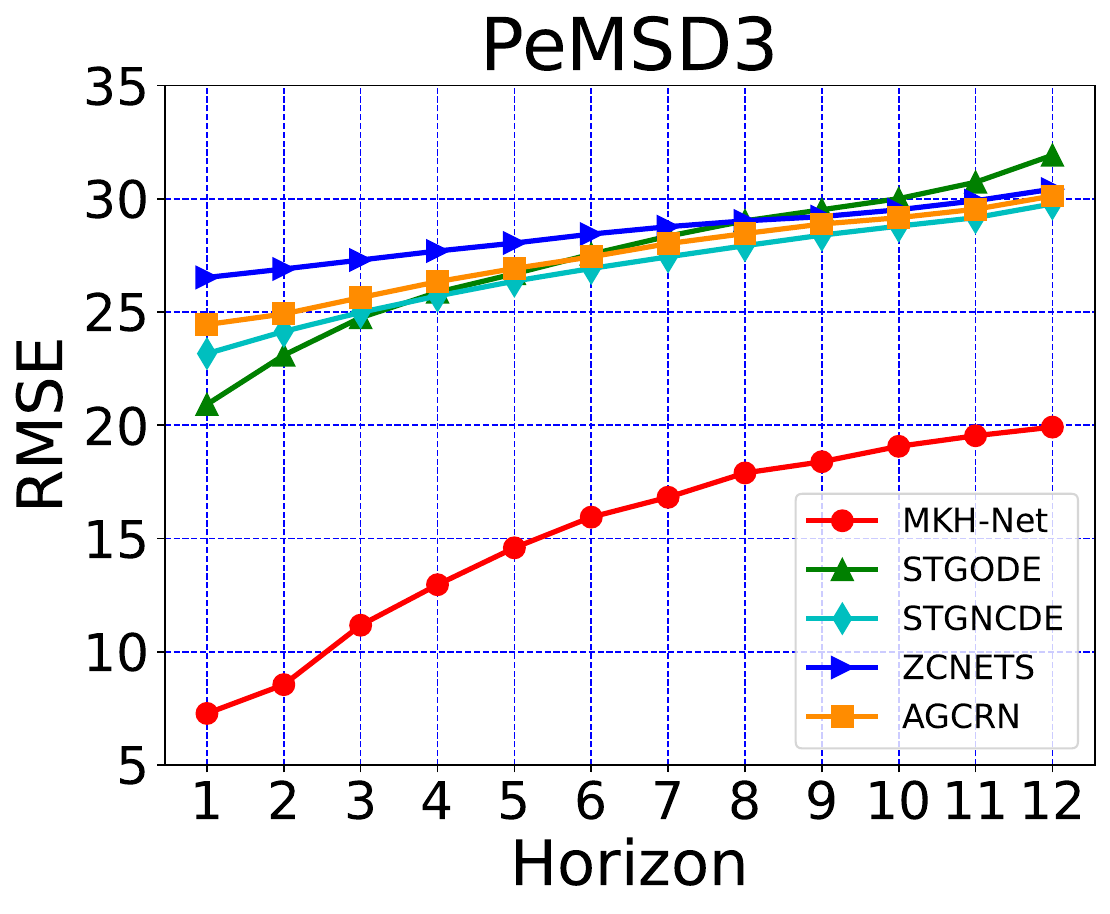}
}
\subfloat[MAE on PeMSD4]{
 \includegraphics[width=45mm]{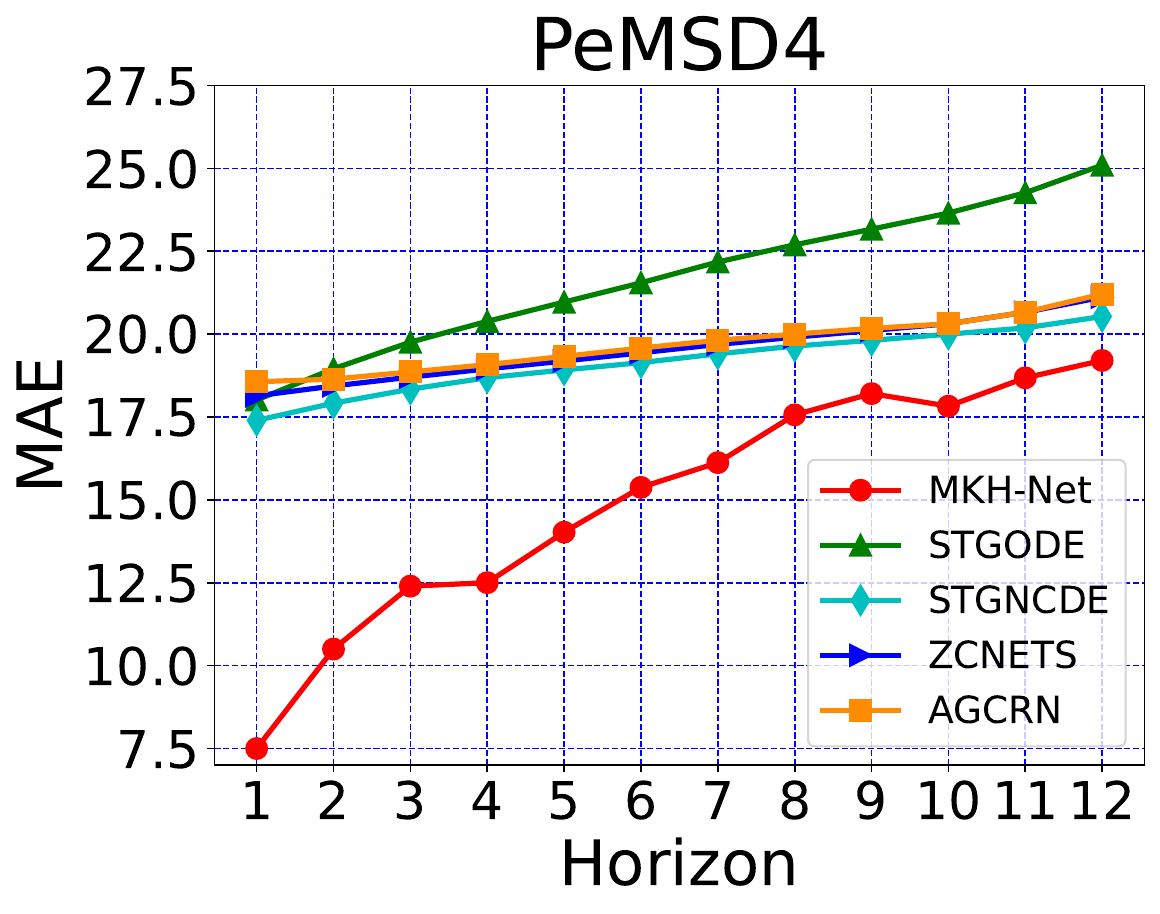}
}
\hspace{-2mm}
\subfloat[MAPE on PeMSD4]{
 \includegraphics[width=45mm]{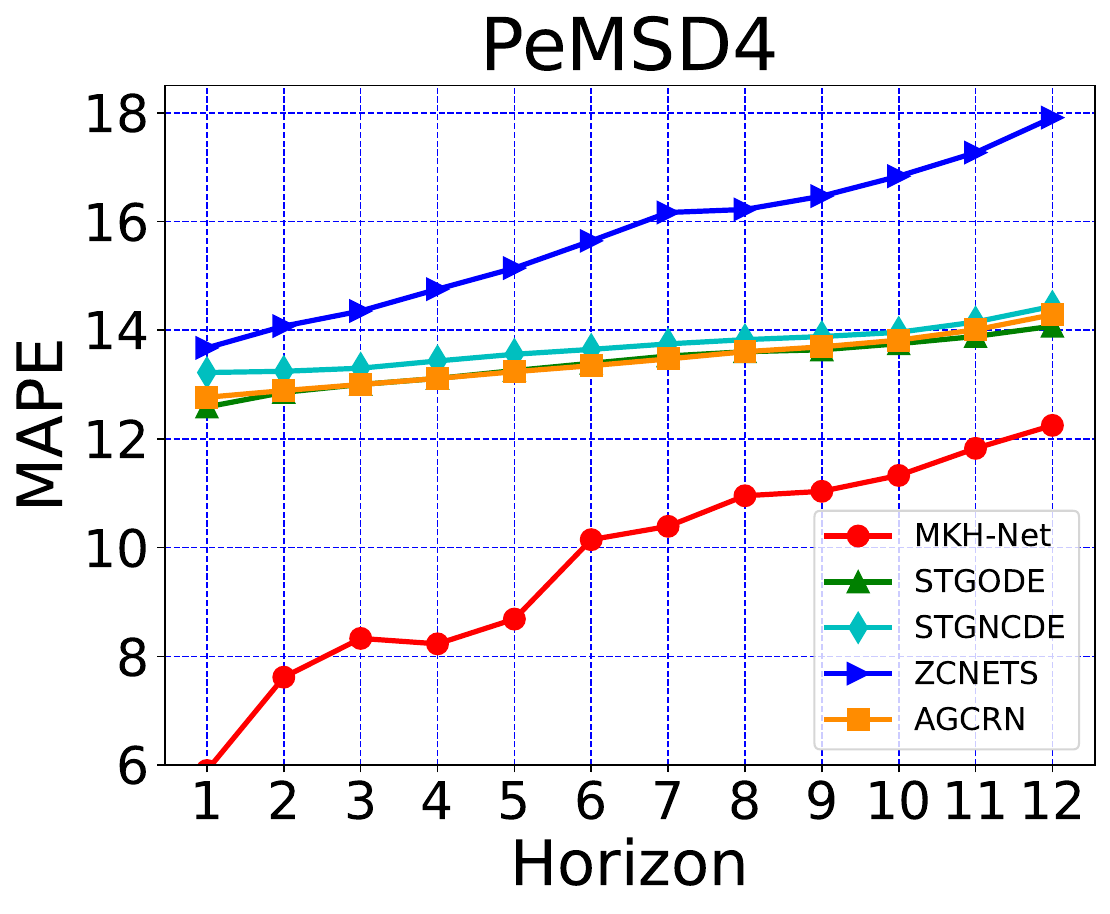}
}
\subfloat[RMSE on PeMSD4]{
 \includegraphics[width=45mm]{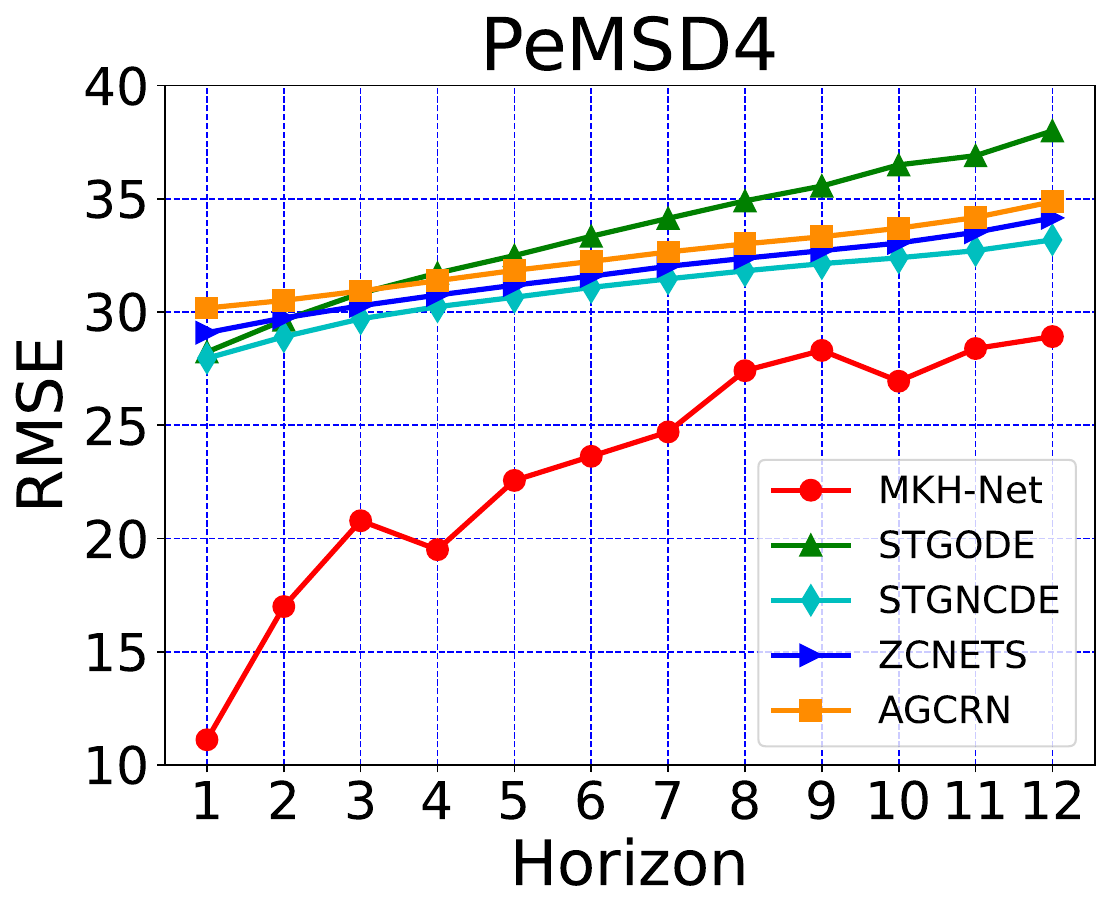}
}
\subfloat[RMSE on PeMSD7]{
 \includegraphics[width=45mm]{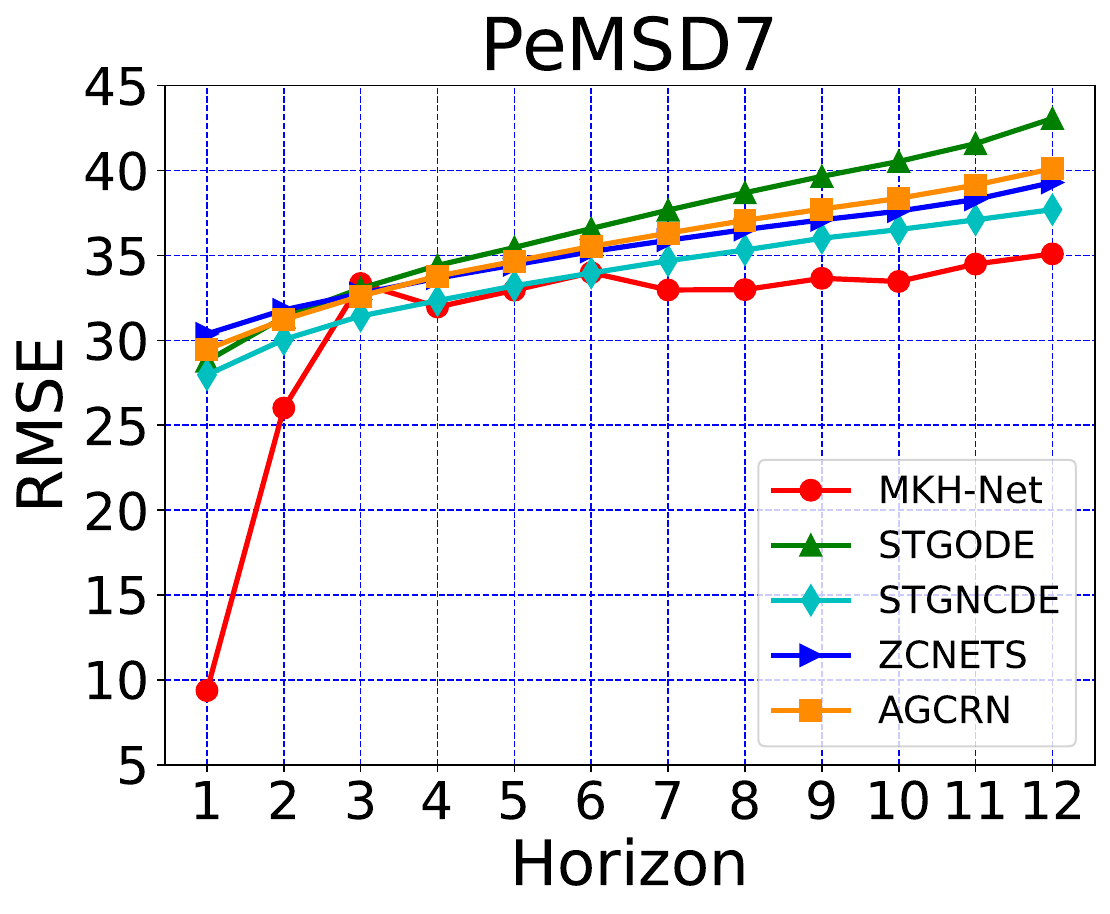}
}
\subfloat[MAE on PeMSD7]{
 \includegraphics[width=45mm]{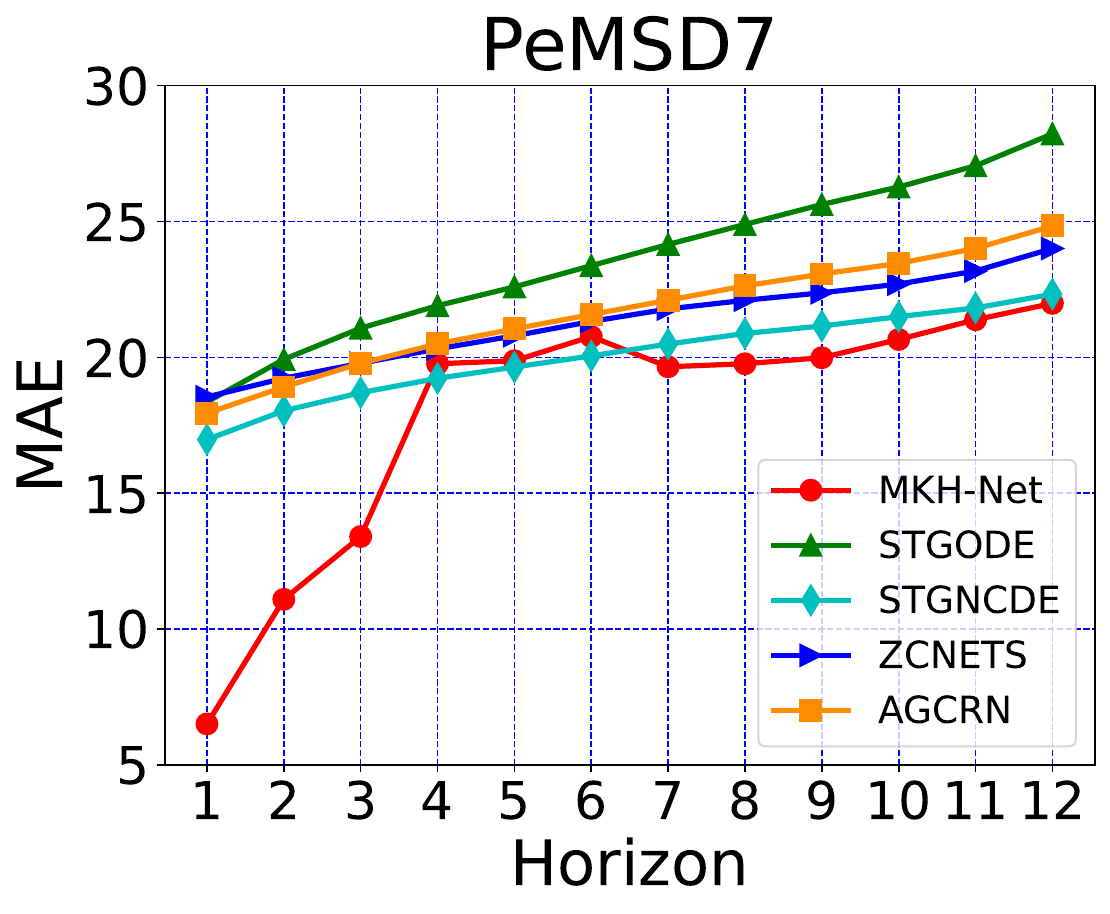}
}
\hspace{-2mm}
\subfloat[MAPE on PeMSD7]{
 \includegraphics[width=45mm]{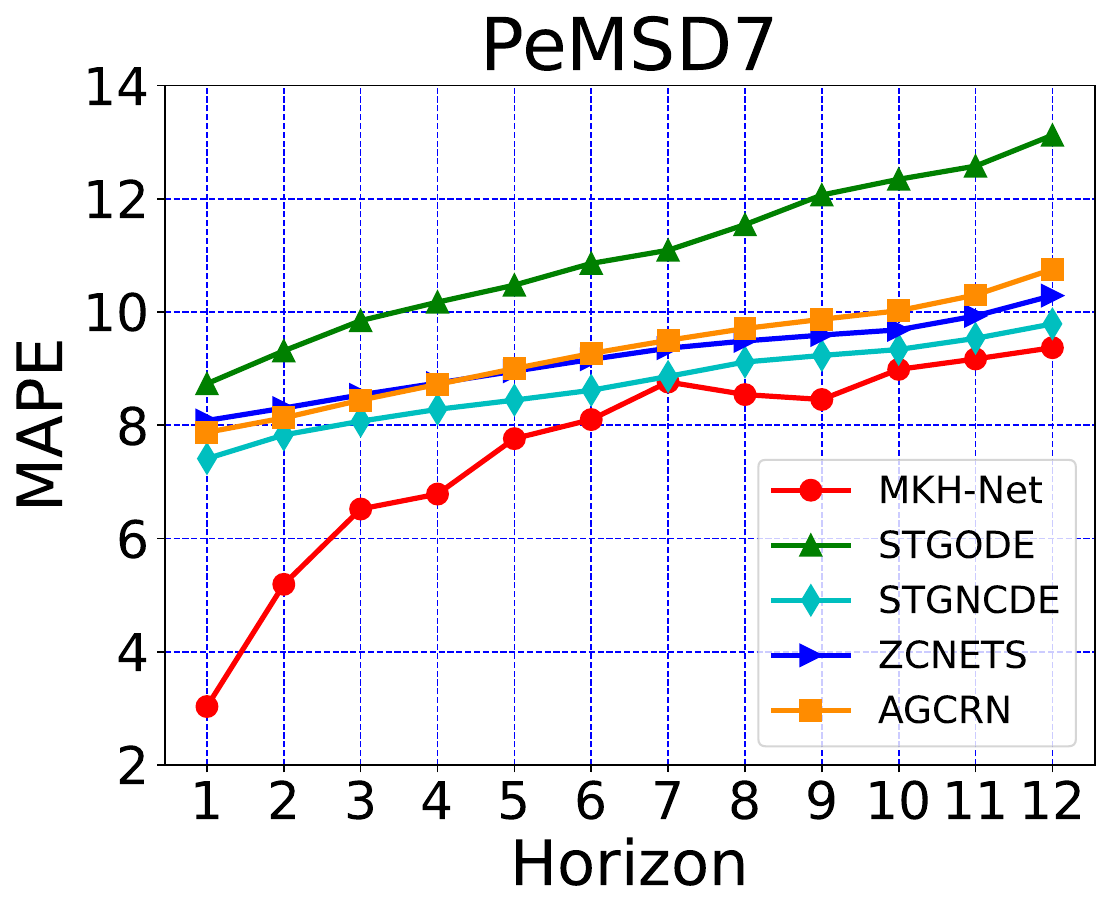}
}
\subfloat[MAE on PeMSD8]{
 \includegraphics[width=45mm]{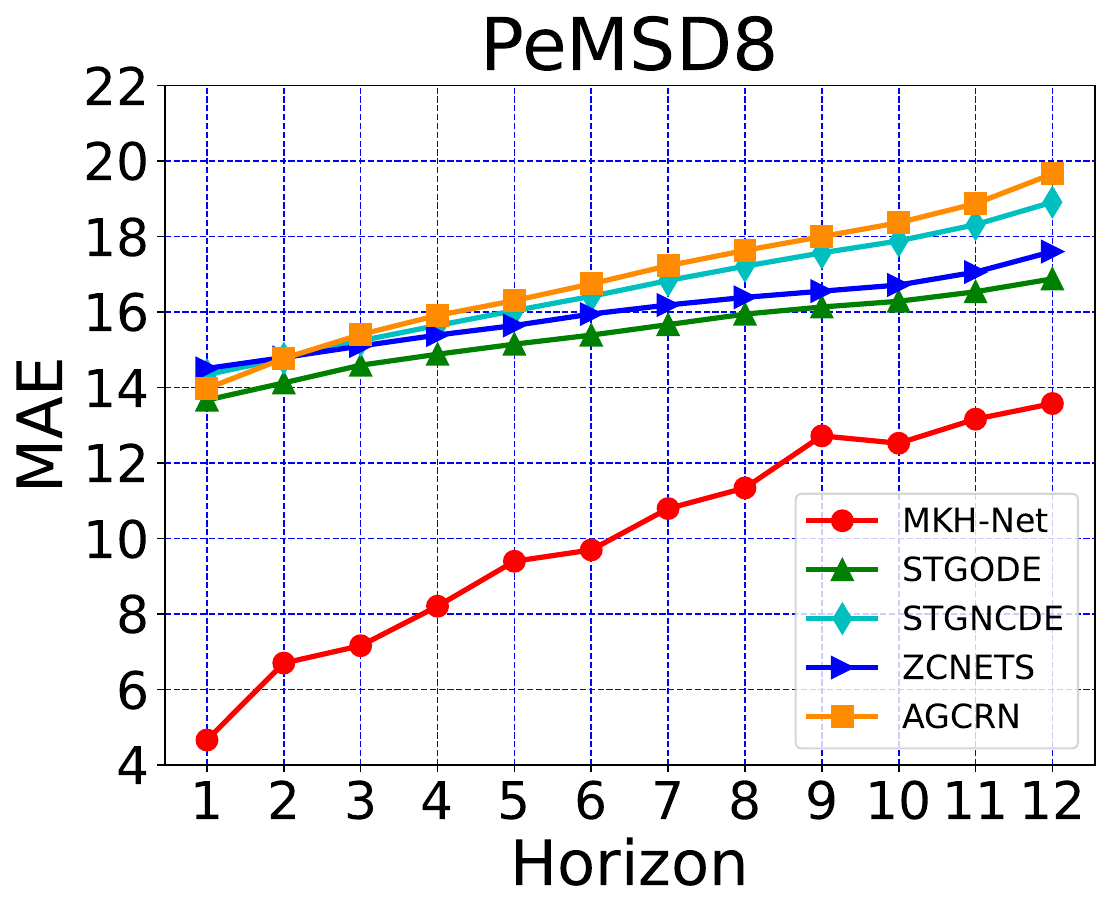}
}
\subfloat[MAPE on PeMSD8]{
 \includegraphics[width=45mm]{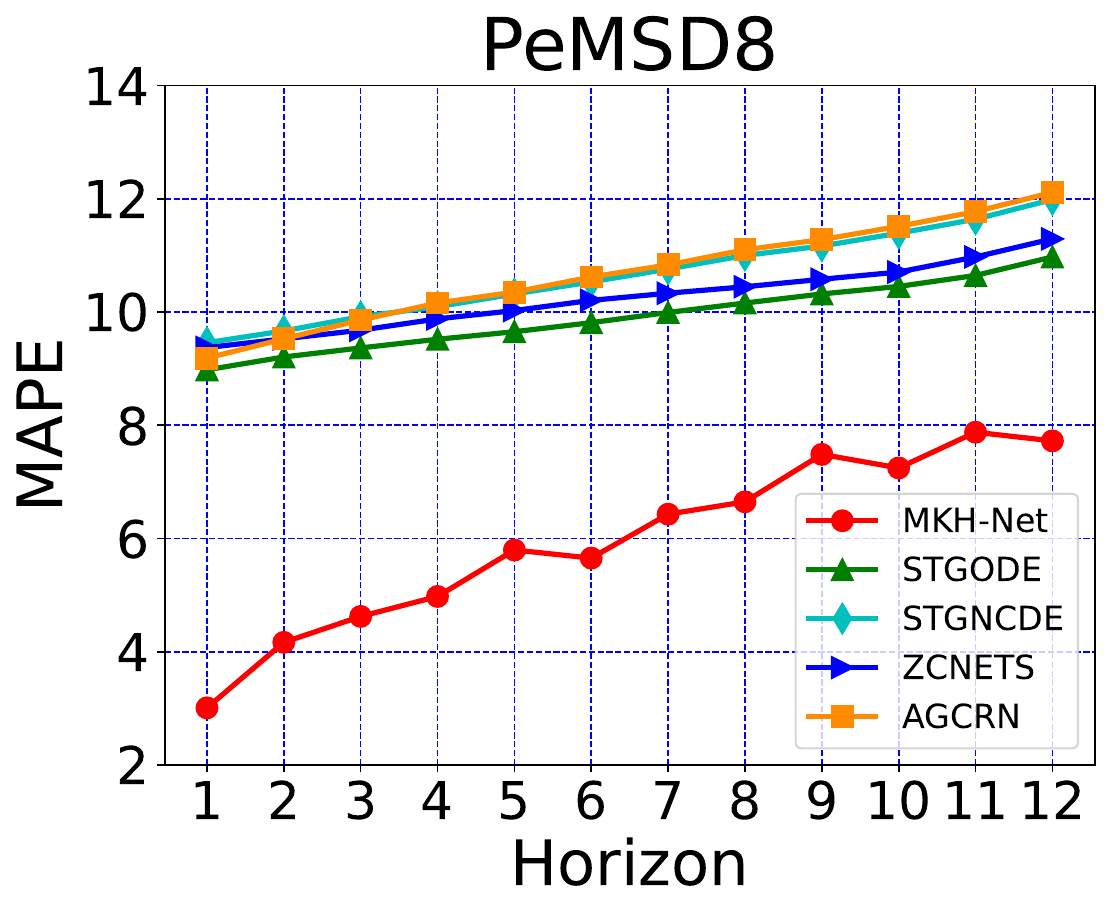}
}
\subfloat[RMSE on PeMSD8]{
 \includegraphics[width=45mm]{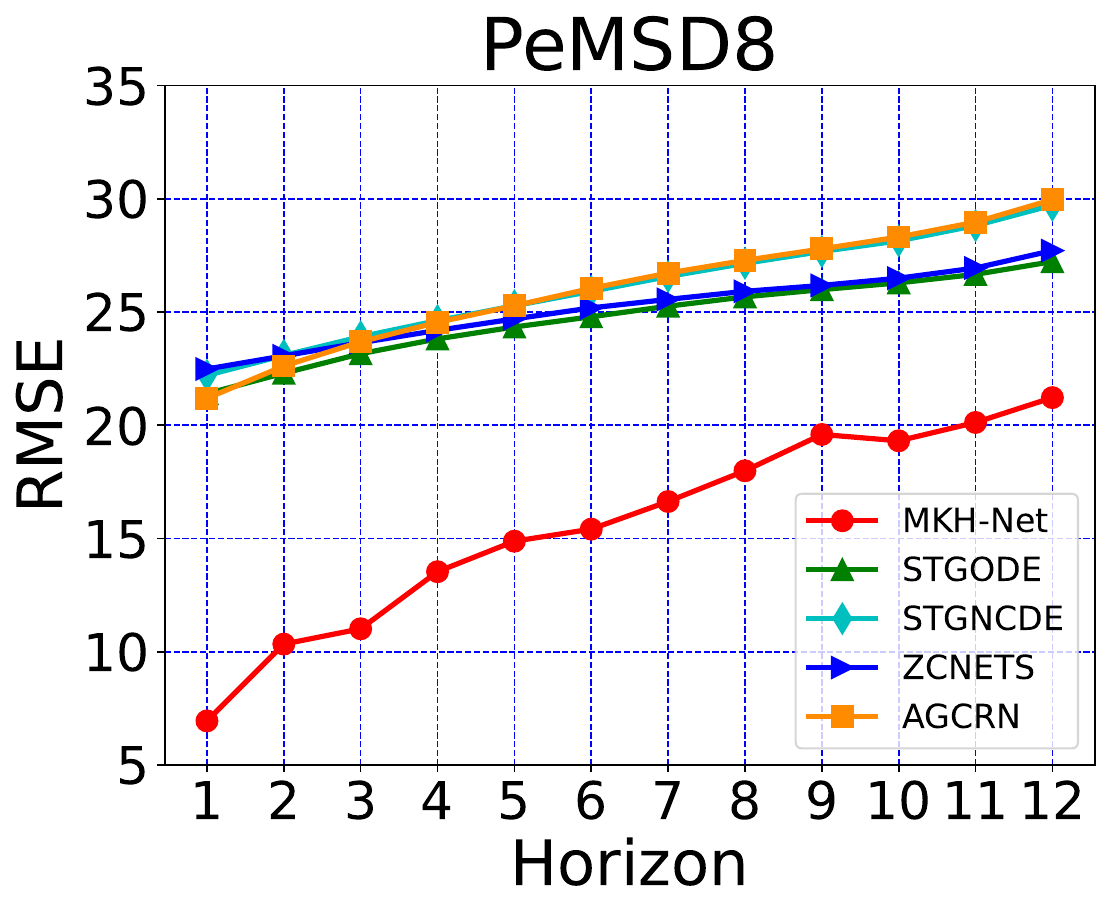}
}
\caption{The figure provides a comprehensive overview of the pointwise prediction errors for multi-horizon forecasting tasks, presented in relation to multiple evaluation metrics.
It reports the results of evaluations conducted on benchmark datasets.}
\label{fig-prediction}
\end{figure*}

\vspace{1mm}
The proposed neural forecasting framework(\textbf{MKH-Net}), has been evaluated for its ability to generate accurate multistep-ahead forecasts on multiple benchmark datasets. The \textbf{MKH-Net} framework performance is evaluated using metrics such as RMSE, MAPE, and MAE, with lower values indicating better model performance. Figure \ref{fig-prediction} shows the multistep-ahead forecast errors of the \textbf{MKH-Net} framework on benchmark datasets. The results show that the \textbf{MKH-Net} outperforms the baselines across all prediction horizons, suggesting that the framework can effectively capture nonlinear spatio-temporal dependencies in structured MTS data to improve forecast accuracy by exploiting relational inductive biases.

\subsection{IRREGULAR TIME SERIES FORECASTING}
Large, complex sensor networks found in various real-world applications often suffer from low-quality data due to intermittent sensor failures and faulty sensors during data acquisition. To evaluate the effectiveness of the \textbf{MKH-Net} framework in handling missing data, we simulate two types of missingness patterns - point-missing and block-missing patterns - which mimic the missingness patterns observed in real-world data of large, complex sensor networks. In the point-missing pattern, observations of each variable are randomly dropped within a historical window, with a missing ratios of 10$\%$, 30$\%$ and 50$\%$. Similarly, in the block-missing pattern, the available data for each variable is randomly masked within a historical window, where missing ratios can range from 10$\%$ to 50$\%$. Additionally, a sensor failure is simulated with a probability of 0.15$\%$, resulting in blocks of missing data for the multivariate time series data. The experimental studies demonstrate the robustness and reliability of the \textbf{MKH-Net} framework in handling missing data, which is ubiquitous in real-world applications. We split several benchmark datasets into three subsets - training, validation, and testing, based on their chronological order. The METR-LA and PEMS-BAY datasets were split in a ratio of 7:1:2, while the other datasets were split in a ratio of 6:2:2. The \textbf{MKH-Net} framework performance is evaluated using multiple forecasting metrics to understand its ability to handle MTS data with missingness and to analyze the impact of increasing missing data percentage on the framework performance on the MTSF task. The \textbf{MKH-Net} framework, trained on fully observed data, is selected as the benchmark for the MTSF task. Table \ref{tab-missing1},\ref{tab-missing2} report the irregular-time-series forecasting results on the benchmark datasets. The \textbf{MKH-Net} framework performance degrades slightly compared to the benchmark when there is a lower percentage of missing data. With a further increase in the percentage of missing data, the framework performance deteriorates further, resulting in reduced forecast accuracy across all datasets, regardless of the missing data pattern. The proposed framework demonstrates robustness to missing data by utilizing observed data instead of relying on imputed values for model predictions. Moreover, the framework generates more reliable out-of-sample forecasts by capturing the complex dependencies and patterns within multivariate time series data, present in interconnected networks, resulting in improved forecast accuracy.

\begin{table*}[htbp]
\centering
\renewcommand{\arraystretch}{1.2}
\resizebox{\textwidth}{!}{
\begin{tabular}{c|c|c|ccc|c|ccc|c|ccc|c|cll}
\hline
\textbf{\begin{tabular}[c]{@{}c@{}}Missing \\ Scheme\end{tabular}} & \textbf{\begin{tabular}[c]{@{}c@{}}Missing \\ Rate\end{tabular}} & \multirow{8}{*}{\textbf{\rotatebox[origin=c]{90}{PeMSD3}}} & \textbf{RMSE} & \textbf{MAE} & \textbf{MAPE} & \multirow{8}{*}{\textbf{\rotatebox[origin=c]{90}{PeMSD4}}} & \textbf{RMSE} & \textbf{MAE} & \textbf{MAPE} & \multirow{8}{*}{\textbf{\rotatebox[origin=c]{90}{PeMSD7}}} & \textbf{RMSE} & \textbf{MAE} & \textbf{MAPE} & \multirow{8}{*}{\textbf{\rotatebox[origin=c]{90}{PeMSD8}}} & \textbf{RMSE} & \multicolumn{1}{c}{\textbf{MAE}} & \multicolumn{1}{c}{\textbf{MAPE}} \\ \cline{1-2} \cline{4-6} \cline{8-10} \cline{12-14} \cline{16-18} 
\textbf{MKH-Net} & \textbf{0\%} &  & \textbf{19.937} & \textbf{13.177} & \textbf{11.679} &  & \textbf{28.925} & \textbf{19.194} & \textbf{12.259} &  & 35.092 & \textbf{21.988} & \textbf{9.371} &  & \textbf{21.224} & \textbf{13.577} & \textbf{7.725} \\ \cline{1-2} \cline{4-6} \cline{8-10} \cline{12-14} \cline{16-18}
\multirow{3}{*}{Point} & 10\% &  & 22.411 & 15.146 & 13.266 &  & 31.511 & 21.836 & 14.000 &  & 35.252 & 22.954 & 9.687 &  & 24.844 & 16.759 & 9.582 \\
& 30\% &  & 23.737 & 16.082 & 13.885 &  & 33.609 & 23.651 & 14.808 &  & 35.377 & 25.054 & 10.325 &  & 26.319 & 18.018 & 10.299 \\
& 50\% &  & 25.168 & 17.053 & 14.499 &  & 35.910 & 25.211 & 15.719 &  & 37.486 & 25.681 & 11.027 &  & 29.180 & 19.929 & 11.434 \\ \cline{1-2} \cline{4-6} \cline{8-10} \cline{12-14} \cline{16-18} 
\multirow{3}{*}{Block} & 10\% &  & 22.656 & 15.198 & 13.155 &  & 32.316 & 22.148 & 14.135 &  & 38.923 & 25.534 & 11.003 &  & 25.246 & 16.931 & 9.696 \\
& 30\% &  & 23.646 & 15.896 & 13.594 &  & 34.024 & 23.784 & 14.836 &  &  39.042 & 25.638 & 11.528 & & 27.274 & 19.700 & 11.002 \\
& 50\% &  & 25.607 & 17.272 & 14.634 &  & 36.068 & 25.320 & 15.738 &  & 44.386 & 28.176 & 12.102 &  & 29.138 & 20.419 & 11.582 \\ \hline
\end{tabular}
}
\vspace{-1mm}
\caption{The table presents the pointwise forecasting errors obtained on the irregular PeMSD3, PeMSD4, PeMSD7, and PeMSD8 benchmark datasets.}
\label{tab-missing1}
\end{table*}

\vspace{-3mm}
\begin{table*}[htbp]
\centering
\renewcommand{\arraystretch}{1.2}
\resizebox{0.745\textwidth}{!}{
\begin{tabular}{c|c|c|ccc|c|ccc|c|ccc}
\hline
\textbf{\begin{tabular}[c]{@{}c@{}}Missing \\ Scheme\end{tabular}} & \textbf{\begin{tabular}[c]{@{}c@{}}Missing \\ Rate\end{tabular}} & \multirow{8}{*}{\textbf{\rotatebox[origin=c]{90}{PeMSD7(M)}}} & \textbf{RMSE} & \textbf{MAE} & \textbf{MAPE} & \multirow{8}{*}{\textbf{\rotatebox[origin=c]{90}{METR-LA}}} & \textbf{RMSE} & \textbf{MAE} & \textbf{MAPE} & \multirow{8}{*}{\textbf{\rotatebox[origin=c]{90}{PeMS-BAY}}} & \textbf{RMSE} & \textbf{MAE} & \textbf{MAPE} \\ \cline{1-2} \cline{4-6} \cline{8-10} \cline{12-14} 
\textbf{MKH-Net} & \textbf{0\%} &  & \textbf{5.817} & \textbf{3.329} & \textbf{6.972} &  & \textbf{7.495} & \textbf{4.594} & \textbf{8.578} &  & \textbf{3.069} & \textbf{1.659} & \textbf{3.070} \\ \cline{1-2} \cline{4-6} \cline{8-10} \cline{12-14} 
\multirow{3}{*}{Point} & 10\% &  & 6.617 & 3.802 & 7.736 &  & 8.624 & 5.534 & 9.283 &  & 3.289 & 1.940 & 3.585 \\
& 30\% &  & 6.758 & 4.294 & 9.193 &  & 9.008 & 5.917 & 10.059 &  & 3.633 & 2.209 & 4.011 \\
& 50\% &  & 7.054 & 4.569 & 9.352 &  & 9.390 & 6.298 & 10.857 &  & 3.817 & 2.364 & 4.273 \\ \cline{1-2} \cline{4-6} \cline{8-10} \cline{12-14} 
\multirow{3}{*}{Block} & 10\% &  & 9.316 & 4.503 & 8.492 &  & 8.906 & 5.605 & 9.779 &  & 3.265 & 1.900 & 3.502 \\
& 30\% &  & 11.787 & 4.793 & 8.590 &  & 9.330 & 6.003 & 10.011 &  & 3.495 & 2.118 & 3.892 \\
& 50\% &  & 14.264 & 6.227 & 10.713 &  & 9.463 & 6.286 & 10.375 &  & 3.710 & 2.294 & 4.181 \\ \hline
\end{tabular}
}
\vspace{-1mm}
\caption{The table presents the pointwise forecasting errors obtained on the irregular PeMSD7(M), METR-LA and PeMS-BAY benchmark datasets.}
\label{tab-missing2}
\end{table*}

\vspace{1mm}
\subsection{SENSITIVITY ANALYSIS}
\label{sensitivityanalysis}
\vspace{0mm}
We conducted a hyperparameter study to determine the impact of specific hyperparameters on the performance of the proposed framework. The goal is to determine the optimal set of hyperparameter values that lead to the best performance on benchmark datasets. We have tuned four hyperparameters - embedding size($\textit{d}$), number of hyperedges($|\mathcal{HE}|$), batch size($\textit{b}$), and learning rate($\textit{lr}$) - over specific ranges of values. The ranges were as follows: embedding dimension($\textit{d}$) $\in \{2, 6, 10, 18, 24\}$, the number of hyperedges($|\mathcal{HE}|$) $\in \{2, 5, 8\}$, batch size($\textit{b}$) $ \in \{2, 6, 10, 18, 24, 32, 64\}$, and the learning rate($\textit{lr}$) $ \in \{\num{1e-1}, \num{1e-2}, \num{1e-3}, \num{1e-4}\}$. The ranges were chosen to avoid memory errors and limit model size. The hyperparameters of the framework were optimized using grid search, while the model performance was evaluated using multiple metrics such as MAE and RMSE. The study revealed the impact of varying the critical hyperparameters, including embedding size and the number of hyperedges in a predetermined range across all datasets on the \textbf{MKH-Net} framework performance. These experimental findings provided crucial insights into the effect of these specific hyperparameters on the framework's ability to generate accurate forecasts in multivariate time series analysis, facilitating a better understanding of its overall performance. The optimal set of hyperparameter configurations that yielded the best performance for each dataset are discussed below,

\vspace{1mm}
\begin{itemize}
    \item For PeMSD3, we set the batch size($\textit{b}$) to 6, the initial learning rate($\textit{lr}$) to \num{1e-3}, and the embedding size($\textit{d}$) to 16. Additionally, the number of hyperedges is 8.    
    \item For PeMSD4, we set the batch size($\textit{b}$) to 32, the initial learning rate($\textit{lr}$) to \num{1e-3}, and the embedding size($\textit{d}$) to 18. Additionally, the number of hyperedges is 5.    
    \item For PeMSD7, we set the batch size($\textit{b}$) to 6, the initial learning rate($\textit{lr}$) to \num{1e-3}, and the embedding size($\textit{d}$) to 18. Additionally, the number of hyperedges is 8.    
    \item For PeMSD8, we set the batch size($\textit{b}$) to 18, the initial learning rate($\textit{lr}$) to \num{1e-3}, and the embedding size($\textit{d}$) to 16. Additionally, the number of hyperedges is 8.     
    \item For PeMSD7(M), we set the batch size($\textit{b}$) to 18, the initial learning rate($\textit{lr}$) to \num{1e-3}, and the embedding size($\textit{d}$) to 12. Additionally, the number of hyperedges is 4. 
    \item For METR-LA, we set the batch size($\textit{b}$) to 6, the initial learning rate($\textit{lr}$) to \num{1e-3}, and the embedding size($\textit{d}$) to 16. Additionally, the number of hyperedges is 6. 
    \item For PEMS-BAY, we set the batch size($\textit{b}$) to 6, the initial learning rate($\textit{lr}$) to \num{1e-3}, and the embedding size($\textit{d}$) to 16. Additionally, the number of hyperedges is 8. 
\end{itemize}

\begin{figure*}[!ht]
\centering
\subfloat[PeMSD3]{
 \includegraphics[width=45mm]{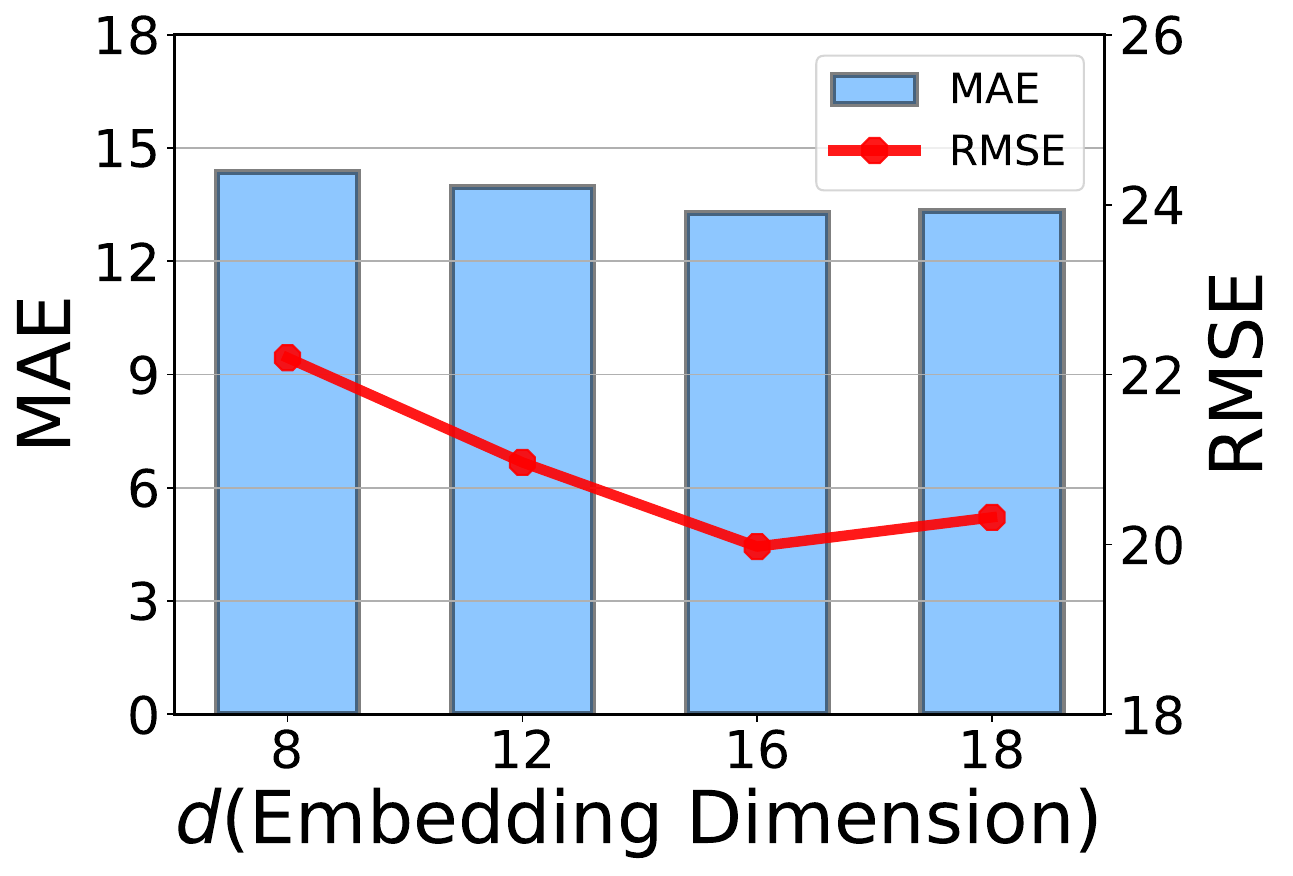}
}
\subfloat[PeMSD4]{
 \includegraphics[width=45mm]{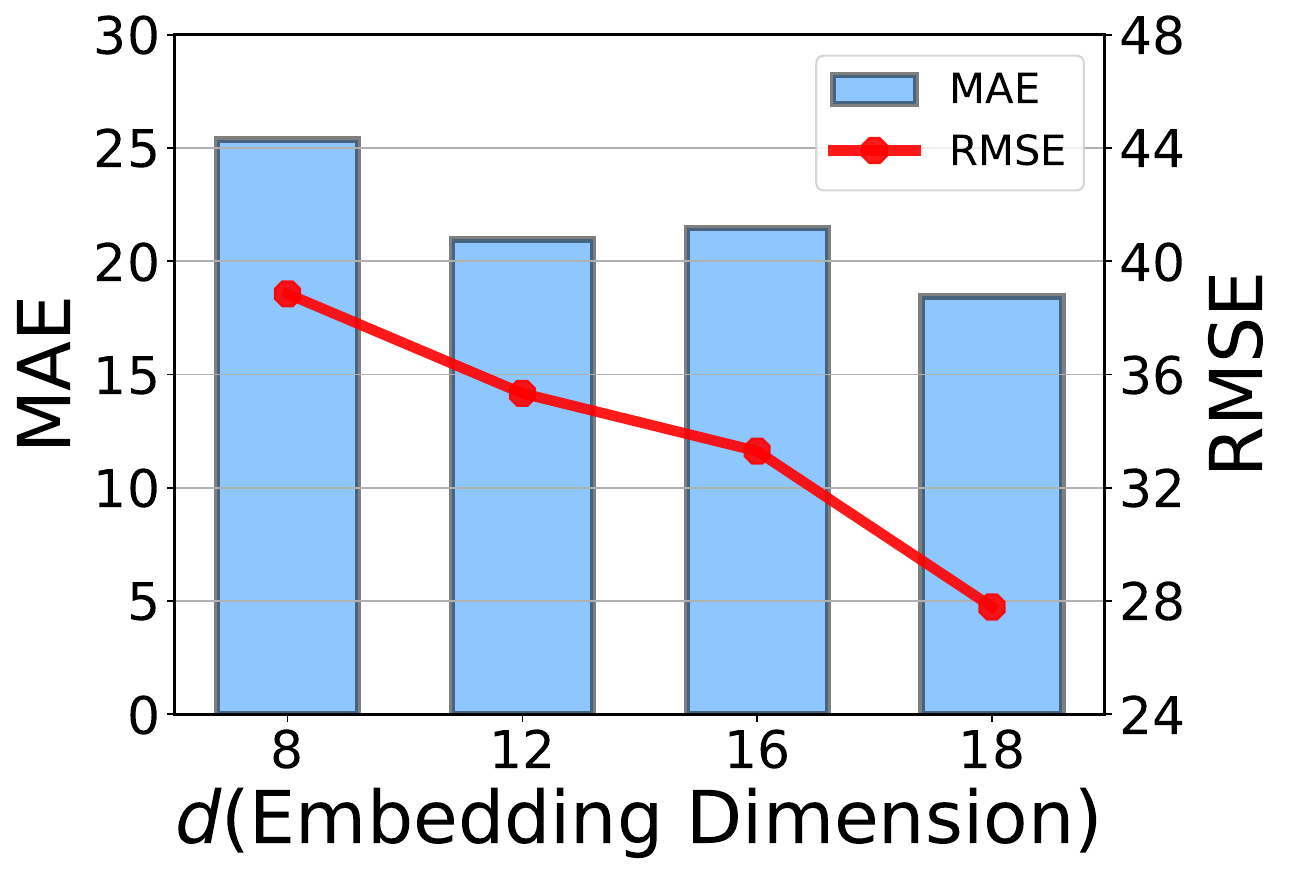}
}
\subfloat[PeMSD7]{
 \includegraphics[width=45mm]{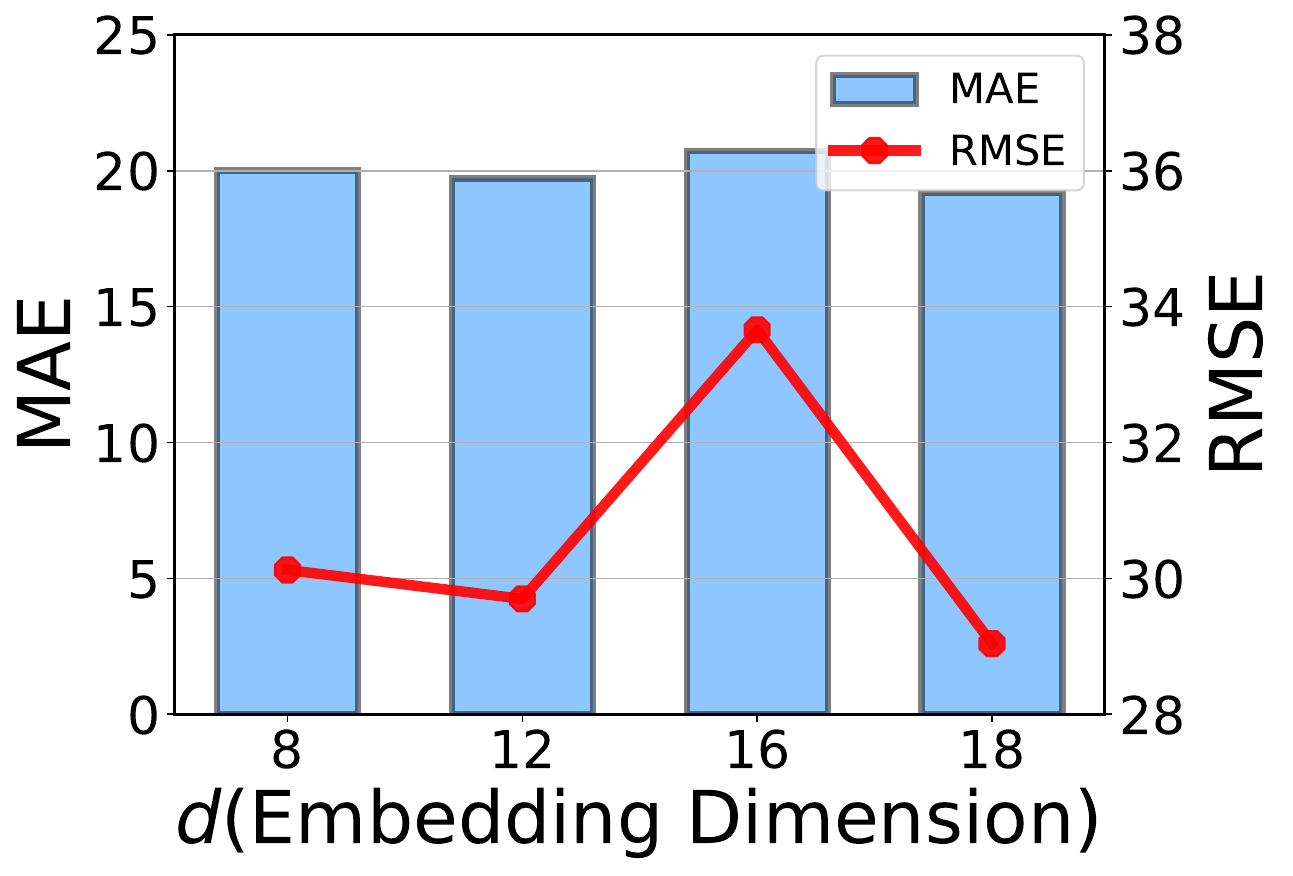}
}
\subfloat[PeMSD7(M)]{
 \includegraphics[width=45mm]{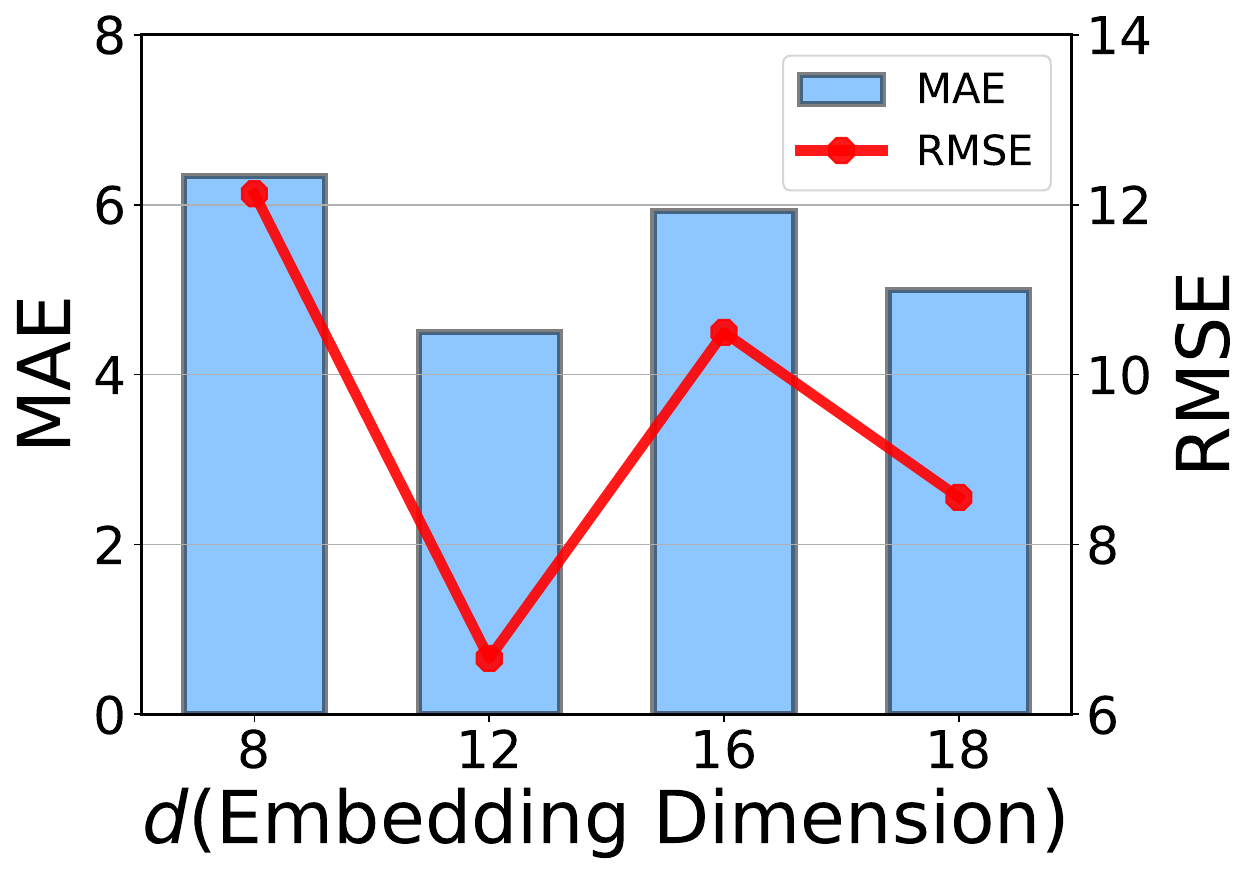}
} \\[-1ex]
\subfloat[PeMSD8]{
 \includegraphics[width=45mm]{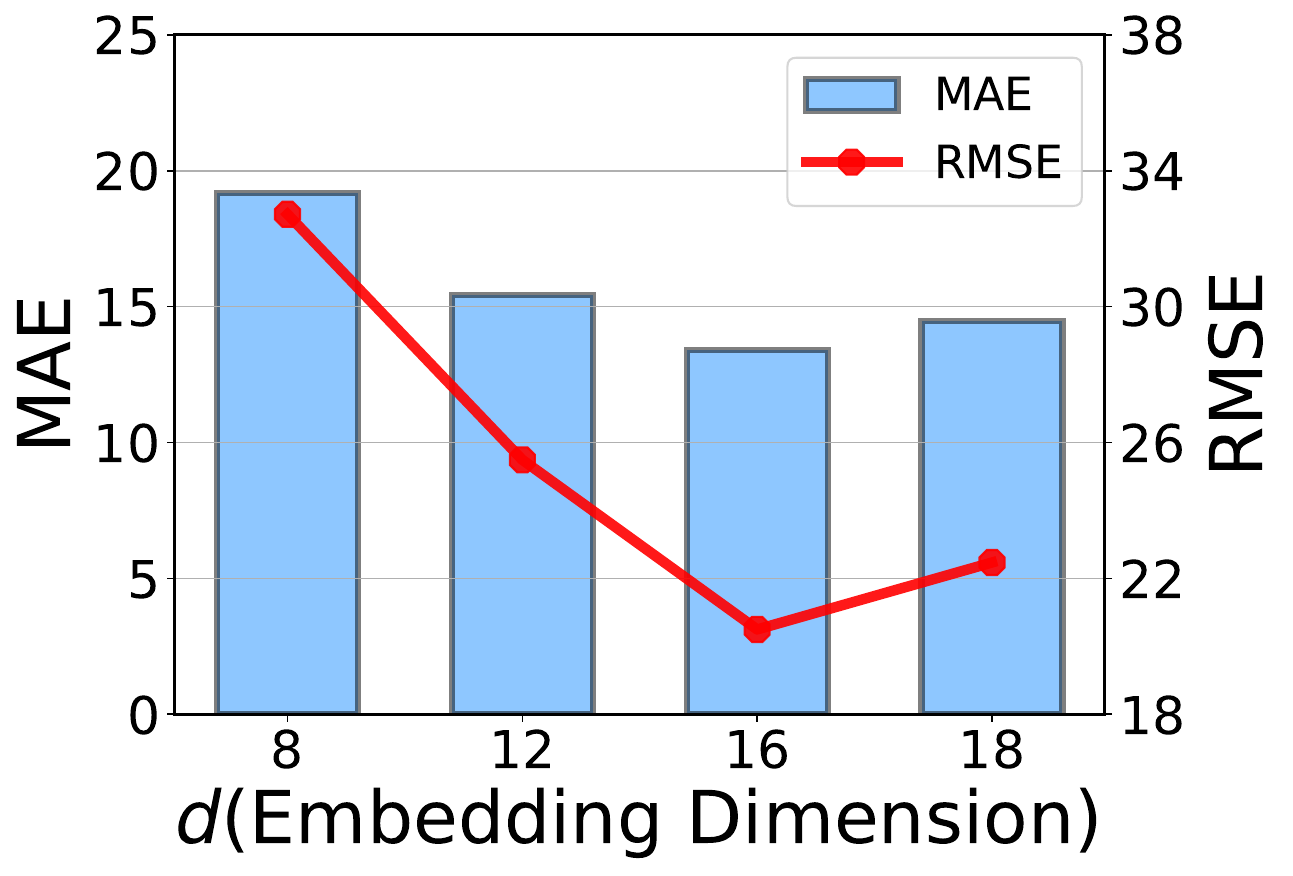}
}
\subfloat[METR-LA]{
 \includegraphics[width=45mm]{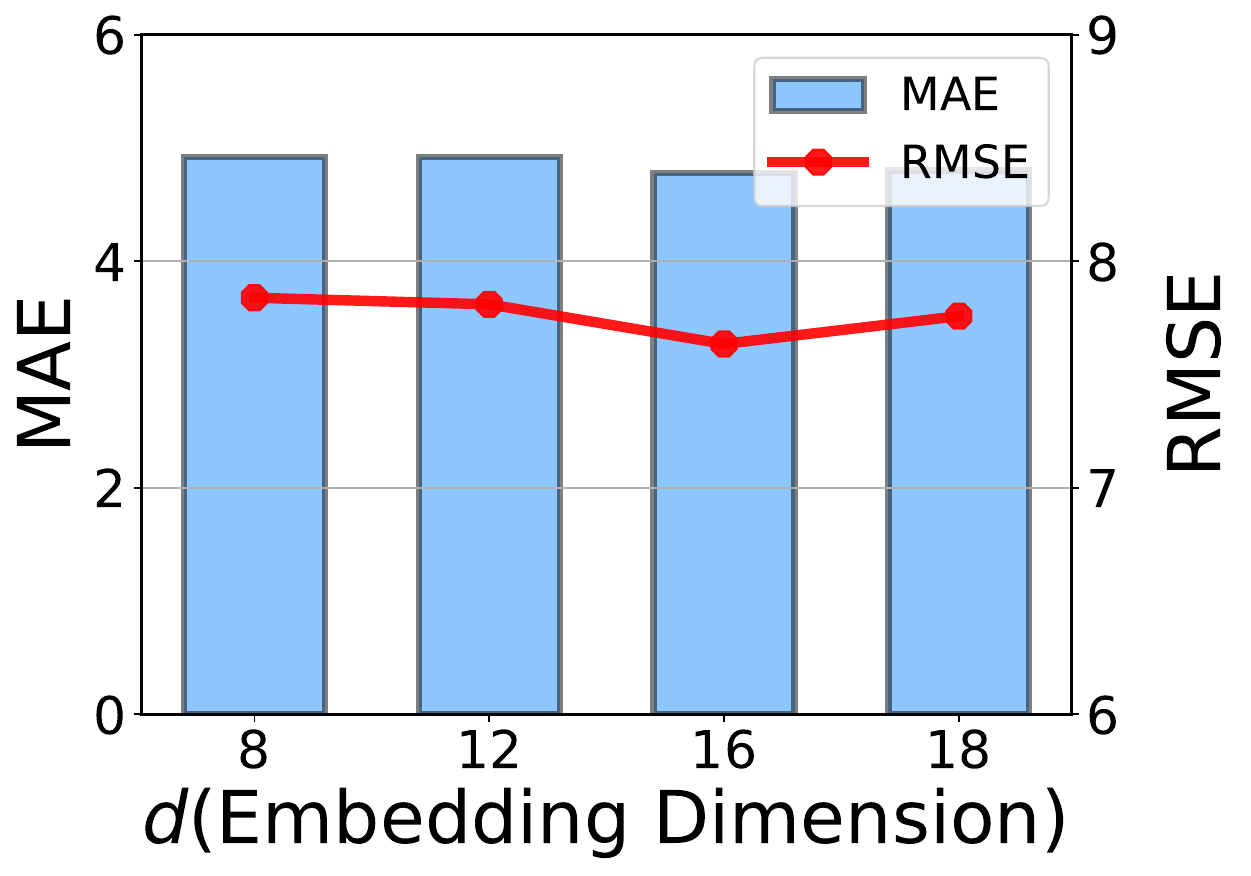}
}
\subfloat[PEMS-BAY]{
 \includegraphics[width=45mm]{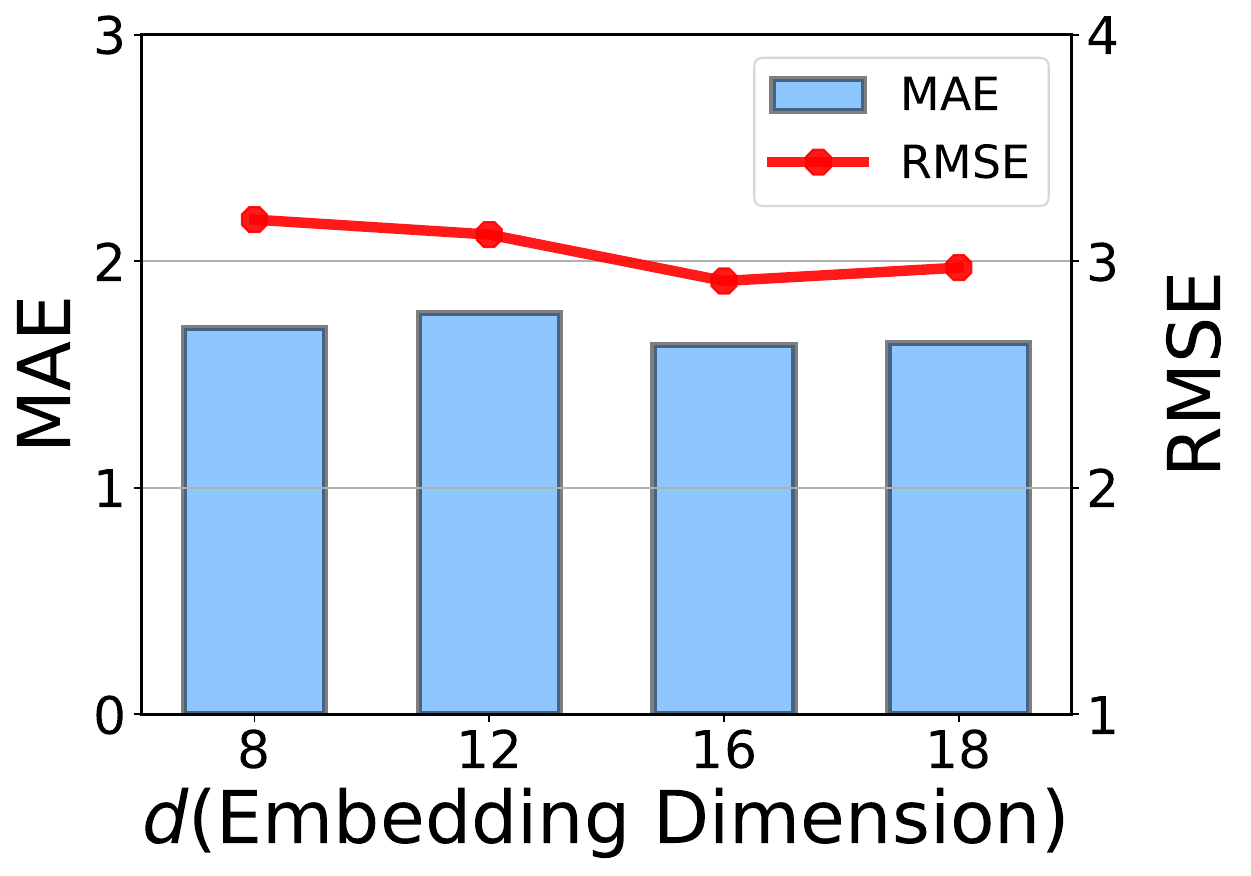}
}
\subfloat[PeMSD3]{
 \includegraphics[width=45mm]{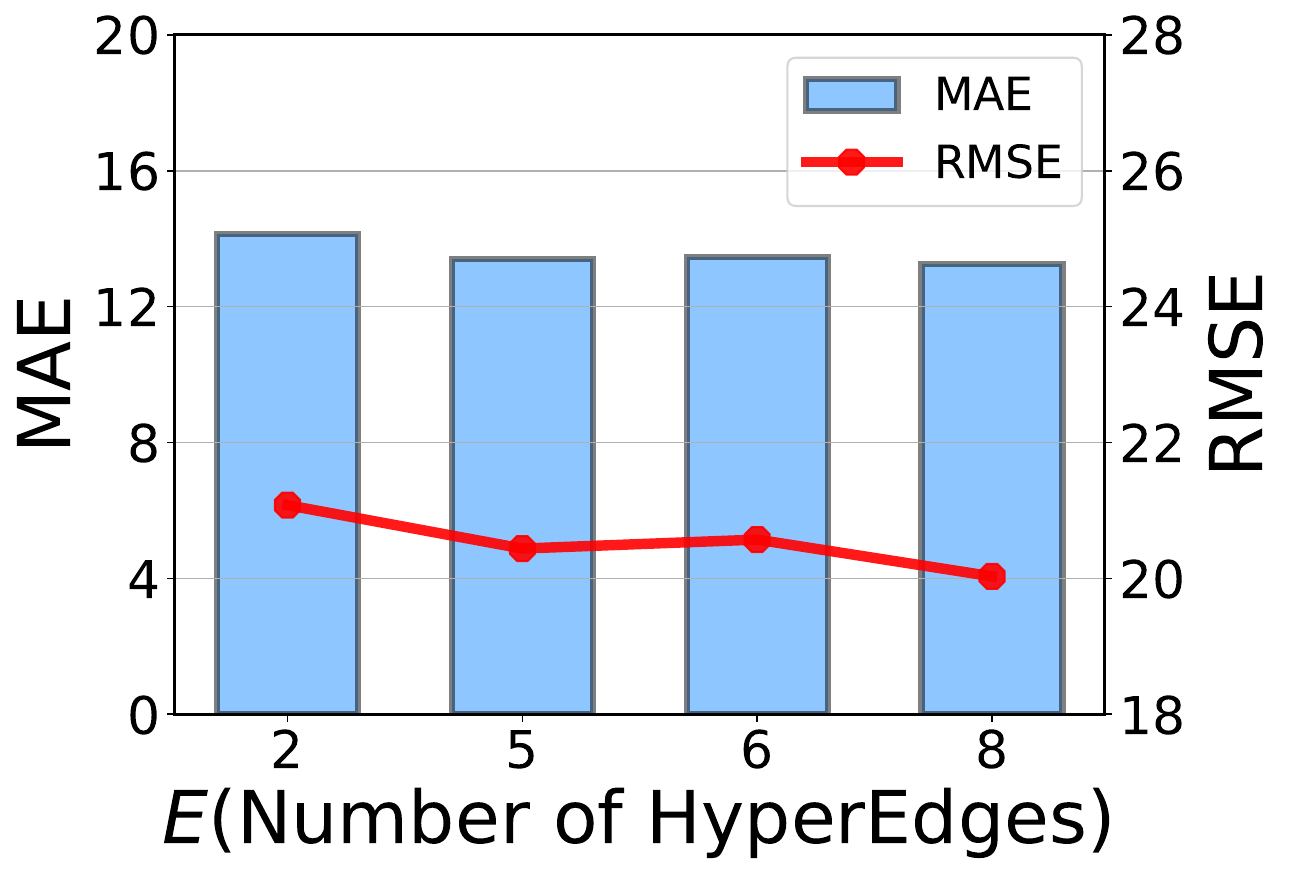}
}\\[-1ex]
\subfloat[PeMSD4]{
 \includegraphics[width=45mm]{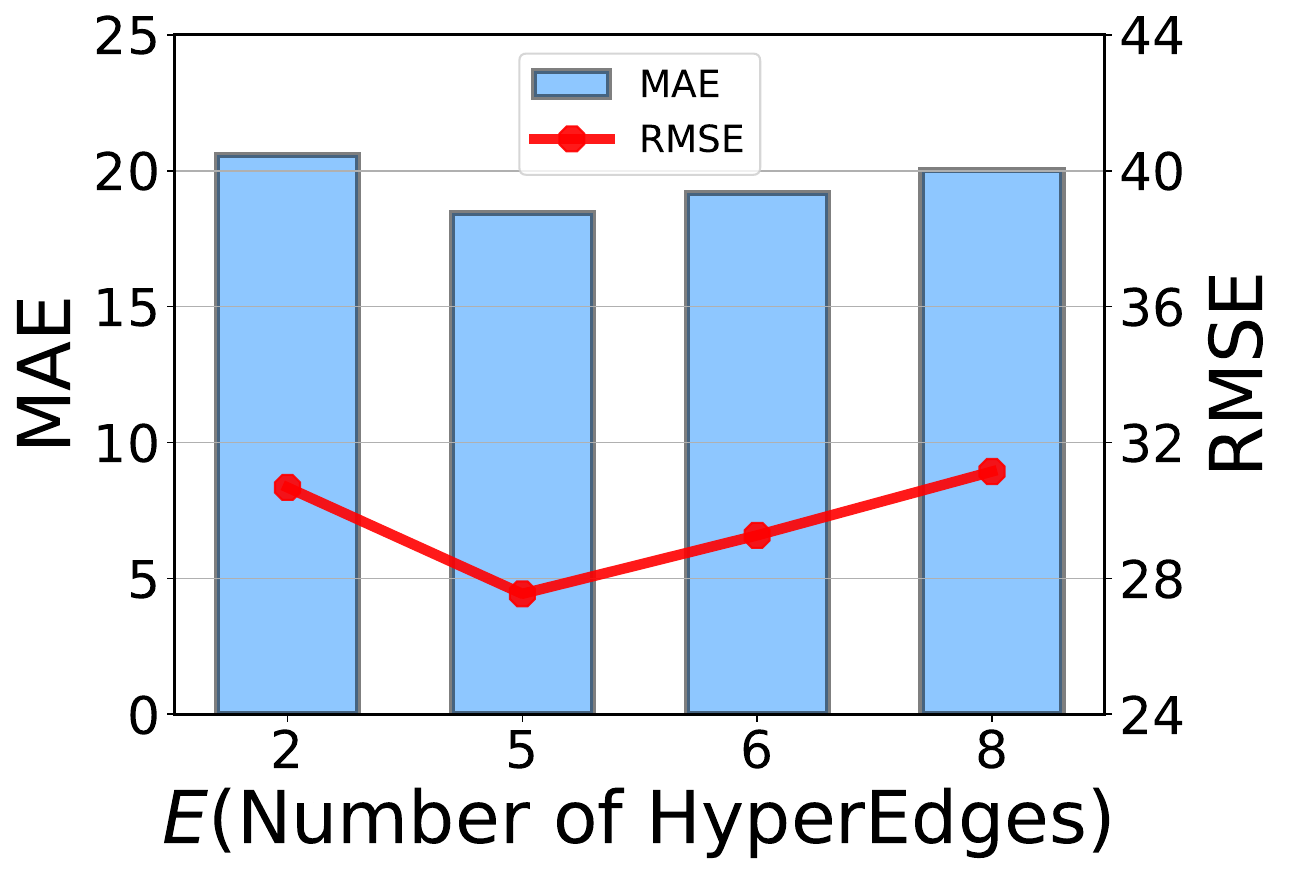}
}
\subfloat[PeMSD7]{
 \includegraphics[width=45mm]{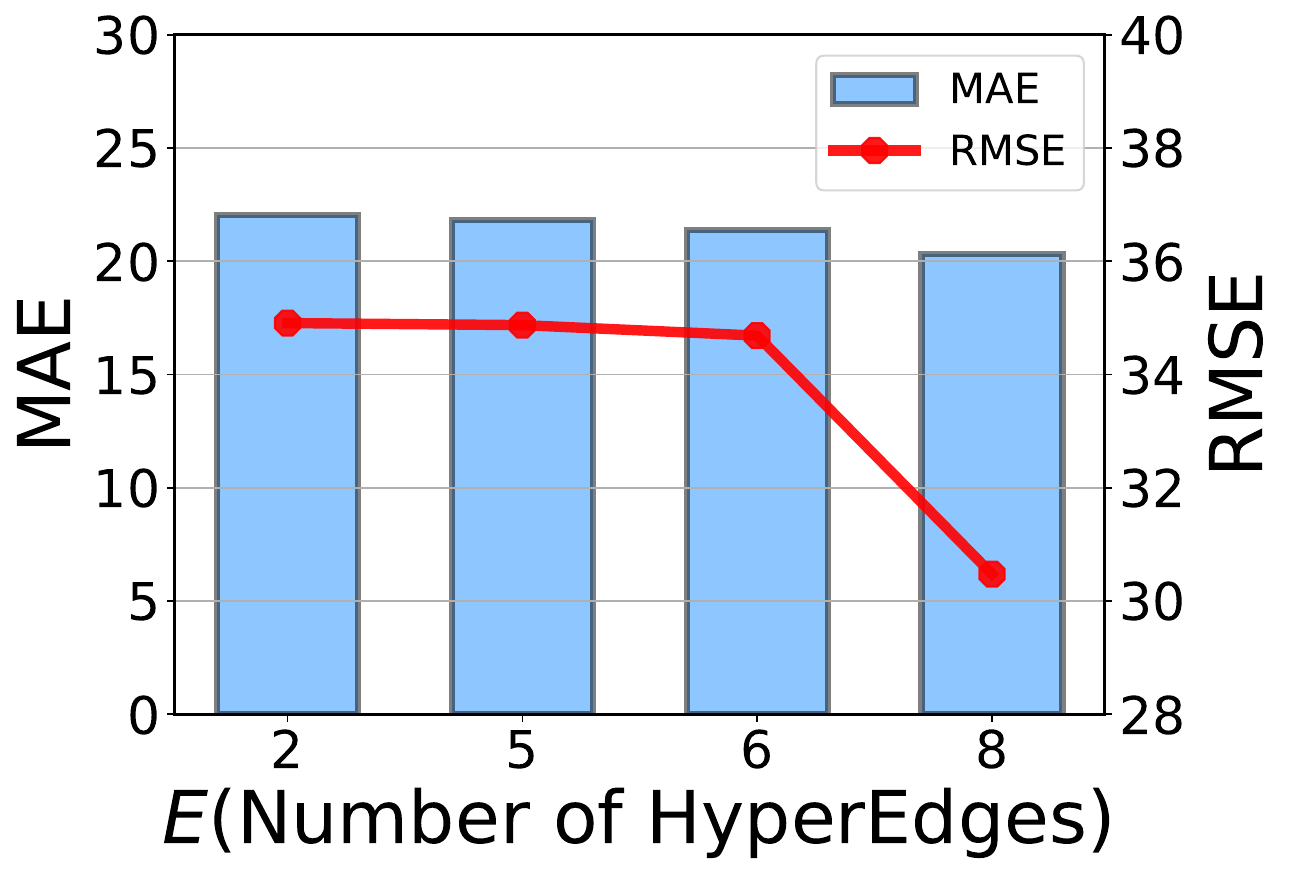}
}
\subfloat[PeMSD7(M)]{
 \includegraphics[width=45mm]{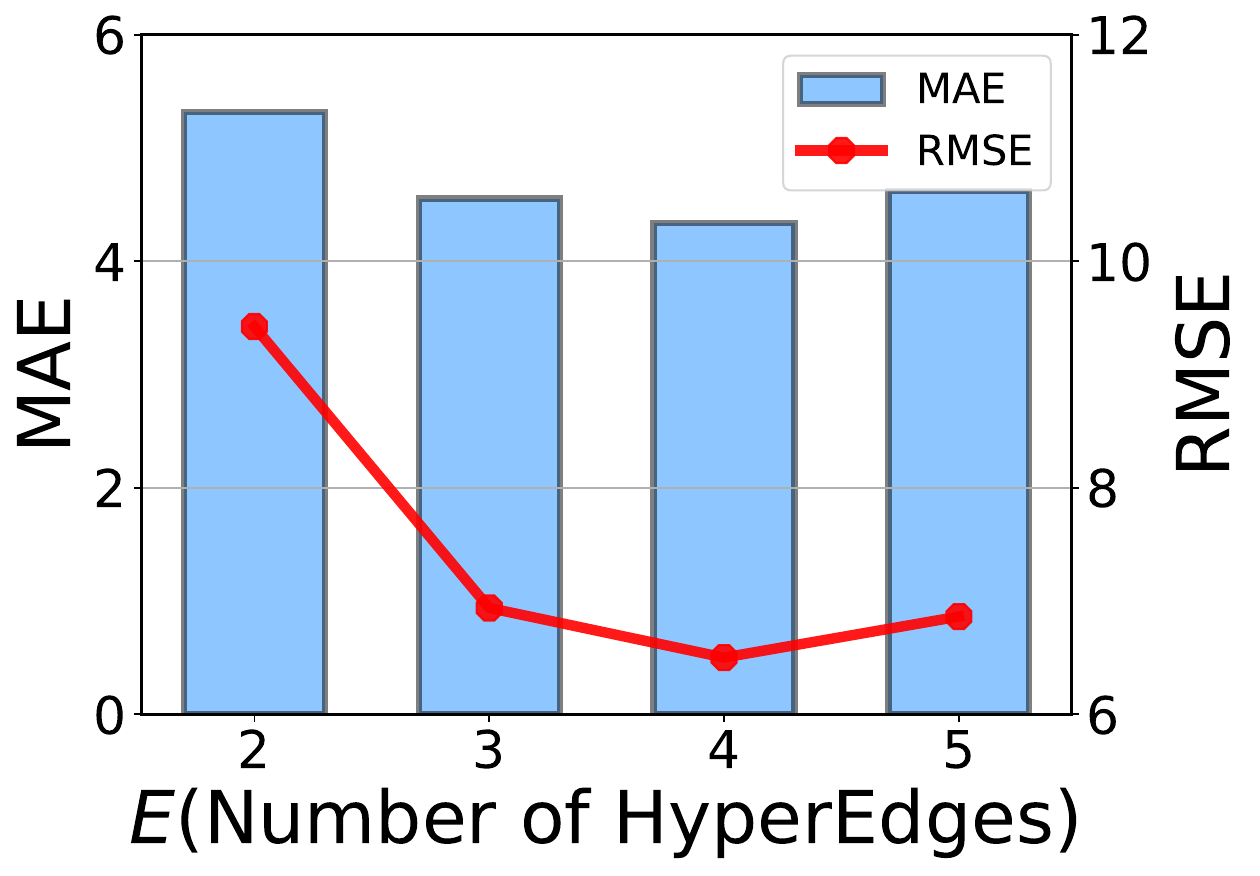}
} 
\subfloat[PeMSD8]{
 \includegraphics[width=45mm]{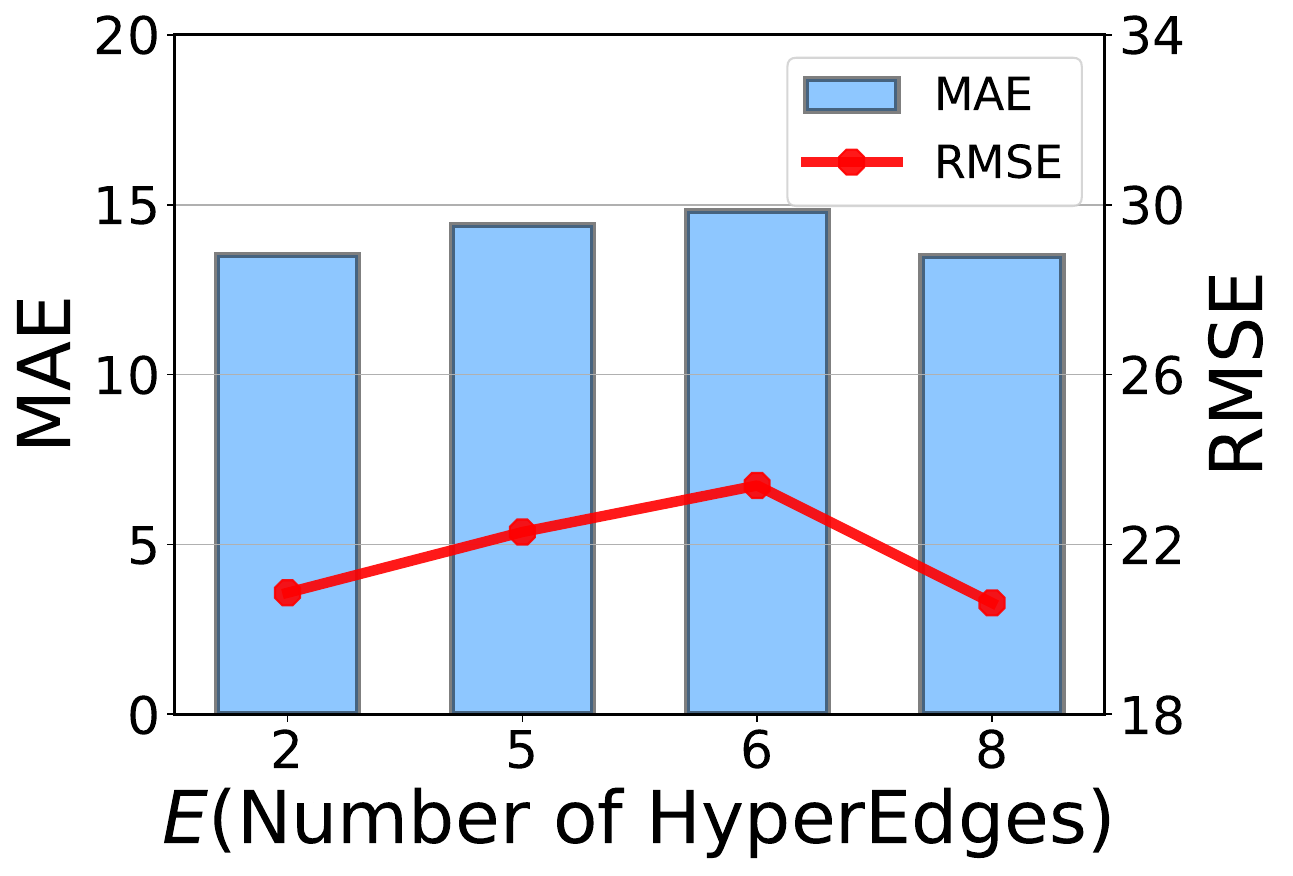}
}\\[-1ex]
\subfloat[METR-LA]{
 \includegraphics[width=45mm]{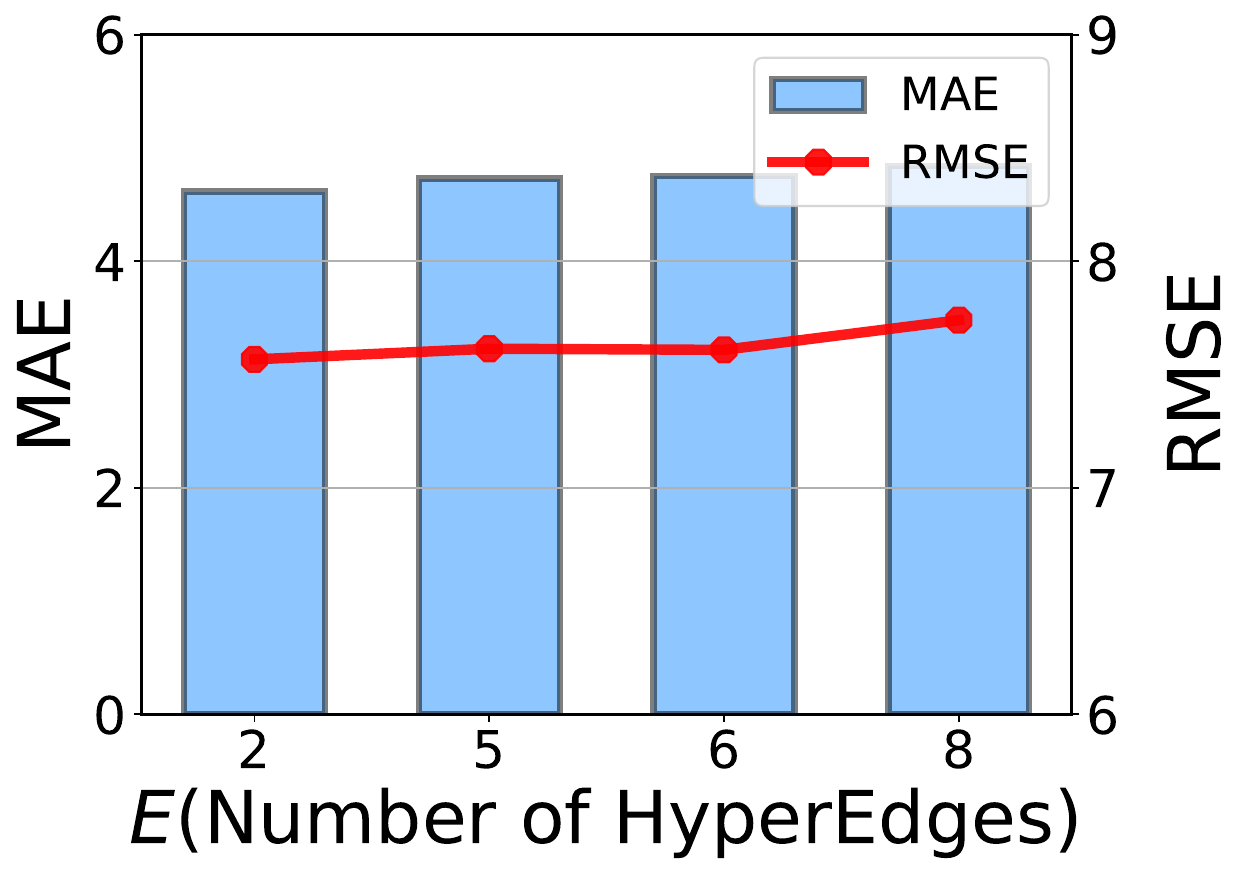}
}
\subfloat[PEMS-BAY]{
 \includegraphics[width=45mm]{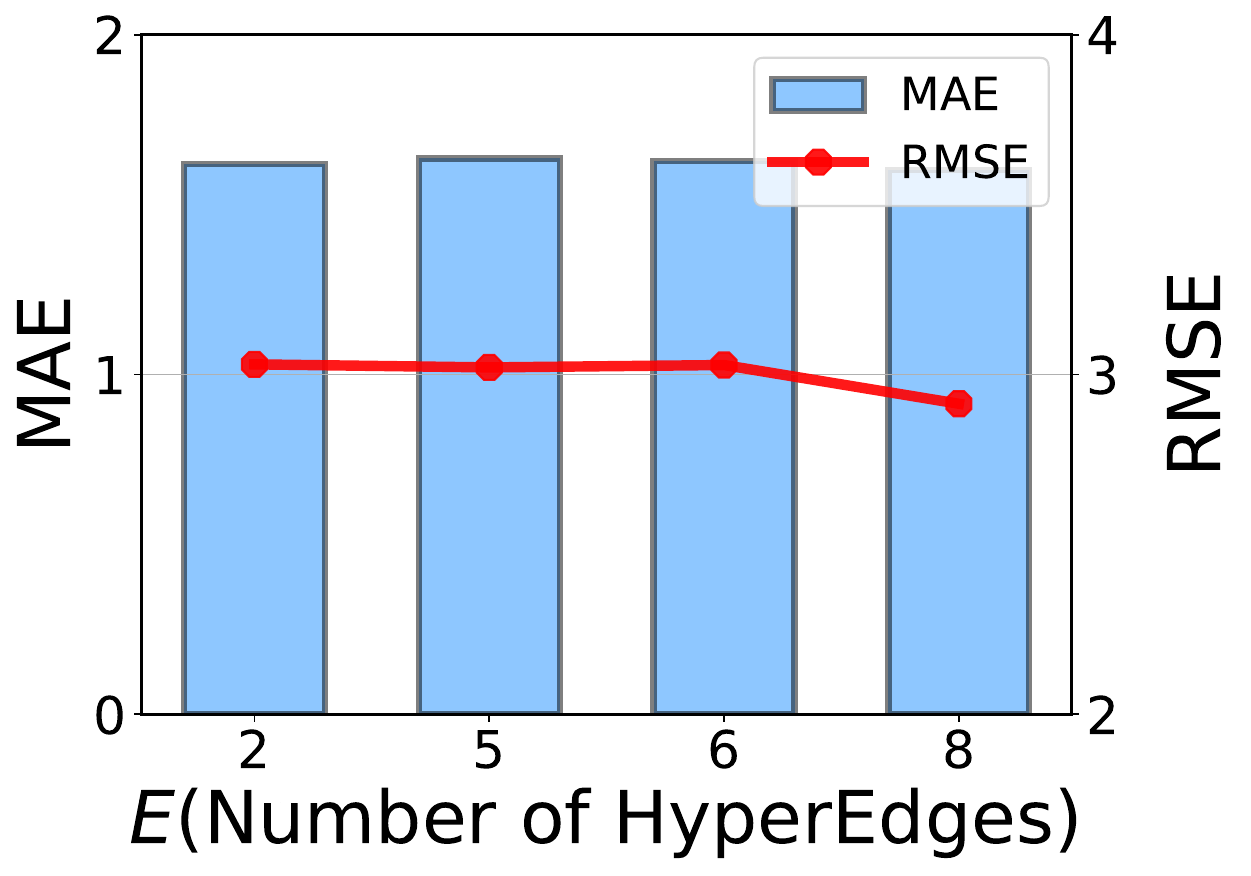}
}
\\[-1ex]
\caption{The figure showcases the results of a sensitivity analysis on the embedding dimension and number of hyperedges.}
\label{fig-sensitivity}
\end{figure*}

\vspace{-2mm}
The fraction of hypernodes connected to hyperedges in a hypergraph provides information about the ``edge density" of the network. A higher fraction of hypernodes connected to hyperedges indicates a denser network, while a lower fraction suggests a sparser network. Controlling the density of the hypergraph is possible by changing the number of hyperedges.
This experiment was carried out to determine the optimal number of hyperedges for an MTSF task by examining the impact of the number of predefined hyperedges on the learned hypergraph structures. This study helps to understand how the density of the hypergraph changes as the number of hyperedges increases or decreases for a given dataset on the MTSF task. Figure \ref{fig-sensitivity} shows the experimental results. In another experiment, we fixed the number of hyperedges to 5 and studied how the fraction of incident hypernodes varies for each hyperedge across the datasets in the MTSF task. We observed that the fraction of hypernodes connected to the hyperedges remained the same across the benchmark datasets. However, further exploration is necessary to gain better insights and understand the trends.

\subsection{TIME SERIES FORECASTING VISUALIZATION}
\begin{figure*}[htbp]
\centering
\subfloat[Node 12 in PeMSD3]{
 \includegraphics[width=45mm]{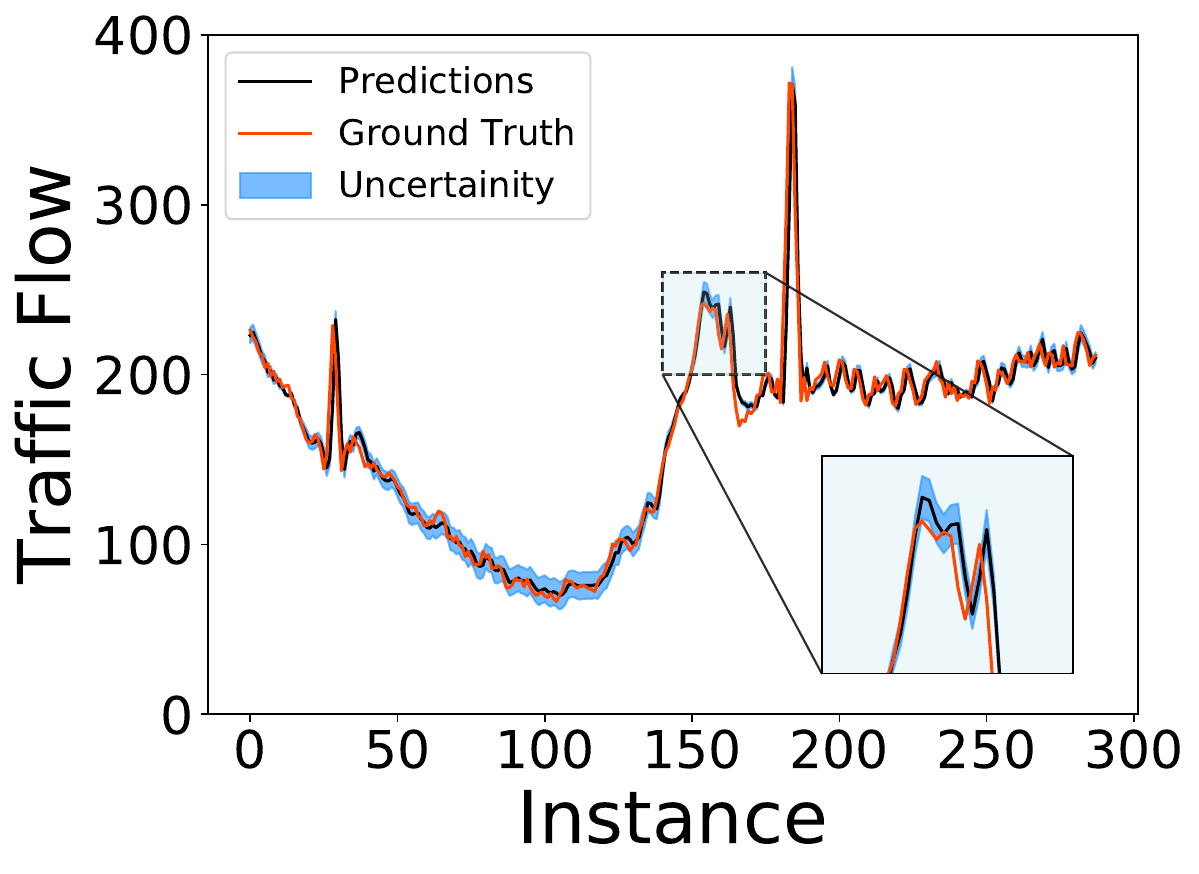}
}
\subfloat[Node 99 in PeMSD3]{
 \includegraphics[width=45mm]{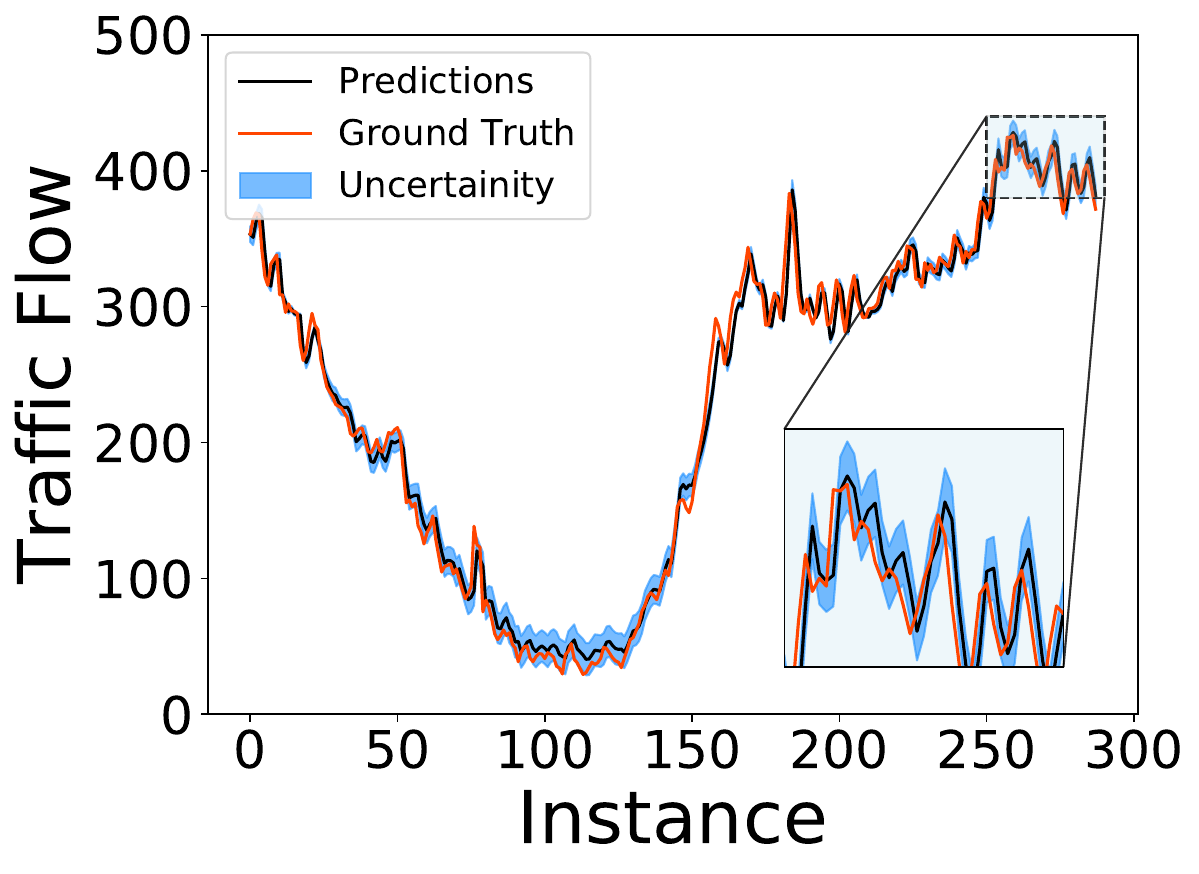}
}
\subfloat[Node 108 in PeMSD3]{
 \includegraphics[width=45mm]{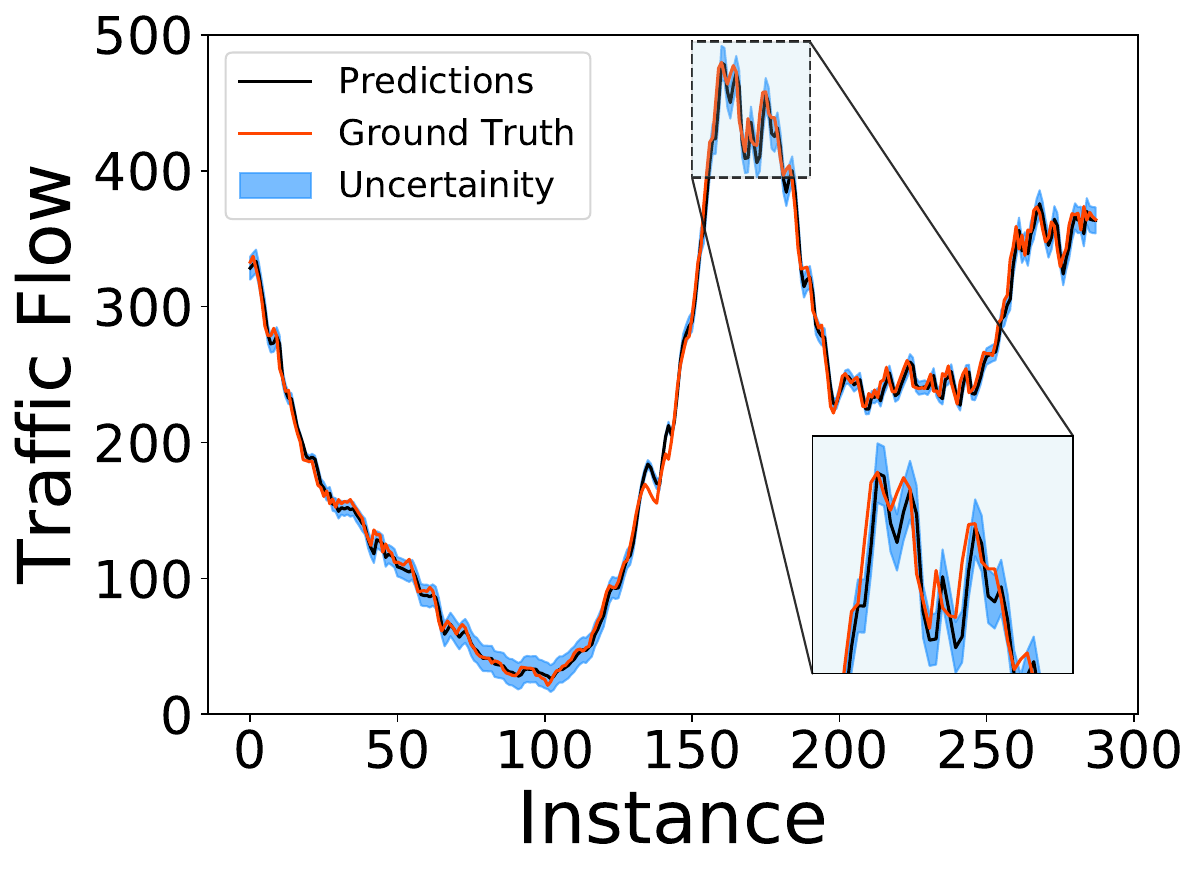}
}
\subfloat[Node 141 in PeMSD3]{
 \includegraphics[width=45mm]{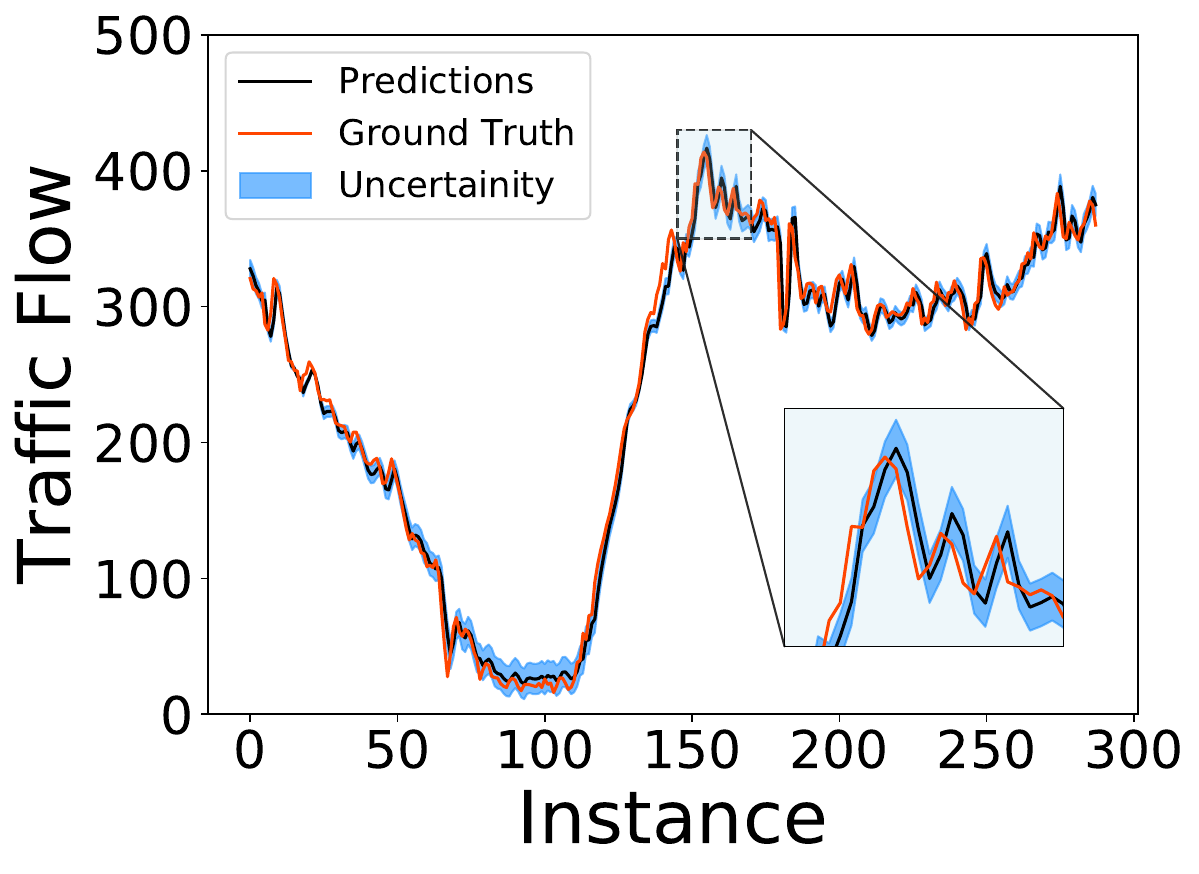}
} \\[-1.5ex]
\hspace{0mm}
\subfloat[Node 149 in PeMSD4]{
 \includegraphics[width=45mm]{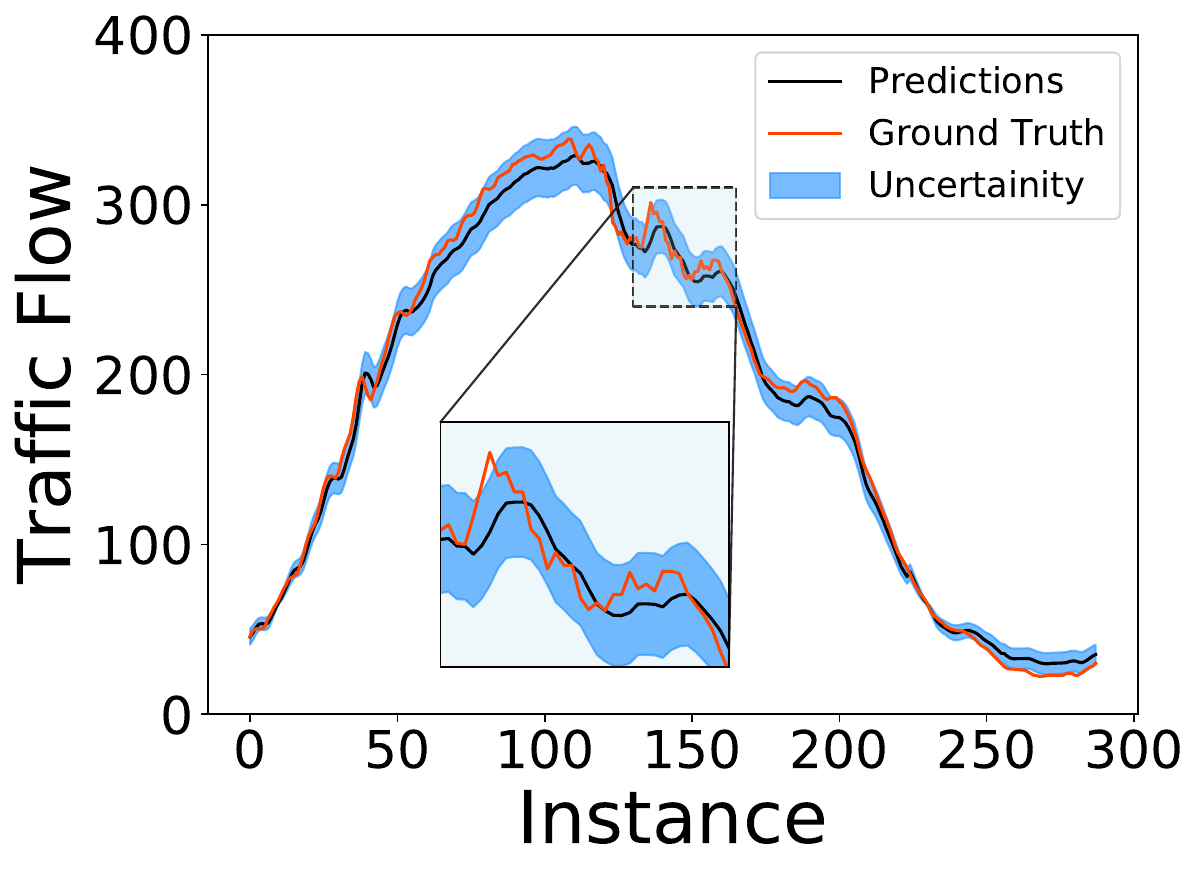}
}
\subfloat[Node 170 in PeMSD4]{
 \includegraphics[width=45mm]{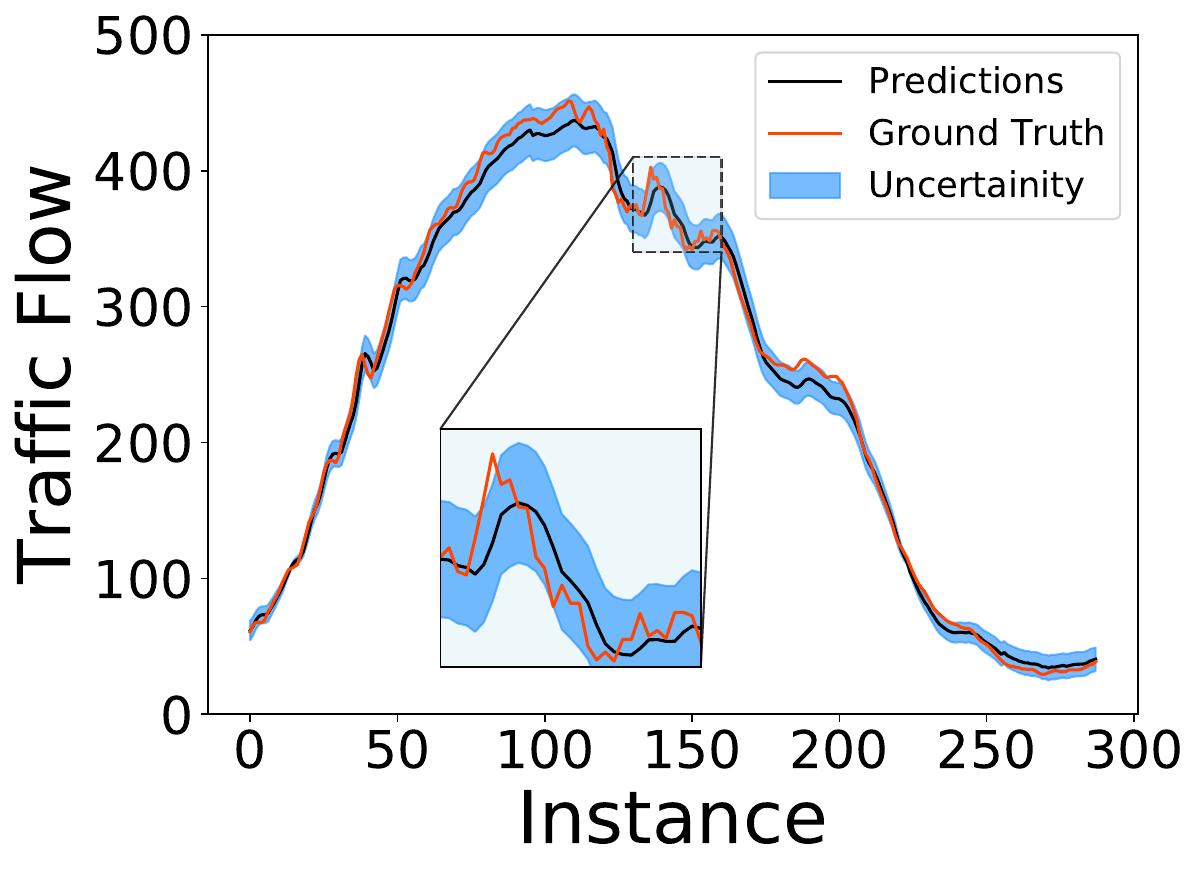}
}
\subfloat[Node 211 in PeMSD4]{
 \includegraphics[width=45mm]{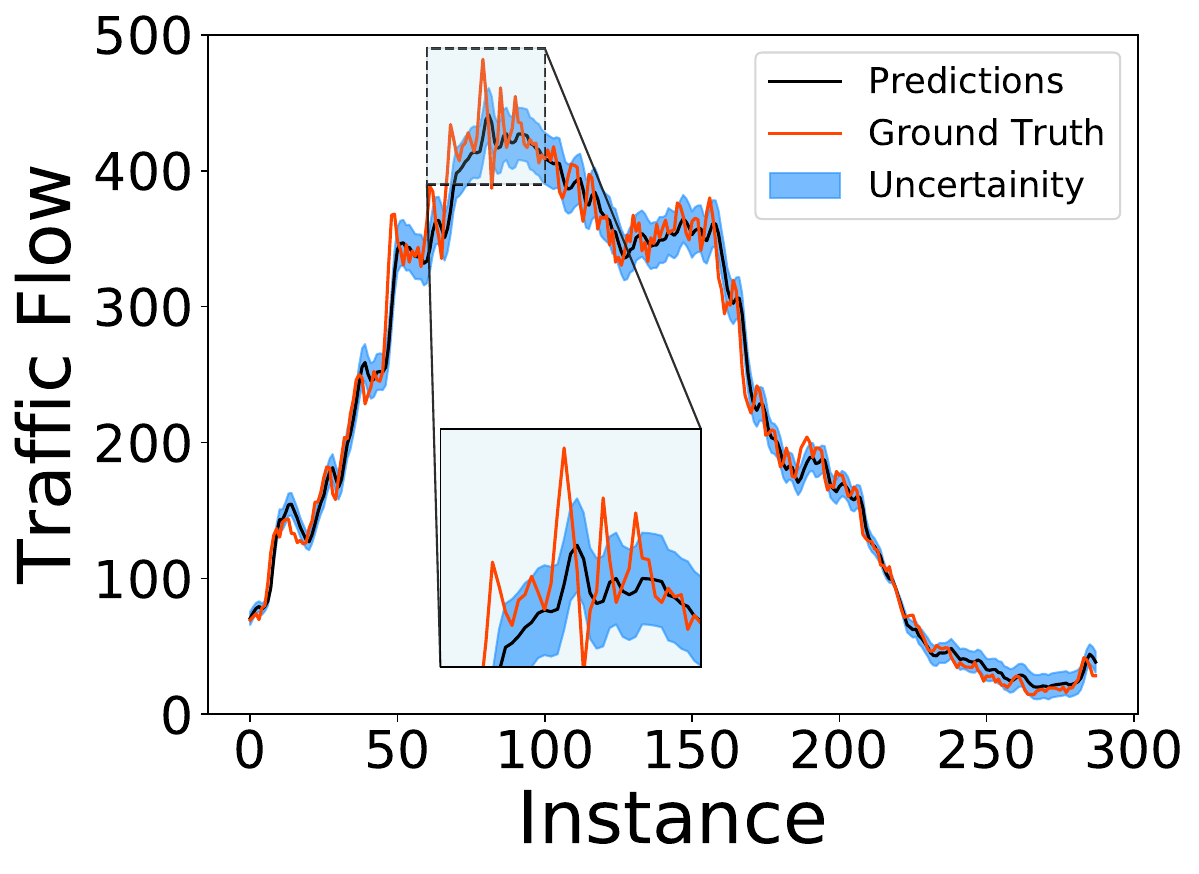}
}
\subfloat[Node 287 in PeMSD4]{
 \includegraphics[width=45mm]{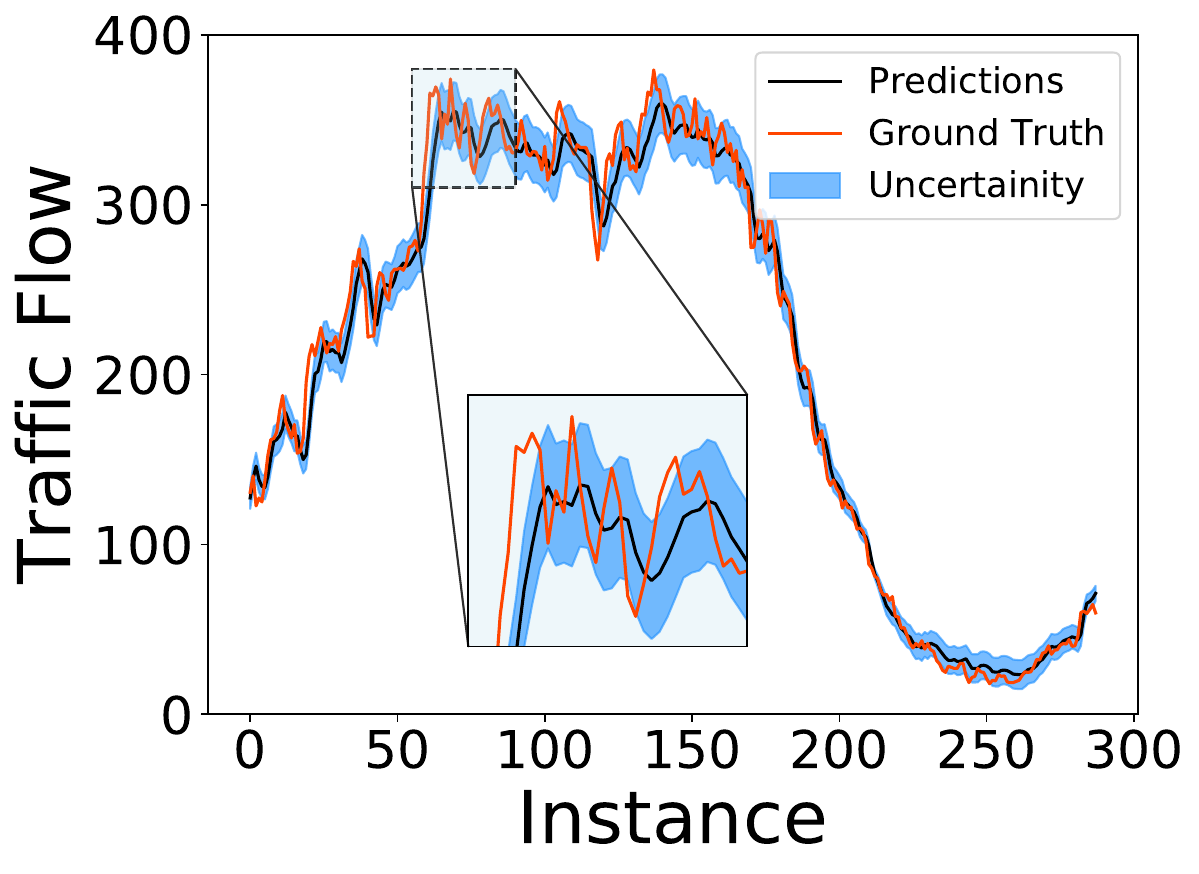}
} \\[-1.5ex]
\hspace{0mm}
\subfloat[Node 26 in PeMSD7]{
 \includegraphics[width=45mm]{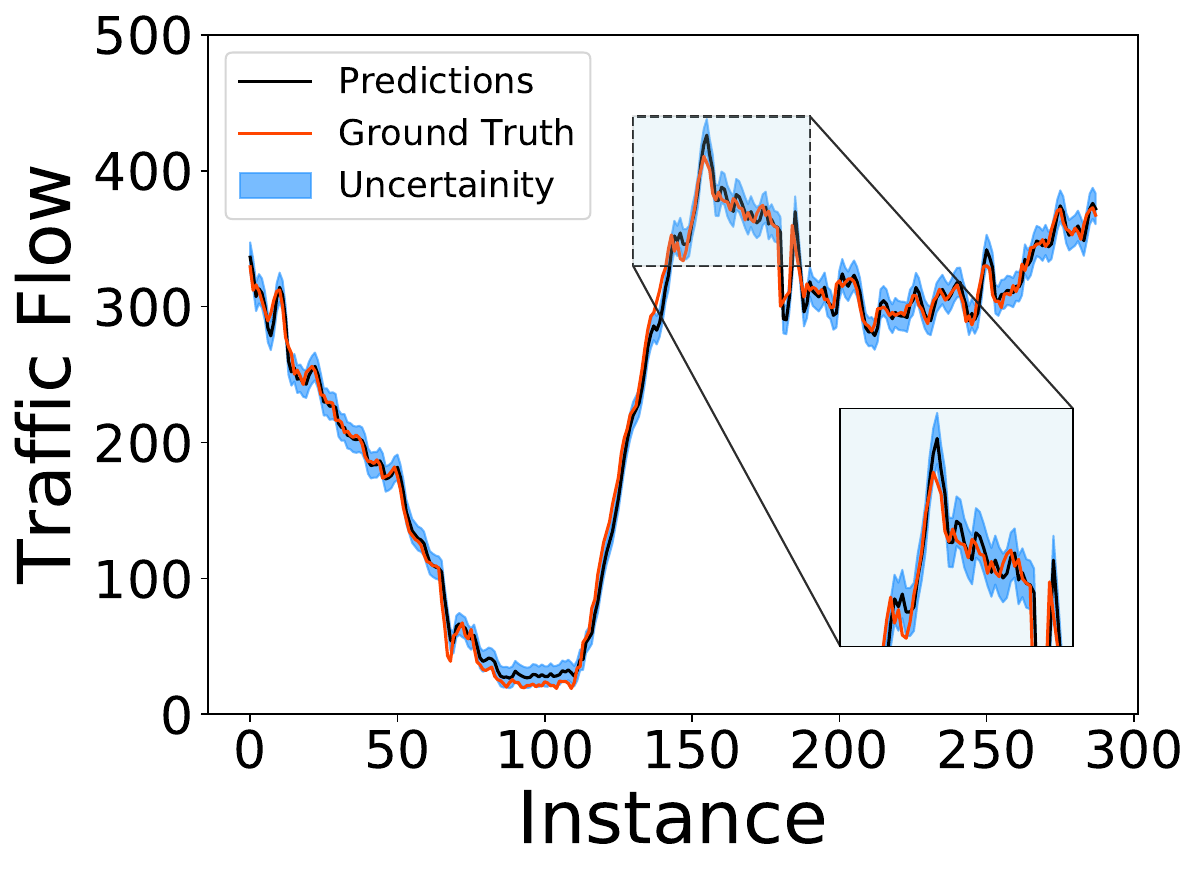}
}
\subfloat[Node 66 in PeMSD7]{
 \includegraphics[width=45mm]{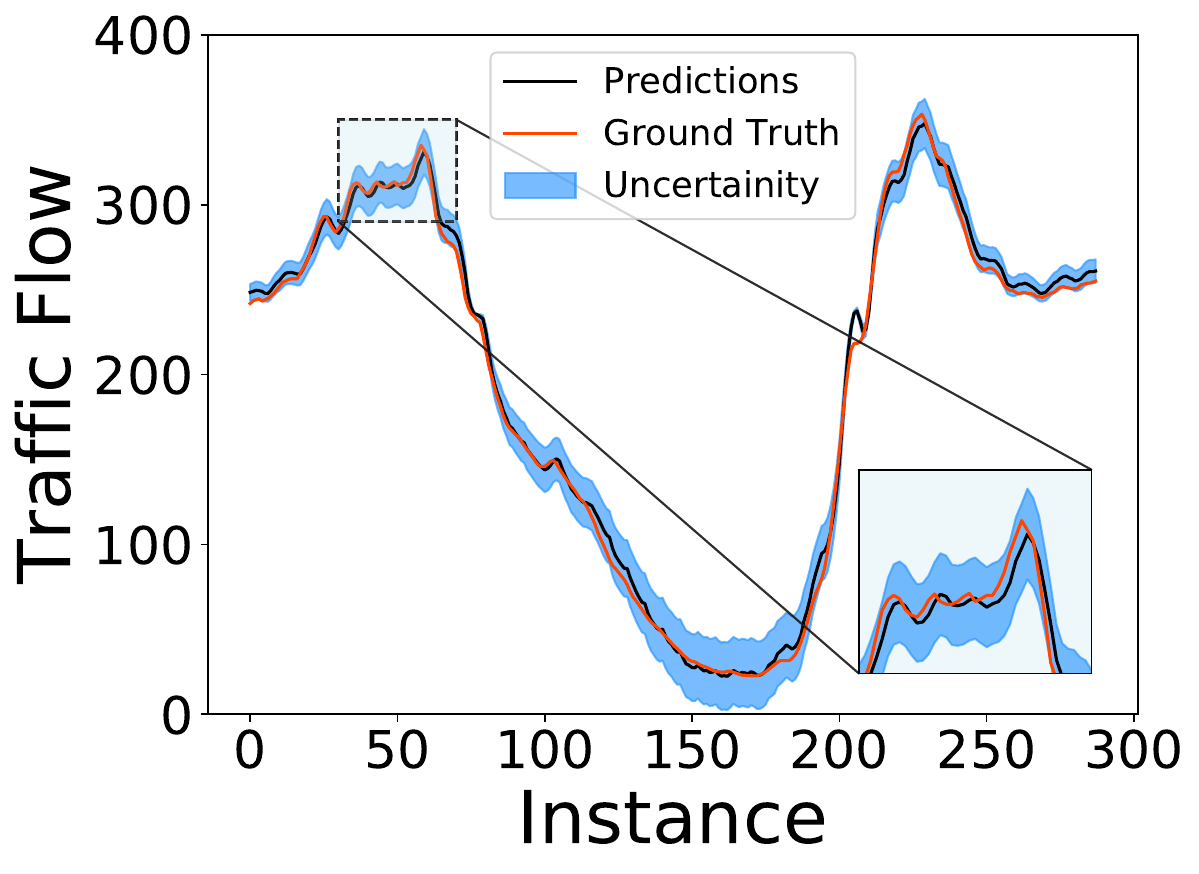}
}
\subfloat[Node 139 in PeMSD7]{
 \includegraphics[width=45mm]{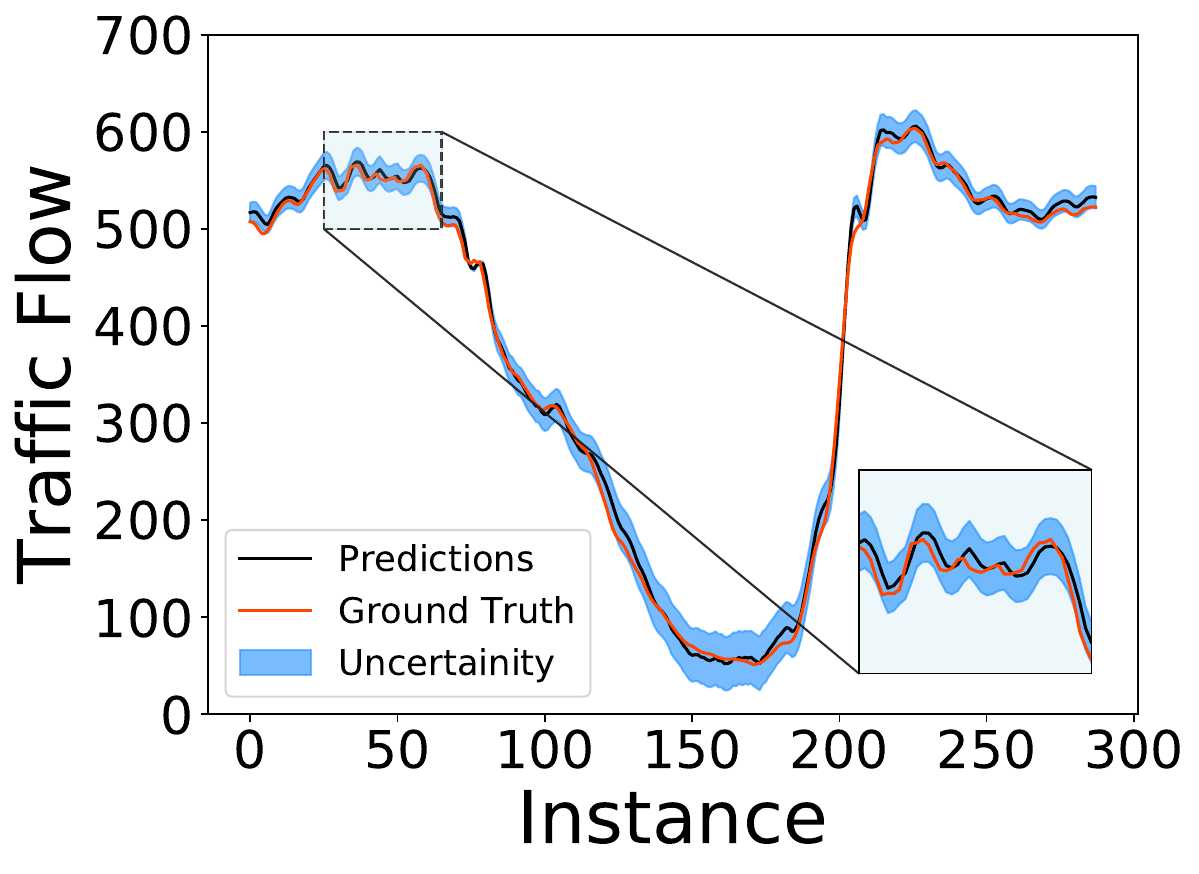}
}
\subfloat[Node 277 in PeMSD7]{
 \includegraphics[width=45mm]{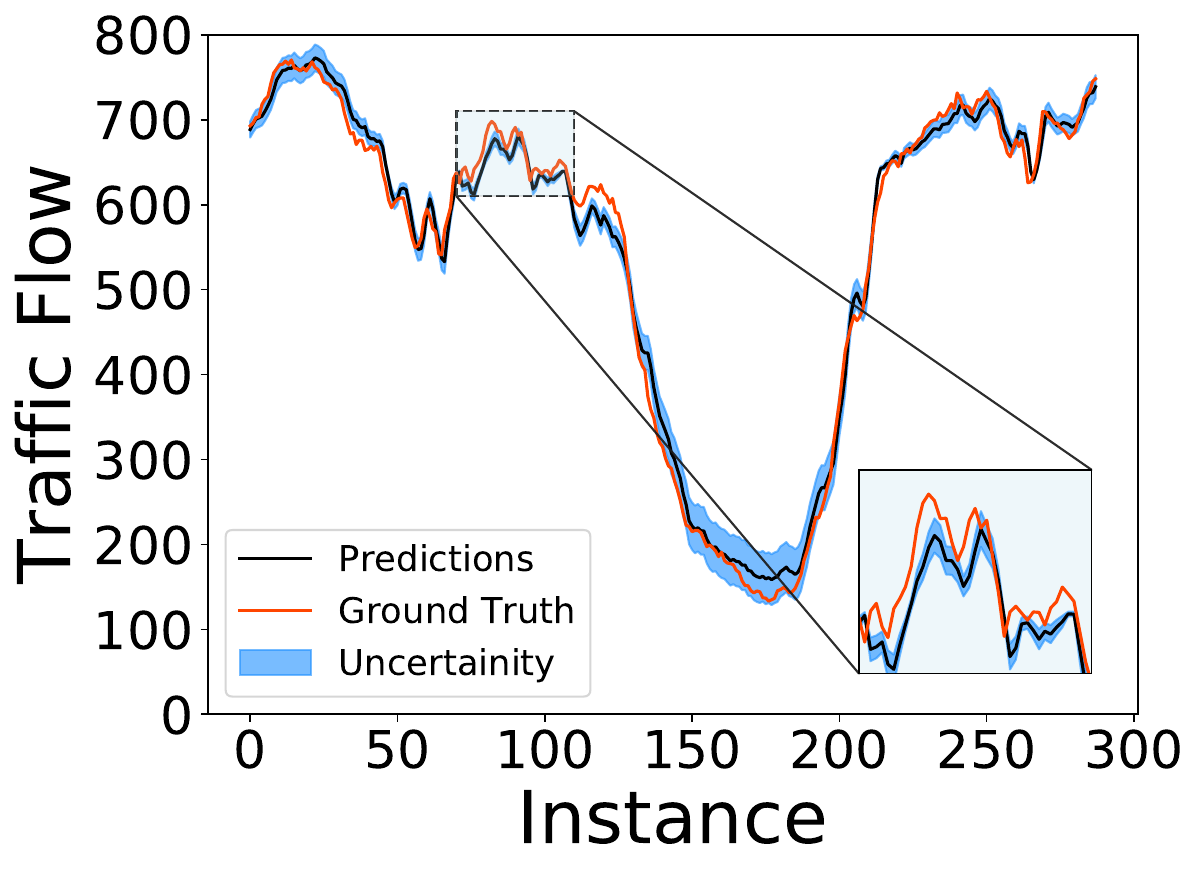}
} \\[-1.5ex]
\hspace{0mm}
\subfloat[Node 85 in PeMSD8]{
 \includegraphics[width=45mm]{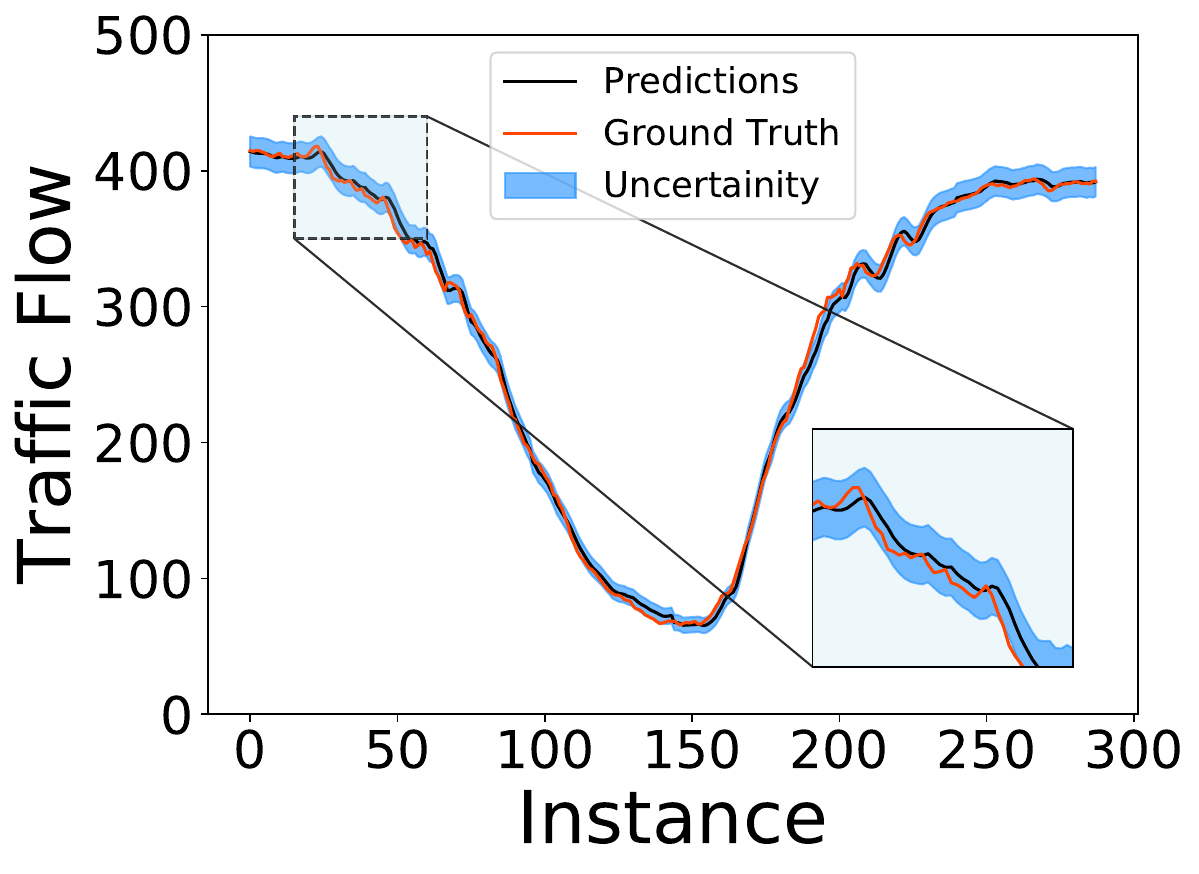}
}
\subfloat[Node 104 in PeMSD8]{
 \includegraphics[width=45mm]{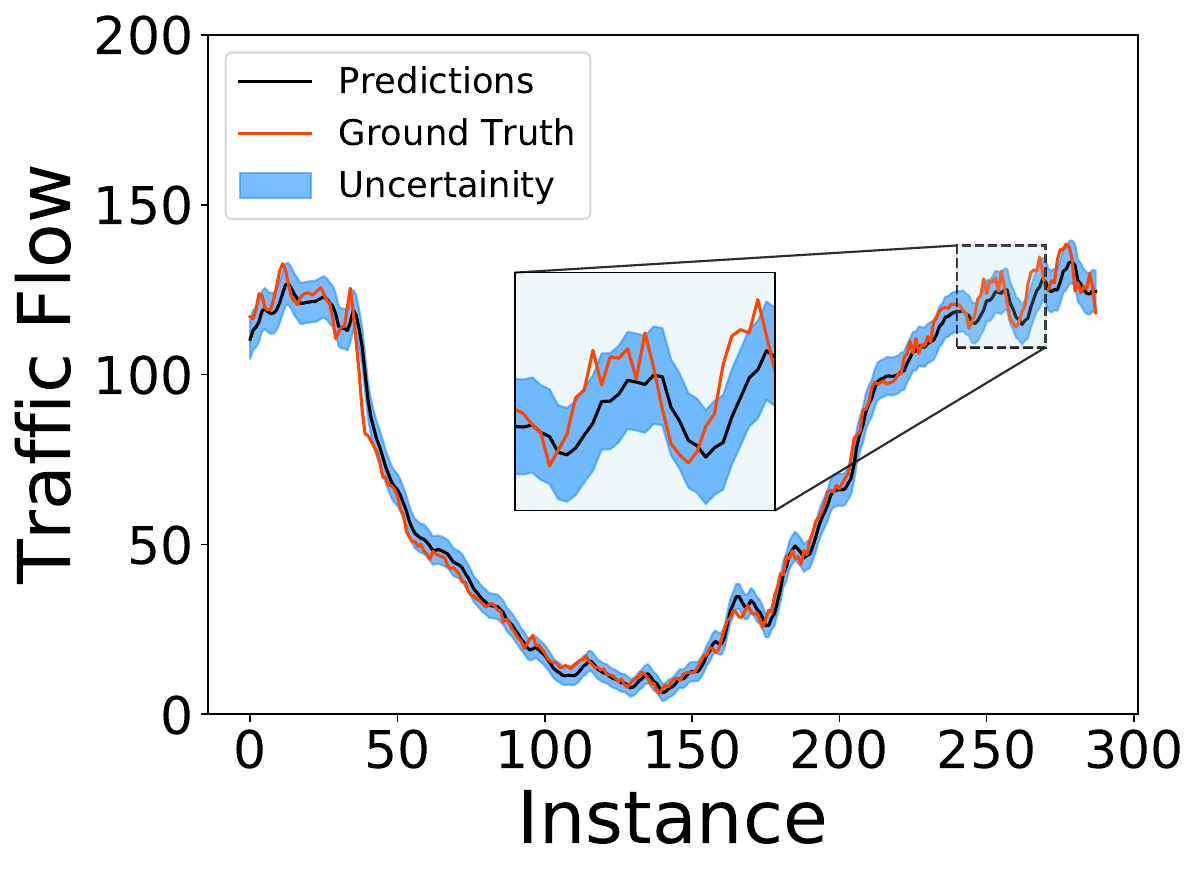}
}
\subfloat[Node 155 in PeMSD8]{
 \includegraphics[width=45mm]{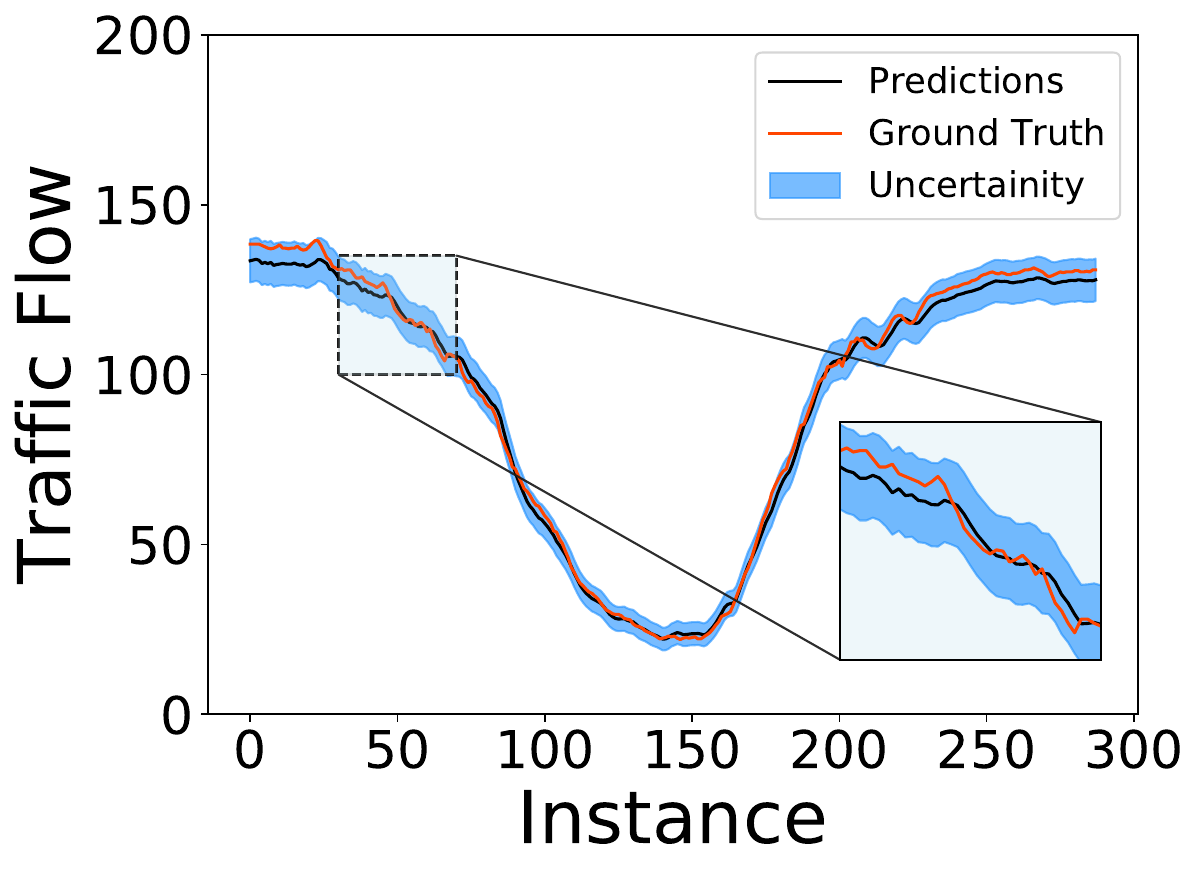}
}
\subfloat[Node 162 in PeMSD8]{
 \includegraphics[width=45mm]{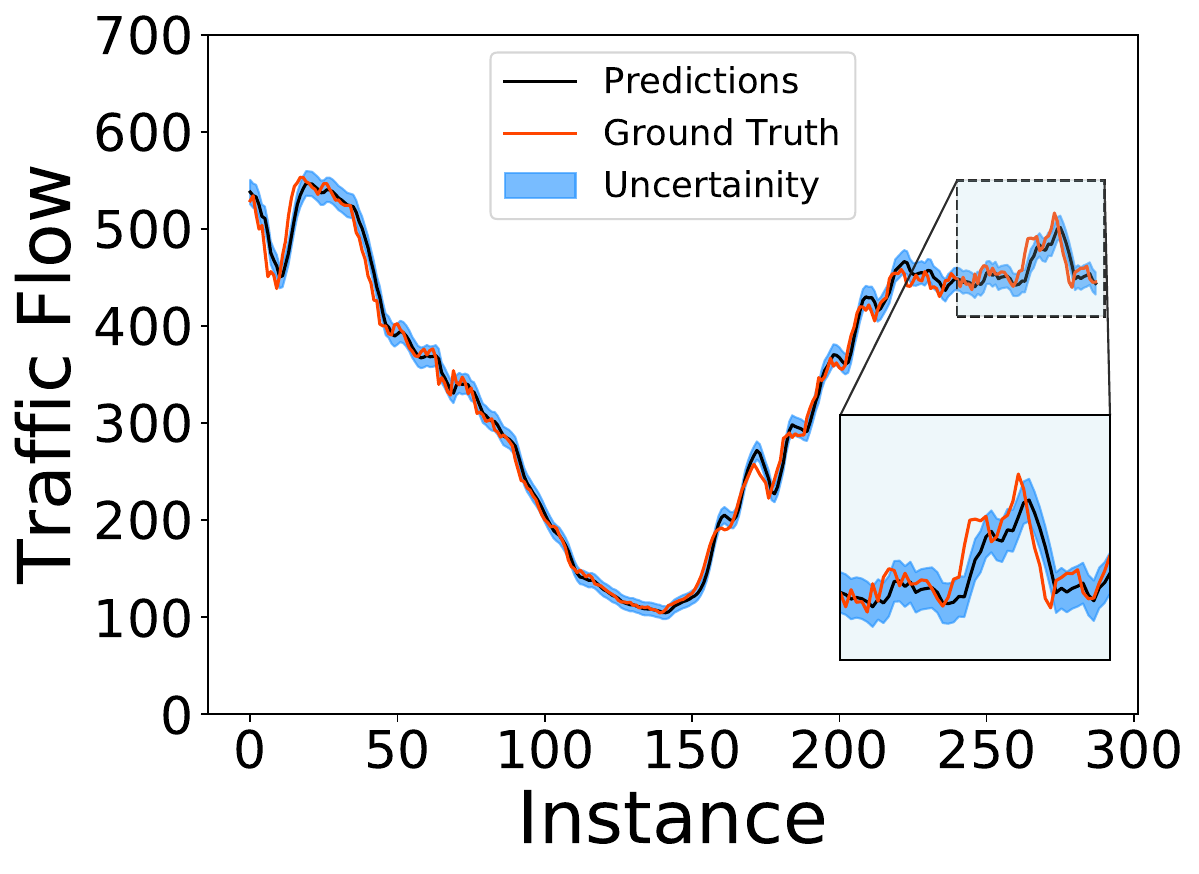}
}\\[-1.5ex]
\hspace{0mm}
\subfloat[Node 37 in METR-LA]{
 \includegraphics[width=45mm]{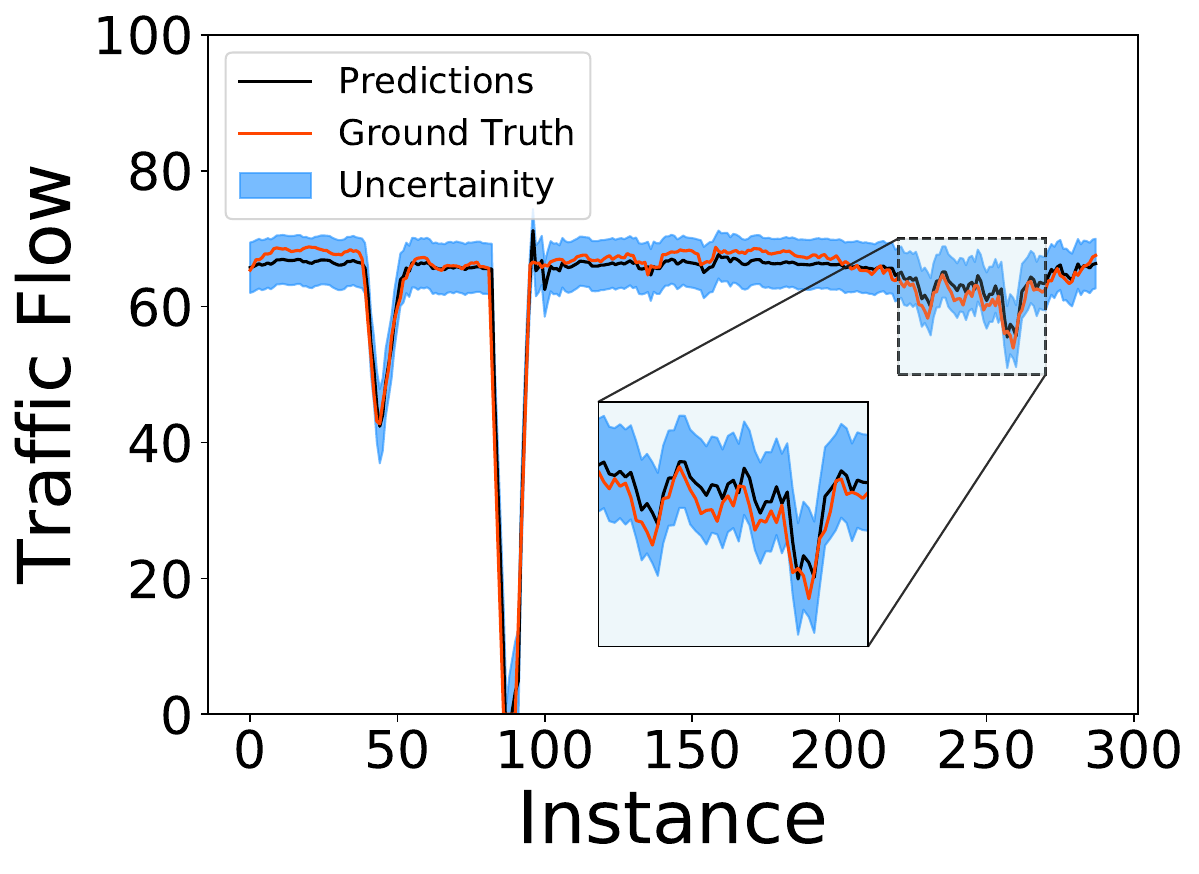}
}
\subfloat[Node 88 in METR-LA]{
 \includegraphics[width=45mm]{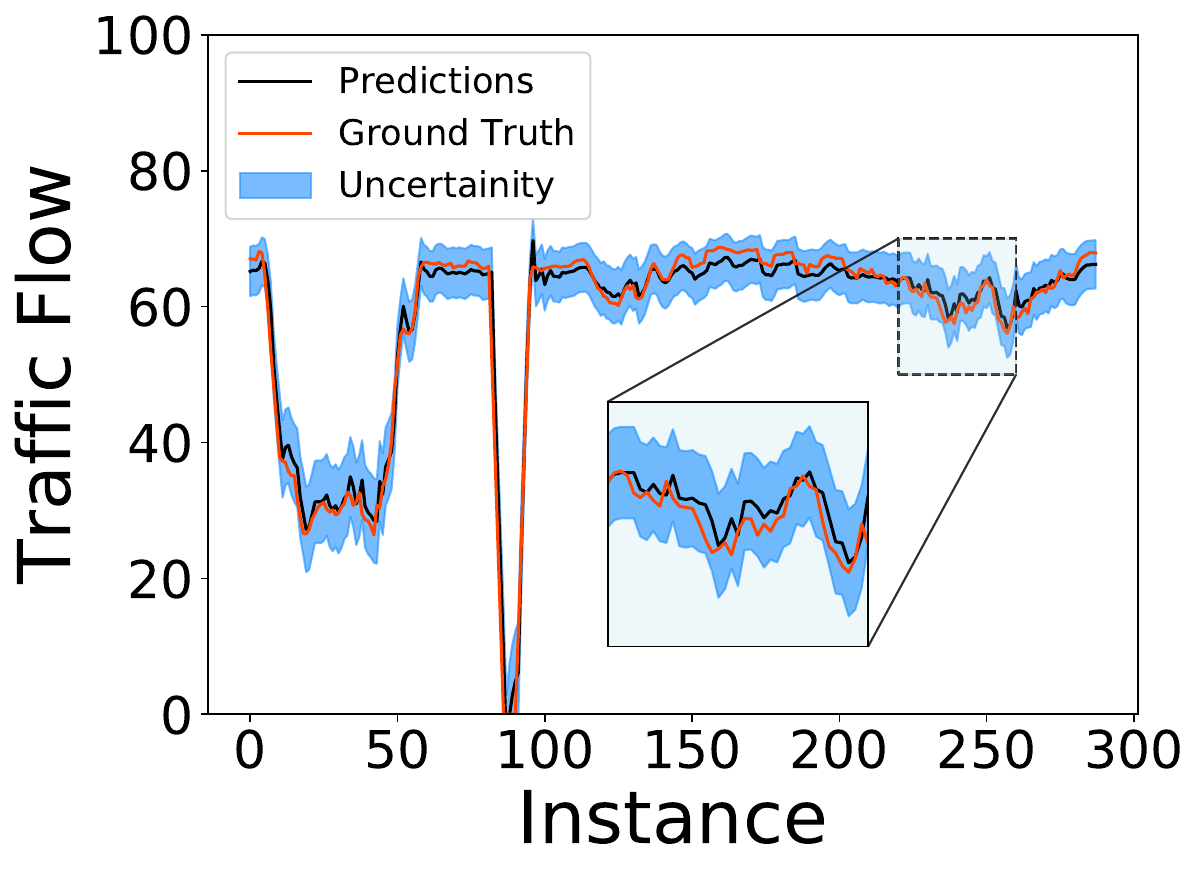}
}
\subfloat[Node 137 in METR-LA]{
 \includegraphics[width=45mm]{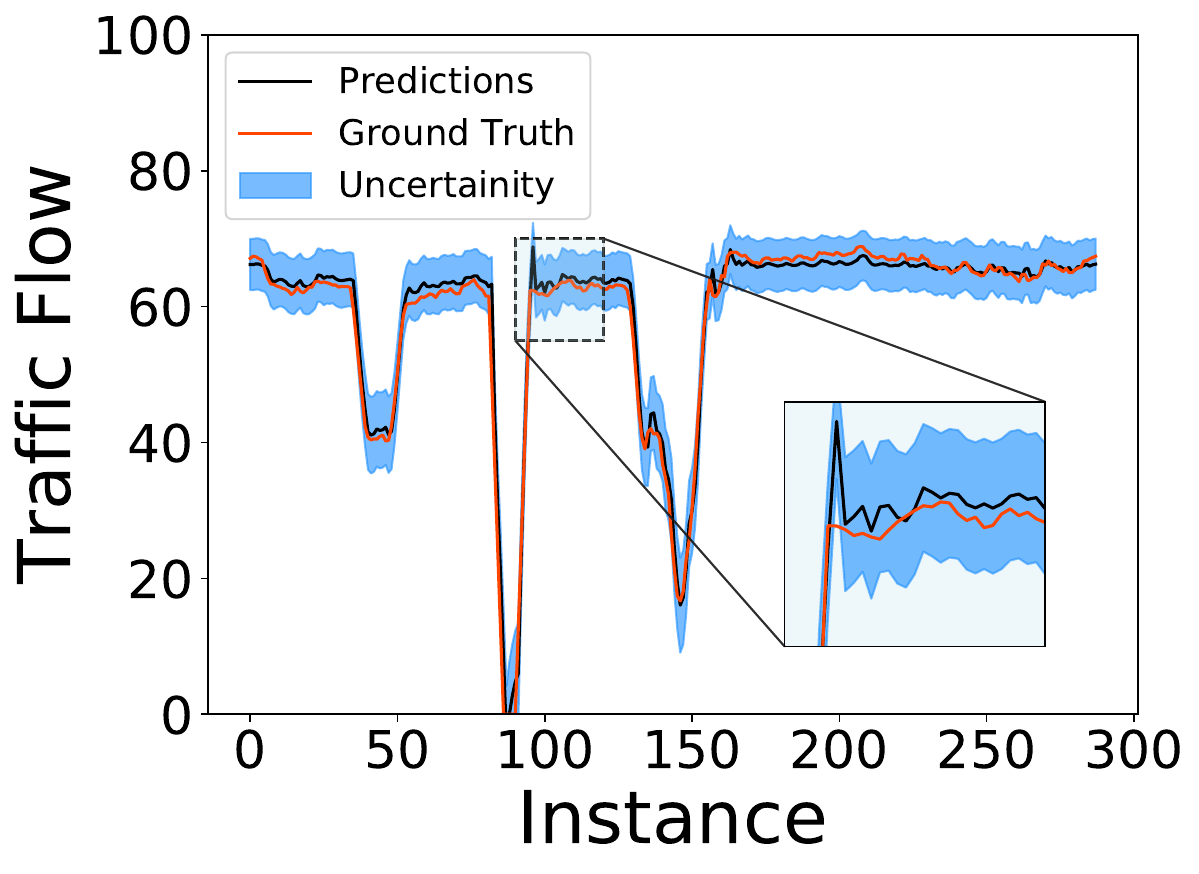}
}
\subfloat[Node 189 in METR-LA]{
 \includegraphics[width=45mm]{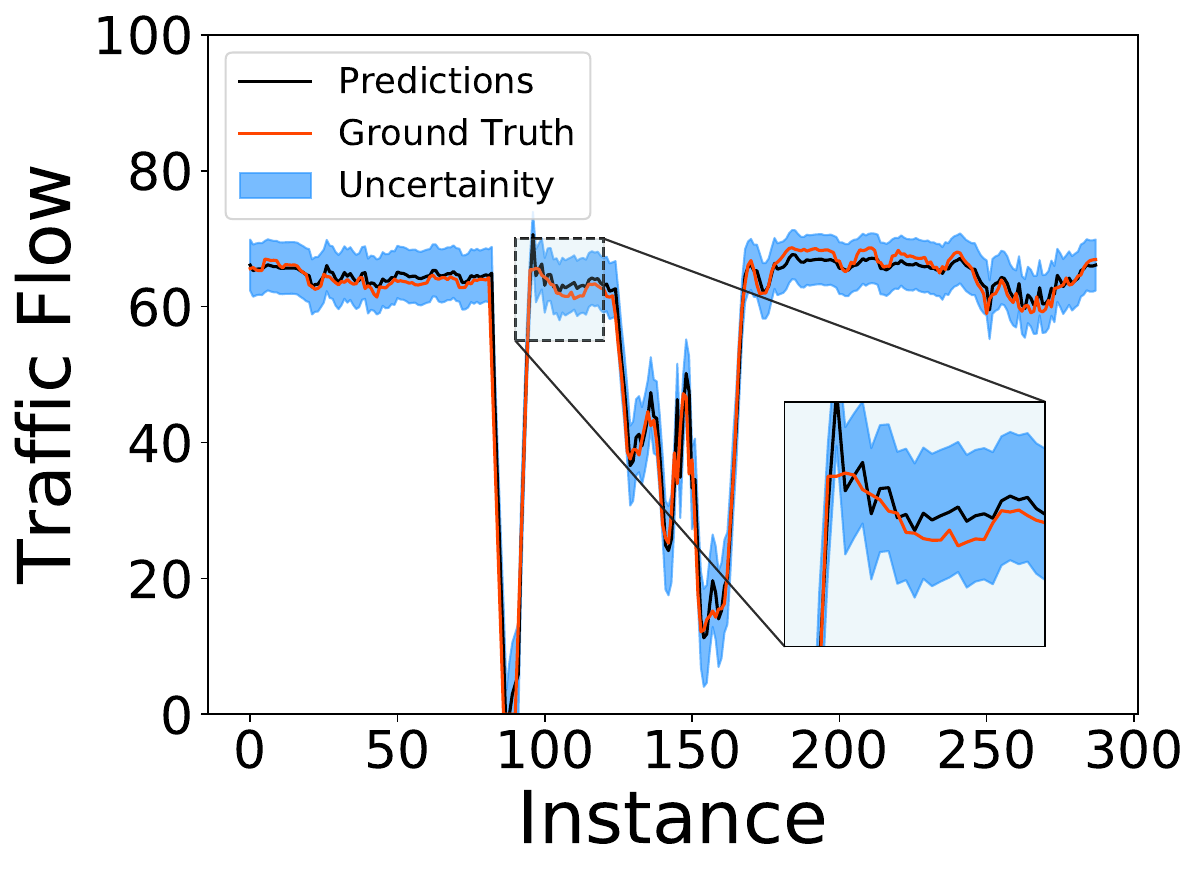}
}\\[-1.5ex]
\hspace{0mm}
\subfloat[Node 58 in PeMS-BAY]{
 \includegraphics[width=45mm]{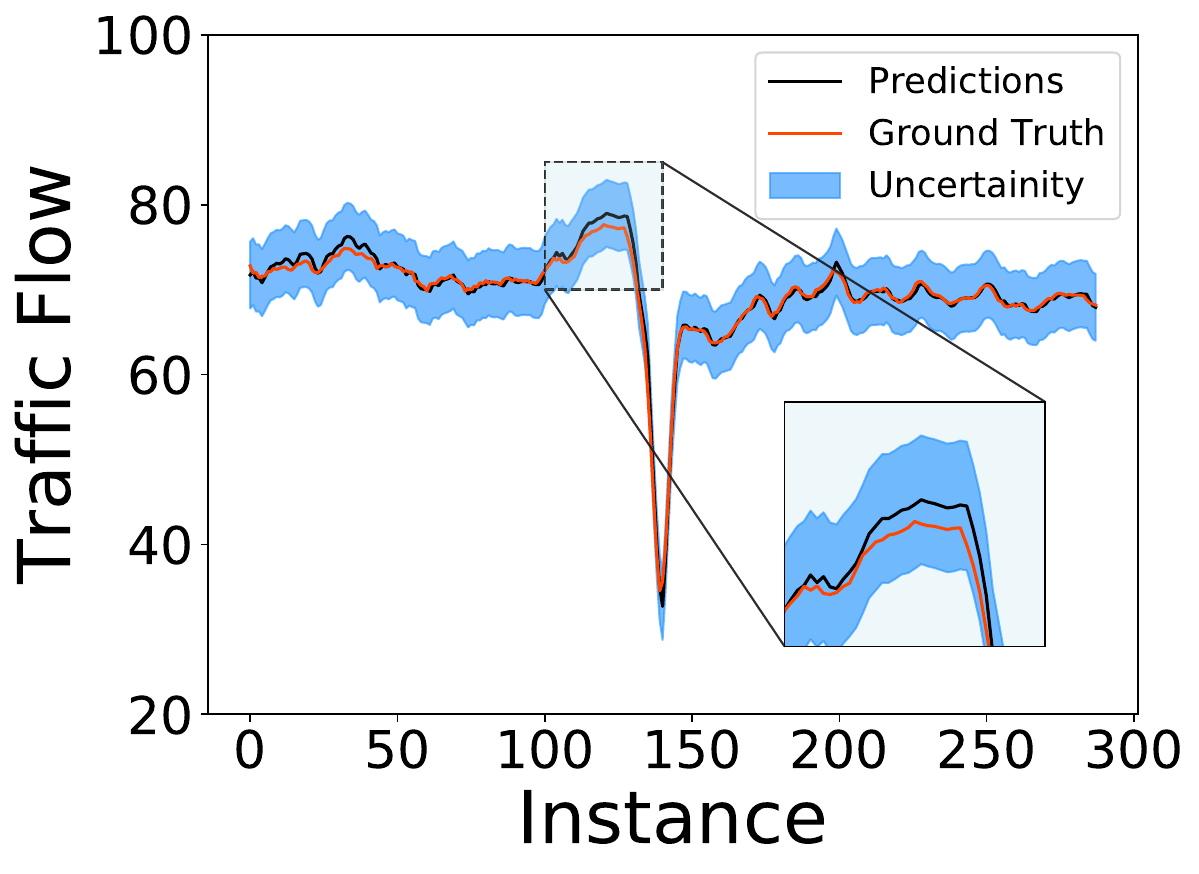}
}
\subfloat[Node 155 in PeMS-BAY]{
 \includegraphics[width=45mm]{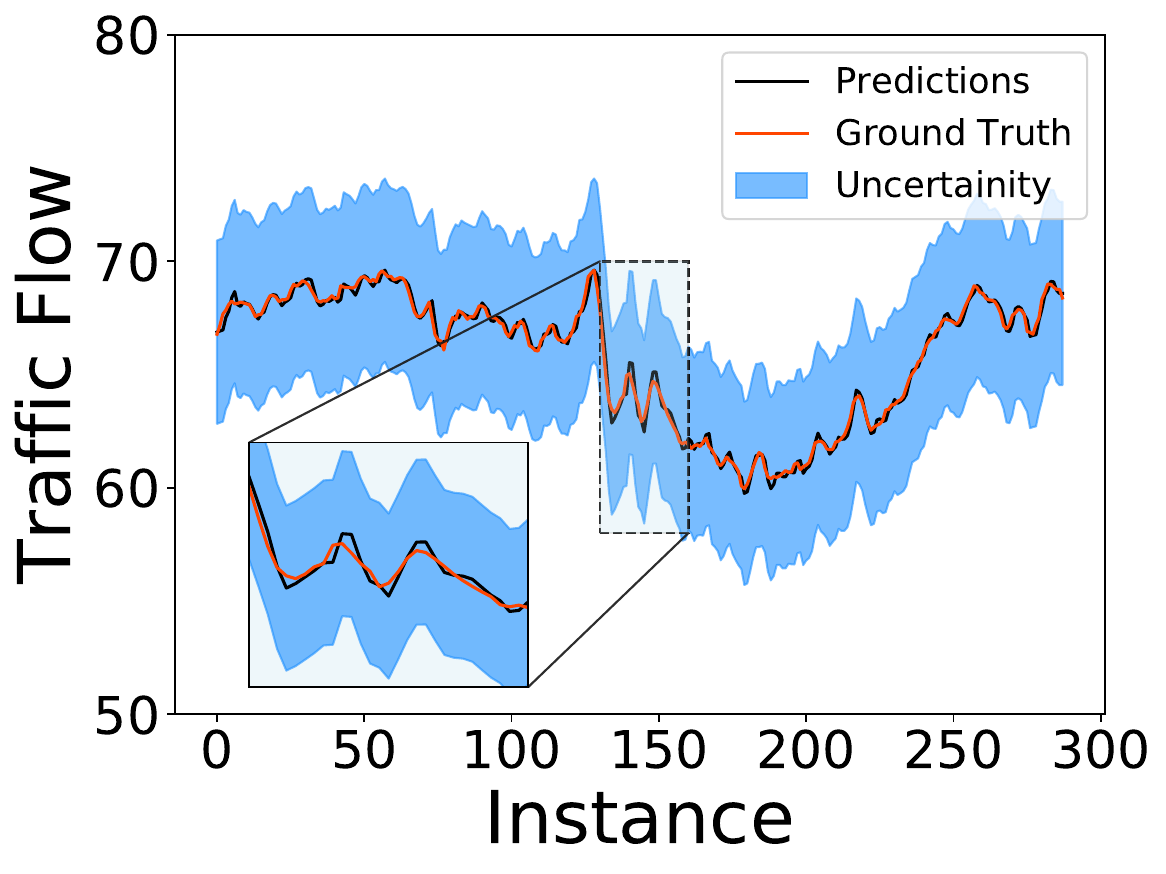}
}
\subfloat[Node 237 in PeMS-BAY]{
 \includegraphics[width=45mm]{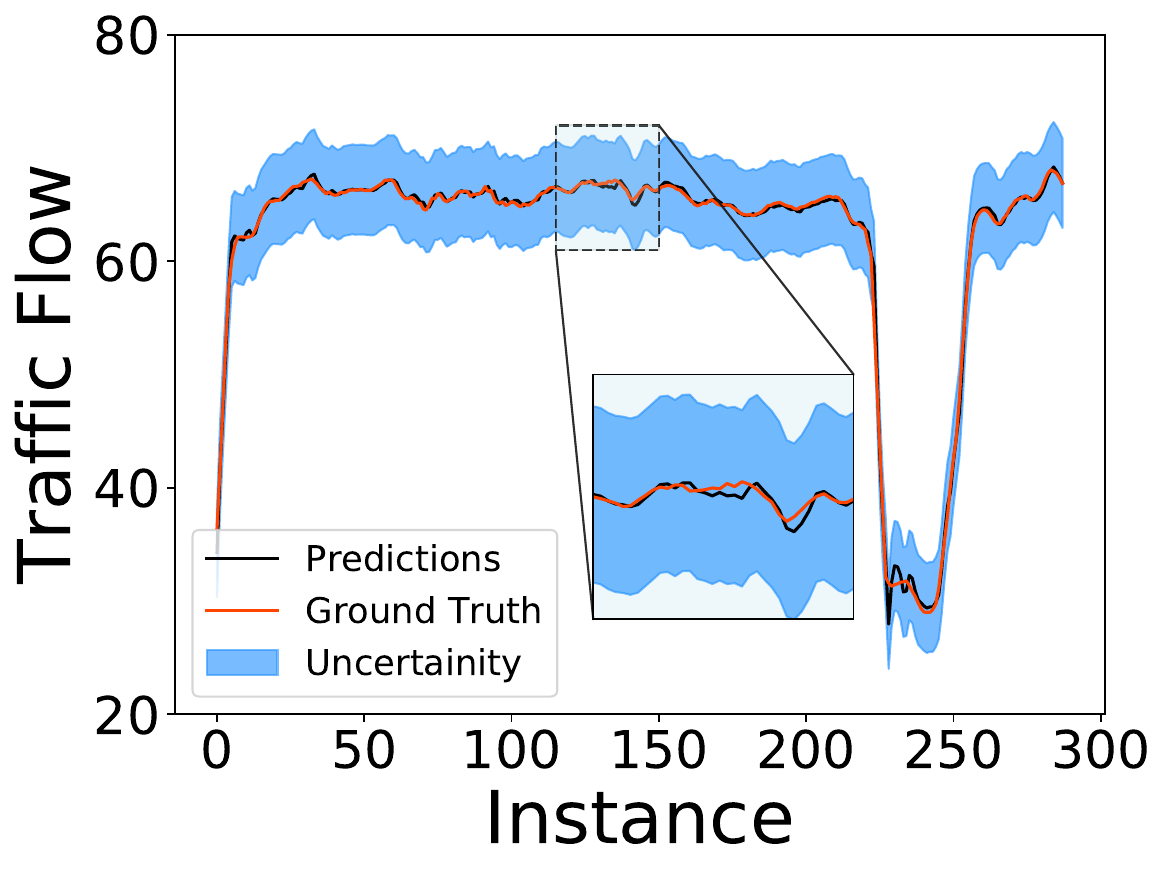}
}
\subfloat[Node 300 in PeMS-BAY]{
 \includegraphics[width=45mm]{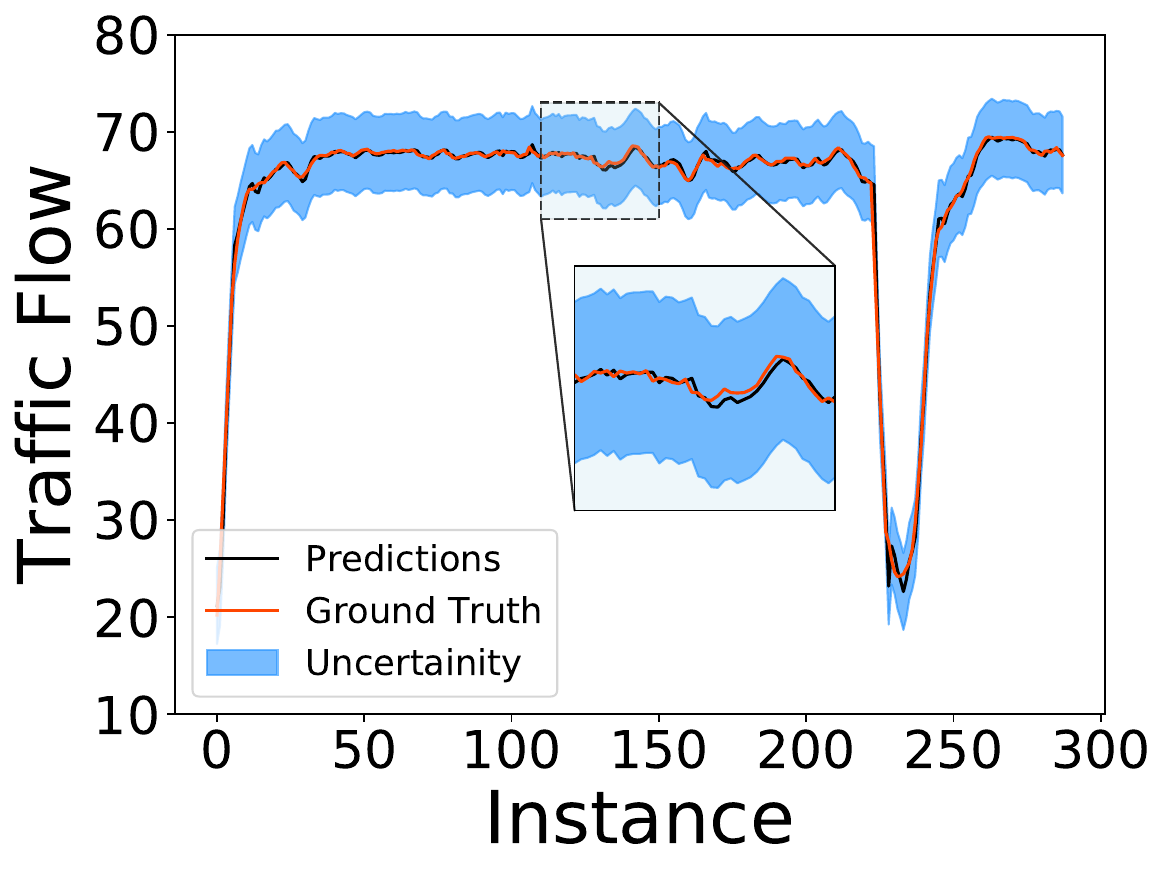}
}
\caption[Short Caption]{The figure provides a visualization of traffic forecasting results on benchmark datasets.}
\label{fig-traffic}
\end{figure*}

Figure \ref{fig-traffic} provides a visual depiction of the ground truth, pointwise forecasts, and time-varying uncertainty estimates obtained from the proposed \textbf{w/Unc- MKH-Net} framework. This representation offers valuable insights into the framework performance, facilitating a comprehensive analysis and interpretation of the results. While existing methods for multivariate time series forecasting can model nonlinear spatial-temporal dependencies within interconnected sensor networks, they typically fall short in providing accurate measures of uncertainty for multi-horizon forecasts. In contrast, the proposed \textbf{w/Unc- MKH-Net} framework(\textbf{MKH-Net} with local uncertainty estimation) effectively utilizes relational inductive bias through the spatio-temporal propagation architecture to quantitatively estimate uncertainty of model predictions. This framework is successful in accurately estimating uncertainty, as demonstrated through multifaceted visualizations, providing a significant improvement over existing methods that only provide pointwise forecasts for MTSF. The results highlight the effectiveness of the \textbf{w/Unc- MKH-Net} framework in time series representation learning for the MTSF task, making it a valuable addition to the field of multivariate time series analysis for uncertainty estimation.

\subsection{FORECASTING UNCERTAINITY}
The \textbf{MKH-Net} framework presents a supervised learning algorithm that employs the mean absolute error(MAE) as its loss function for model training. The MAE error is calculated by comparing the pointwise forecasts of the model predictions(denoted as \resizebox{.085\textwidth}{!}{$\hat{\mathbf{X}}_{(t  : t + \upsilon-1)}$}) to the corresponding ground-truth data (denoted as \resizebox{.085\textwidth}{!}{$\mathbf{X}_{(t  : t + \upsilon-1)}$}), as expressed in the equation below:

\vspace{-3mm}
\resizebox{0.935\linewidth}{!}{
\begin{minipage}{\linewidth}
\begin{align}
\mathcal{L}_{\text{MAE}}\left(\theta\right) =\frac{1}{\upsilon}\left|\mathbf{X}_{(t  : t + \upsilon-1)}-\hat{\mathbf{X}}_{(t  : t + \upsilon-1)}\right| \label{eq:UCE} \nonumber
\end{align} \nonumber
\end{minipage}
}

\vspace{1mm}
The training process strives to minimize the MAE loss function, denoted as $\mathcal{L}_{\text{MAE}}\left(\theta\right)$, through the fine-tuning of the model parameters $\theta$. The \textbf{w/Unc- MKH-Net} serves as a variant of the \textbf{MKH-Net}, specifically designed to evaluate the uncertainty in model predictions, thereby facilitating more 
trustability for decision-making in real-world applications. The \textbf{w/Unc- MKH-Net} framework offers a robust approach for predicting time-varying uncertainty in multistep-ahead forecasts, enhancing the overall reliability of the model's output. The forecasted predictions, represented as \resizebox{.085\textwidth}{!}{$\hat{\mathbf{X}}_{(t  : t + \upsilon-1)}$}, are characterized by a heteroscedastic Gaussian distribution with mean and variance denoted by \resizebox{.12\textwidth}{!}{$\mu_\phi\big(\bar{\mathbf{X}}_{(t  : t + \upsilon-1)}\big)$} and \resizebox{.12\textwidth}{!}{$\sigma_\phi^2\big(\bar{\mathbf{X}}_{(t  : t + \upsilon-1)}\big)$}, respectively. Here, \resizebox{.075\textwidth}{!}{$\bar{\mathbf{X}}_{(t  : t + \upsilon-1)}$}  refers to the input time series obtained from the projection layer. It is mathematically described as follows, 

\vspace{-1mm}
\resizebox{0.93\linewidth}{!}{
\begin{minipage}{\linewidth}
\begin{align}
\hat{\mathbf{X}}_{(t  : t + \upsilon-1)} \hspace{0.5mm} \sim \hspace{1mm}\mathcal{N}\big(\mu_\phi\big(\bar{\mathbf{X}}_{(t  : t + \upsilon-1)}\big), \sigma_\phi^2\big(\bar{\mathbf{X}}_{(t  : t + \upsilon-1)}\big)\big) \nonumber
\end{align}  
\end{minipage} \nonumber
}

\vspace{1mm}
The predicted mean and standard deviation of the heteroscedastic Gaussian distribution are obtained from the following equation,

\vspace{-1mm}
\resizebox{0.945\linewidth}{!}{
\begin{minipage}{\linewidth}
\begin{align}
\mu_\phi\big(\bar{\mathbf{X}}_{(t  : t + \upsilon-1)}\big), \sigma_\phi^2\big(\bar{\mathbf{X}}_{(t  : t + \upsilon-1)}\big) &= f_\theta\big( \mathbf{H}_{t  : t + \upsilon-1}\big) \nonumber
\end{align} 
\end{minipage}
}

\vspace{2mm}
The neural network, denoted as $f_{\theta}$, receives the output of the temporal inference component, represented by  \resizebox{.07\textwidth}{!}{$\mathbf{H}_{t  : t + \upsilon-1}$}, as its input. Utilizing this input, the network predicts the mean \resizebox{.12\textwidth}{!}{$\big(\mu_\phi\big(\bar{\mathbf{X}}_{(t  : t + \upsilon-1)}\big)\big)$} and standard deviation \resizebox{.12\textwidth}{!}{$\big(\sigma_\phi^2\big(\bar{\mathbf{X}}_{(t  : t + \upsilon-1)}\big)\big)$} of a normal distribution for future observations, symbolized as \resizebox{.08\textwidth}{!}{$\hat{\mathbf{X}}_{(t  : t + \upsilon-1)}$}. The maximum likelihood estimate(MLE) of the predicted Gaussian distribution is represented by $\hat{\mathbf{X}}_{(t : t + \upsilon-1)}$. It is mathematically described as follows,

\vspace{-2mm}
\resizebox{0.965\linewidth}{!}{
\begin{minipage}{\linewidth}
\begin{align}
\hat{\mathbf{X}}_{(t  : t + \upsilon-1)} = \mu_\phi\big(\bar{\mathbf{X}}_{(t  : t + \upsilon-1)}\big) \nonumber
\end{align}  
\end{minipage}
}

\vspace{2mm}
In short, \resizebox{.12\textwidth}{!}{$\mu_\phi\big(\bar{\mathbf{X}}_{(t  : t + \upsilon-1)}\big)$} provides an estimation of the expected value of the Gaussian distribution of future values, i.e., \resizebox{.085\textwidth}{!}{$\hat{\mathbf{X}}_{(t  : t + \upsilon-1)}$}, given the observed values from time points \resizebox{.1\textwidth}{!}{$t - \tau$ to $t-1$}, i.e., \resizebox{.085\textwidth}{!}{$\bar{\mathbf{X}}_{(t  : t + \upsilon-1)}$}. Furthermore, \resizebox{.12\textwidth}{!}{$\sigma_\phi^2\big(\bar{\mathbf{X}}_{(t  : t + \upsilon-1)}\big)$}  represents the model's prediction uncertainty for the subsequent $\upsilon$ time steps, starting from the current $t^{th}$ time point. The regular Gaussian likelihood of future values, given parameters \resizebox{.12\textwidth}{!}{$\mu_\phi\big(\bar{\mathbf{X}}_{(t  : t + \upsilon-1)}\big)$} and \resizebox{.12\textwidth}{!}{$\sigma_\phi^2\big(\bar{\mathbf{X}}_{(t  : t + \upsilon-1)}\big)$} described as follows:

\vspace{-2mm}
\resizebox{0.895\linewidth}{!}{
\begin{minipage}{\linewidth}
\begin{multline}
\mathcal{N}(\hat{\mathbf{X}}_{(t  : t + \upsilon-1)};\mu_\phi\big(\bar{\mathbf{X}}_{(t - \tau : \hspace{1mm}t-1)}\big),\sigma_\phi\big(\bar{\mathbf{X}}_{(t - \tau : \hspace{1mm}t-1)}\big))= 
\\
\\ {\frac {1}{\sigma_\phi\big(\bar{\mathbf{X}}_{(t - \tau : \hspace{1mm}t-1)}\big) {\sqrt {2\pi }}}}\hspace{1.5mm} e^{-{\dfrac {1}{2}}\left({\dfrac {\mathbf{X}_{(t  : t + \upsilon-1)}-\mu_\phi\big(\bar{\mathbf{X}}_{(t - \tau : \hspace{1mm}t-1)}\big) }{\sigma_\phi\big(\bar{\mathbf{X}}_{(t - \tau : \hspace{1mm}t-1)}\big)}}\right)^{2}}  \nonumber
\end{multline}  \nonumber
\end{minipage} \nonumber
}

\vspace{2mm}
The uncertainty modeling framework aims to optimize the Gaussian negative log likelihood loss(i.e., negative log gaussian probability density function(pdf)), as described in \cite{nix1994estimating}. This optimization process relies on the mean and variance estimates of the model and forms a solid foundation for understanding and quantifying the inherent uncertainty in the model's predictions. We apply logarithm transformation on both sides of the equation,  it is described as follows,  

\vspace{-2mm}
\resizebox{0.91\linewidth}{!}{
\begin{minipage}{\linewidth}
\begin{multline}
\log\ \mathcal{N}(\hat{\mathbf{X}}_{(t  : t + \upsilon-1)};\mu_\phi\big(\bar{\mathbf{X}}_{(t  : t + \upsilon-1)}\big),\sigma_\phi\big(\bar{\mathbf{X}}_{(t  : t + \upsilon-1)}\big)) 
\\ = \log\left[{\frac {1}{\sigma_\phi\big(\bar{\mathbf{X}}_{(t  : t + \upsilon-1)}\big) {\sqrt {2\pi }}}}\right] + \log \left[e^{-{\dfrac {1}{2}}\left({\dfrac {\mathbf{X}_{(t  : t + \upsilon-1)}-\mu_\phi\big(\bar{\mathbf{X}}_{(t  : t + \upsilon-1)}\big) }{\sigma_\phi\big(\bar{\mathbf{X}}_{(t  : t + \upsilon-1)}\big)}}\right)^{2}}\right] \\
 = \log\ {\frac {1}{\sigma_\phi\big(\bar{\mathbf{X}}_{(t  : t + \upsilon-1)}\big)}} + \log\ {\frac {1}{{\sqrt {2\pi }}}} -{\frac {1}{2}}\left(\dfrac {\mathbf{X}_{(t  : t + \upsilon-1)}-\mu_\phi\big(\bar{\mathbf{X}}_{(t  : t + \upsilon-1)}\big) }{\sigma_\phi\big(\bar{\mathbf{X}}_{(t - \tau : \hspace{1mm}t-1)}\big)}\right)^{2} \\
 = -\log\ \sigma_\phi\big(\bar{\mathbf{X}}_{(t  : t + \upsilon-1)}\big) + C -{\frac {1}{2}}\left(\dfrac {\mathbf{X}_{(t  : t + \upsilon-1)}-\mu_\phi\big(\bar{\mathbf{X}}_{(t  : t + \upsilon-1)}\big) }{\sigma_\phi\big(\bar{\mathbf{X}}_{(t  : t + \upsilon-1)}\big)}\right)^{2}  \nonumber
\end{multline}   
\end{minipage} \nonumber
}

\vspace{1mm}
We drop the constant(C) and the Gaussian negative log likelihood loss is described by, 

\vspace{-3mm}
\resizebox{0.90\linewidth}{!}{
\hspace{-8mm}\begin{minipage}{\linewidth}
\begin{multline}
\mathcal{L}_{\text{GaussianNLLLoss}} = 
\\
\\ \sum_{t=1}^{\text{T}_{train}} \left[\frac{\log \sigma_\phi\big(\bar{\mathbf{X}}_{(t  : t + \upsilon-1)}\big)^2}{2}+\frac{\left(\mathbf{X}_{(t  : t + \upsilon-1)}-\mu_\phi\left(\bar{\mathbf{X}}_{(t  : t + \upsilon-1)}\right)\right)^2}{2 \sigma_\phi\left(\bar{\mathbf{X}}_{(t  : t + \upsilon-1)}\right)^2}\right] \nonumber
\end{multline}   
\end{minipage} \nonumber
}

\vspace{3mm}
Where $\text{T}_{train}$ denotes the time points in the training set. The negative Gaussian log-likelihood quantifies the likelihood of the observations, considering the estimated mean and variance of the Gaussian distribution. A lower negative Gaussian log-likelihood value signifies a better fit of the Gaussian distribution to the observed values, thus indicating a more accurate representation of the underlying data.  The  \textbf{w/Unc- MKH-Net} framework, utilizes a Gaussian likelihood function for modeling the future values, and the mean and variance of the gaussian distribution are modeled by the neural network. The mean is the predicted value, and the variance represents the uncertainty of the predictions. By minimizing the negative log-likelihood of the Gaussian distribution, we determine the set of model parameters that provide the best fit to the data and also provides the uncertainty estimates. To put it briefly, the \textbf{MKH-Net} framework minimizes the LMSE(Least Mean Square Error) to find the set of model parameters that best fit the data. On the other hand, the \textbf{w/Unc- MKH-Net} framework(\textbf{MKH-Net} with local uncertainty estimation) minimizes the $\textbf{L}_\textbf{GaussianNLLLoss}$(negative log-likelihood of a Gaussian distribution) to estimate uncertainty quantitatively.

\newpage
\vspace{-1mm}
\subsection{DATASETS}
\vspace{-2mm}

\begin{table}[htbp]
\center
\setlength{\tabcolsep}{0.25em} 
\renewcommand\arraystretch{1.15} 
\centering
\hspace*{-10mm}
 \resizebox{0.55\textwidth}{!}{
\begin{tabular}{c|c|c|c|c|c}
\hline
\textbf{Dataset} & \textbf{Variables} & \textbf{Timepoints} & \textbf{Time-Range} & \multicolumn{1}{l|}{\textbf{Split-Ratio}} & \multicolumn{1}{l}{\textbf{Granularity}} \\ \hline
PeMSD3 & 358 & 26,208 & 09/2018 - 11/2018 & \multirow{5}{*}{6 / 2 / 2} & \multirow{7}{*}{\rotatebox[origin=c]{270}{5 mins}} \\
PeMSD4 & 307 & 16,992 & 01/2018 - 02/2018 &  &  \\
PeMSD7 & 883 & 28,224 & 05/2017 - 08/2017 &  &  \\
PeMSD8 & 170 & 17,856 & 07/2016 - 08/2016 &  &  \\
PeMSD7(M) & 228 & 12,672 & 05/2012 - 06/2012 &  &  \\ \cline{1-5}
METR-LA & 207 & 34,272 & 03/2012 - 06/2012 & \multirow{2}{*}{7 / 1 / 2} &  \\
PEMS-BAY & 325 & 52,116 & 01/2017 - 05/2017 &  &  \\ \hline
\end{tabular}
}
\vspace{-2mm}
\caption{Summary of the various traffic-related benchmark datasets.}
\label{tab:summarydatasets}
\end{table}

\vspace{-2mm}
We evaluated the effectiveness of our novel \textbf{MKH-Net} framework by comparing it to existing benchmark models on various real-world datasets. These datasets include PeMSD3, PeMSD4, PeMSD7, PeMSD7(M), PeMSD8, METR-LA, and PEMS-BAY. To generate the pre-defined graphs(explicit graphs) for traffic-related benchmark datasets, kernel-based similarity metric functions were used, based on the distance between sensors on the road network(\cite{choi2022graph}, \cite{LiYS018}). Table \ref{tab:summarydatasets} provides more details of the benchmark datasets.

\vspace{-1mm}
\subsection{EXPERIMENTAL SETUP}
The traffic-related benchmark datasets were divided into three mutually exclusive subsets with different ratios: a training set for model learning, a validation set for hyperparameter tuning, and a test set to evaluate model performance on unseen data. The PEMS-BAY and METR-LA datasets were split with a 7/1/2 ratio, while all other datasets were split with a 6/2/2 ratio into training, validation, and test sets, respectively. The time-series data preprocessing step involved scaling each time-series variable to have zero mean and unit variance. Various forecast accuracy metrics such as MAE(Mean Absolute Error), RMSE(Root Mean Squared Error), and MAPE(Mean Absolute Percentage Error) were computed on the original scale of the MTS data during the training and evaluation of forecasting models. The \textbf{MKH-Net} framework was trained for a predefined number of 30 epochs to update the trainable parameters of the model iteratively, with the goal of minimizing the forecast error. We used Validation MAE for early stopping to prevent overfitting. This approach helps to avoid suboptimal solutions and select the best model that can generalize well to unseen data. As a result, the overall performance of the model is improved. The \textbf{MKH-Net} framework performance was evaluated on the test set to estimate its ability to generalize to new and unseen data. To optimize the learning process, a learning rate scheduler was utilized, which reduced the learning rate by half after 5 epochs of no improvement in the validation set performance. The Adam optimizer was used to train the model to fine-tune the learnable parameters and improve convergence. An initial learning rate of \num{1e-3} was set to minimize the MAE loss for the \textbf{MKH-Net} model and the negative Gaussian log-likelihood for the \textbf{w/Unc- MKH-Net} model between the ground truth and the model predictions. The models were trained on powerful GPUs, including the NVIDIA Tesla T4, Nvidia Tesla v100, and GeForce RTX 2080 GPUs, to speed up the training process and enable the use of larger models and datasets built upon the PyTorch framework. Multiple independent experimental runs were performed, and the ensemble average was reported to provide a reliable evaluation of the models performance. The hyperparameters of the learning algorithm were the embedding size($\textit{d}$), number of hyperedges($|\mathcal{HE}|$), batch size($\textit{b}$), and learning rate($\textit{lr}$). The optimum hyperparameter values for each dataset were reported in Section \ref{sensitivityanalysis}.

\subsection{BASELINES}
In the context of evaluating the performance of the proposed neural forecasting models(\textbf{MKH-Net} and \textbf{w/Unc-MKH-Net}) on the MTSF(multivariate time series forecasting) task, it is customary to use widely recognized algorithms as benchmarks. These benchmark algorithms are usually chosen based on their widespread use in the literature and their established performance on benchmark datasets.

\vspace{0mm}
\begin{itemize}
\item HA~\cite{hamilton2020time}, involves using the average of a predefined historical window of observations to predict the next value in a time series. 
\item ARIMA is a statistical analysis model for handling the non-stationary time series data. However, they do have some limitations, such as difficulty in handling long-term trends or seasonal patterns that change over time.
\item VAR(\cite{hamilton2020time}) is an extension of the univariate autoregressive(AR) model, designed to capture the inter-dependencies among multiple time series variables, enabling the analysis and forecasting of complex systems.
\item TCN(~\cite{BaiTCN2018}) architecture comprises a stack of causal convolutions and dilation layers for handling sequential data in multistep-ahead time series prediction tasks. The use of causal convolutions allows TCN to incorporate past information into the model, making it effective for handling time series data. The dilation layers increase the receptive field of the convolutional filters exponentially, allowing the model to learn long-range correlations in the MTS data.
\item FC-LSTM(~\cite{sutskever2014sequence}) framework, which is an encoder-decoder architecture based on LSTM units with a peephole connection has been an effective technique for multi-horizon forecasting task. It captures the complex and nonlinear relationships between multiple time series variables in MTS data, including both short-term and long-term relationships.
\item GRU-ED(~\cite{cho2014grued}) an encoder-decoder architecture based on GRU units is an effective framework for handling sequential data for multistep-ahead time series prediction by capturing relevant information from previous time steps.
\item DSANet(~\cite{Huang2019DSANet}) is a correlated time series prediction model that uses CNN networks to capture long-range intra-temporal dependencies of multiple time series and does not rely on recurrence to capture temporal dependencies in MTS data. It also utilizes self-attention blocks that can adaptively capture inter-dependencies for predicting multiple-steps-ahead forecasts in the MTS data.
\item DCRNN(~\cite{li2018dcrnn_traffic}), is a powerful technique that leverages bidirectional random walks on graphs and integrates graph convolution with recurrent neural networks to capture the spatial-temporal dependencies in the MTS data. This unique approach enables the prediction of multistep-ahead forecasts in MTS data through an encoder-decoder architecture, with an improved forecast accuracy than conventional approaches. 
\item STGCN(~\cite{bing2018stgcn}) combines the graph convolution and gated temporal convolution to effectively capture the spatial-temporal correlations between multiple time series variables for predicting multistep-ahead forecasts in MTS data.
\item GraphWaveNet(~\cite{wu2019graphwavenet}), is a framework that jointly learns an adaptive dependency matrix using a wave-based propagation mechanism and graph representations with dilated causal convolution layers to capture spatial-temporal dependencies for predicting multistep-ahead forecasts in the time series. This approach allows the model to effectively capture dependencies between multiple time series variables, resulting in imroved forecast accuracy.
\item ASTGCN(~\cite{guo2019astgcn}), utilizes an attention-based spatio-temporal graph convolutional network to capture inter- and intra-dependencies for predicting multiple-steps-ahead forecasts in the time series. This approach allows the model to effectively capture spatial-temporal relationships between multiple time series variable using attention mechanisms. 
\item STG2Seq(~\cite{bai2019STG2Seq}) predicts multistep-ahead forecasts in MTS data by using a combination of gated graph convolutional networks(GGCNs) and a sequence-to-sequence(seq2seq) architecture with attention mechanisms. This approach captures dynamic temporal correlations within the time series variables, as well as the cross-channel information (correlations among multiple variables), to effectively model the complex relationships between multiple time series variables.
\item STSGCN(~\cite{song2020stsgcn}) predicts multistep-ahead forecasts in MTS data by stacking multiple layers of spatial-temporal graph convolutional networks. It captures localized intra- and inter-dependencies in the graph-structured MTS data to model the complex relationships between multiple time series variables.
\item LSGCN(~\cite{huang2020lsgcn}) predicts multistep-ahead forecasts in MTS data by integrating a graph attention mechanism into a spatial gated block. This approach captures dynamic spatial-temporal dependencies among multiple time-series variables using attention mechanisms, resulting in significantly improved forecast accuracy.
\item AGCRN(~\cite{NEURIPS2020_ce1aad92}) predicts multistep-ahead forecasts in MTS data by using data-adaptive graph structure learning method. It captures node-specific intra- and inter-correlations to effectively model complex spatial-temporal dependencies and relationships between multiple time series variables.
\item STFGNN(~\cite{li2021stfgnn}) predicts multistep-ahead forecasts in MTS data by fusing representations obtained from the temporal graph module and gated convolution module. They operate for different time periods in parallel and learn spatial-temporal correlations to effectively model the complex relationships between multiple time series variables.
\item Z-GCNETs(~\cite{chen2021ZGCNET}) predicts multistep-ahead forecasts in MTS data by integrating a time-aware zigzag topological layer into time-conditioned graph convolutional networks(GCNs). It captures hidden spatial-temporal dependencies and learns salient time-conditioned topological information to effectively model the complex relationships between multiple time series variables while considering their topological properties.
\item STGODE(~\cite{fang2021STODE}) predicts multistep-ahead forecasts by using a tensor-based ordinary differential equation(ODE) to capture inter- and intra-dependency relationships between multiple time series variables in the MTS data. This approach allows for deeper networks to be constructed and captures the complex spatial-temporal dynamics of the data, resulting in significantly improved forecast accuracy.
\end{itemize}

\vspace{-2mm}
\subsection{RELATED WORK}
\vspace{0mm}
Multivariate time series forecasting is a critical task that has widespread applications in diverse domains for strategic decision-making. By predicting multiple variables that are related and evolve over time, it allows for informed decisions about strategy formulation, resource allocation, and other essential business processes. The manufacturing sector employs it for production planning, energy management and inventory management, while finance uses it for investment planning and risk management. The retail and e-commerce benefits from its use in supply chain management, product pricing, and so on. Its versatility makes it an indispensable tool for informed decision-making in modern businesses. The conventional time series forecasting techniques such as VAR(\cite{watson1994vector}), ARIMA(\cite{makridakis1997arma}), and state-space models(SSMs, \cite{hyndman2008forecasting}) have limitations in handling correlated MTS data with complex dependencies and non-linear relationships among multiple time series variables. Over the last few years, novel deep learning methods, such as FC-LSTM(\cite{sutskever2014sequence}) and TPA-LSTM(\cite{shih2019temporal}) have emerged as potential alternatives for  overcome the limitations of traditional statistical techniques in handling complex spatial-temporal dependencies in multivariate time series forecasting. These methods employ an implicit recurrent process to model intra-temporal correlations, enabling them to better capture the non-linear temporal dynamics of time series variables in MTS data. While these deep learning-based methods have shown promising results in handling complex temporal dependencies in multivariate time series forecasting, they may not explicitly account for the inter-correlations among different time series variables. Consequently, these models may face limitations in capturing the entire complexity of MTS data. Recent research has proposed various methods to address the limitations of deep learning-based methods in modeling inter-correlations among different time series variables in addition to intra-correlations within the variables. These methods include convolutional LSTM(ConvLSTM), recurrent attention networks(RAN), dynamic self-attention(DSA), temporal convolutional networks(TCN), and transformer-based architectures. In specific, LSTNet is a well-known hybrid method for multivariate time series forecasting, which utilizes a combination of convolutional and recurrent neural networks to capture the inter-dependencies among variables and intra-temporal dependencies within time series variables. However, the method is inefficient in precisely modeling the complex relationships and dependencies among the multiple variables of MTS data due to its global aggregation of pair-wise semantic relations. Hence, further advancements are necessary to address these limitations and enhance the accuracy of MTSF tasks. Recent advances in multivariate time series forecasting have embraced Graph neural networks(GNNs) as a promising approach for modeling the relational dependencies between time series variables in the MTS data. Researchers such as \cite{zonghanwu2019,Bai2020nips,wu2020connecting,YuYZ18,tampsgcnets2022} and  \cite{LiYS018}) have contributed to this growing trend, and the approach has become increasingly popular. The graph-based modeling approaches for MTSF require a predefined graph that represents the dependencies between variables as edges in the graph. GNNs then leverage this graph to predict the dynamics of multiple time series variables, enabling them to capture the complex relationships and dependencies within the MTS data. Despite the effectiveness of graph-based approaches in MTSF task, in most real-world scenarios, an explicit graph structure is often unavailable or inaccurate. Additionally, the use of a predefined graph structure for modeling the high-dimensional dynamics of interconnected systems may impose limitations, as it fails to encapsulate the full complexity of the underlying relational graph structure among variables, ultimately resulting in suboptimal forecasting. Additionally, the forecast models that rely on predefined graphs often assume a fixed relationship between variables. This approach neglects the potential to incorporate the relational graph structure that accounts for the dynamic relationships among variables underlying the spatio-temporal MTS data. To achieve more accurate predictions, it is crucial to develop models that can effectively capture the non-linear dynamic relationships and dependencies between variables. A novel category of GNN-based models has been proposed in recent research, that utilize a time-varying adaptive attention mechanism to model the complex dynamic correlations among variables within MTS data. Researchers such as \cite{shang2021discrete}, \cite{deng2021graph} and, \cite{wu2020connecting} have contributed to this growing field. These GNN-based models aim to jointly learn the implicit graph structure and representations of the spatio-temporal graph MTS data, instead of solely relying on predefined graphs. However, incorporating prior or domain knowledge of the interconnected sensor networks can further enhance performance on MTSF task in real-world scenarios when an explicit graph structure is completely available and reliable. Incorporating relevant prior knowledge can help to overcome the limitations of data-driven neural relational inference methods, which rely solely on the MTS data. This is particularly beneficial in scenarios when the data is limited or noisy, leading to improved forecast accuracy and robustness. Additionally, incorporating prior knowledge can help to constrain the search space, regularize the model, and improve its overall robustness. Nonetheless, the GNN-based approaches model the multi-hop relationships among time series variables, which can lead to over-smoothing and under-learning(over-squashing) issues. Additionally, existing higher-order graph convolution algorithms in literature or simply, modeling the graph as a simplicial complex are often ineffective in capturing the higher-order correlations in spatio-temporal graph data. Consequently, these models may encounter difficulties in capturing higher-order relationships among the time series variables, leading to potentially less accurate and reliable forecasts. Despite the promising results of deterministic GNN-based models in improving pointwise forecast accuracy, they often generate poor estimates of uncertainty. To address this issue, researchers have put forth a range of deep learning methods, such as Bayesian neural networks, Gaussian processes, ensemble learning techniques, bootstrapping, jackknife resampling methods, and post-hoc uncertainty evaluations. However, more research is needed to advance and improve GNN-based models to effectively estimate uncertainty for MTSF tasks. In summary, there is a need for interpretable and tractable uncertainty estimates of graph-based forecasting neural network predictions.  In this work, we have developed a unique approach to overcome the limitations of existing methods in estimating uncertainty for the MTSF task. Our approach effectively models the complex inter-series correlations among the variables and long-term(intra-series) dependencies within the variables in MTS data, resulting in improved forecast accuracy. Furthermore, our framework reduces computational requirements, making it suitable for handling larger datasets while achieving better forecast accuracy. The use of cost-effective GPU hardware accelerates the training process and reduces memory requirements, making the framework more practical for real-world use. 

\subsection{COMPOSITIONAL GENERALIZATION FOR GRAPH TIME SERIES FORECASTING}
Understanding the compositional structure of multivariate time series data is crucial for accurately modeling complex systems across various sectors, including retail and e-commerce, logistics and tranportation, finance, and more. To achieve this, it is essential for developing domain-agnostic frameworks that can adapt to new dynamic complex systems, with multiple interconnected variables that evolve over time which can be influenced by various factors such as time, seasonality, and external events. In graph time series forecasting, compositional generalization refers to the ability of a neural forecasting architecture to accurately model the complex relationships and dependencies between multiple time series variables in complex systems. By capturing the non-linear relationships between different time series variables and modeling their underlying dynamics, compositional generalization enables accurate forecasting in complex systems and improves the accuracy of multi-horizon predictions. This is because the framework can adapt to new graph structures and make accurate predictions even when data is incomplete or missing. Our proposed framework is specifically tailored for graph time series forecasting using Knowledge-Based Compositional Generalization(KBCG), which combines domain-specific knowledge with implicit knowledge of the relational structure learned from MTS data. The framework is designed to handle multi-horizon forecasting tasks that may involve varying numbers of time series variables and have complex, compositional structures underlying the MTS data. KBCG is used to learn a basis set of multi-relational structures that capture the underlying compositional structure of the time series data. It starts by decomposing the time series data into its latent factors, which include different types of dependencies that occur at various observation scales. For example, the correlations among variables in the short-term view of the MTS data may differ from those in the long-term view. The dependencies are then modeled using a set of multi-relational structures that capture the complex spatio-temporal dependencies among the multiple variables and the underlying dynamics in the time series data. The latent representations generated by multi-relational structures representation learning methods are more flexible and expressive than other approaches. These representations are combined using a set of weights, which determine the contribution of each basis relational structure to the final latent representation that captures the underlying dynamics of the time series data with greater accuracy. \textcolor{black}{Overall, KBCG provides a way to learn the compositional structure of multivariate time series data (i.e., underlying relationships and dependencies between different time series variables within a system) through the complex multi-relational structures by combining both domain-specific knowledge and implicit knowledge of the relational structure learned from the time series data}. As a result, the accuracy of multi-horizon forecasts is improved. KBCG is a powerful tool for graph time series forecasting and can be applied in a wide range of applications, such as traffic prediction, financial forecasting, and weather forecasting and many more.
\end{document}

%% file: math_commands.tex
\usepackage{amsmath,amsfonts,bm}

\def\eqref#1{equation~\ref{#1}}

\def\1{\bm{1}}

\DeclareMathAlphabet{\mathsfit}{\encodingdefault}{\sfdefault}{m}{sl}
\SetMathAlphabet{\mathsfit}{bold}{\encodingdefault}{\sfdefault}{bx}{n}













%% file: ijcai23.bbl
\begin{thebibliography}{}

\bibitem[\protect\citeauthoryear{Ba \bgroup \em et al.\egroup
  }{2016}]{ba2016layer}
Jimmy~Lei Ba, Jamie~Ryan Kiros, and Geoffrey~E Hinton.
\newblock Layer normalization.
\newblock {\em arXiv preprint arXiv:1607.06450}, 2016.

\bibitem[\protect\citeauthoryear{Bai \bgroup \em et al.\egroup
  }{2018}]{BaiTCN2018}
Shaojie Bai, J.~Zico Kolter, and Vladlen Koltun.
\newblock An empirical evaluation of generic convolutional and recurrent
  networks for sequence modeling.
\newblock {\em arXiv:1803.01271}, 2018.

\bibitem[\protect\citeauthoryear{Bai \bgroup \em et al.\egroup
  }{2019}]{bai2019STG2Seq}
Lei Bai, Lina Yao, Salil~S. Kanhere, Xianzhi Wang, and Quan~Z. Sheng.
\newblock Stg2seq: Spatial-temporal graph to sequence model for multi-step
  passenger demand forecasting.
\newblock In {\em IJCAI}, 7 2019.

\bibitem[\protect\citeauthoryear{Bai \bgroup \em et al.\egroup
  }{2020a}]{NEURIPS2020_ce1aad92}
Lei Bai, Lina Yao, Can Li, Xianzhi Wang, and Can Wang.
\newblock Adaptive graph convolutional recurrent network for traffic
  forecasting.
\newblock In {\em NeurIPS}, volume~33, pages 17804--17815, 2020.

\bibitem[\protect\citeauthoryear{Bai \bgroup \em et al.\egroup
  }{2020b}]{Bai2020nips}
Lei Bai, Lina Yao, Can Li, Xianzhi Wang, and Can Wang.
\newblock Adaptive graph convolutional recurrent network for traffic
  forecasting.
\newblock In {\em NeurIPS}, 2020.

\bibitem[\protect\citeauthoryear{Berge}{1973}]{berge1973graphs}
Claude Berge.
\newblock Graphs and hypergraphs.
\newblock 1973.

\bibitem[\protect\citeauthoryear{Chen \bgroup \em et al.\egroup
  }{2001}]{chen2001freeway}
Chao Chen, Karl Petty, Alexander Skabardonis, Pravin Varaiya, and Zhanfeng Jia.
\newblock Freeway performance measurement system: mining loop detector data.
\newblock {\em Transportation Research Record}, 1748(1):96--102, 2001.

\bibitem[\protect\citeauthoryear{Chen \bgroup \em et al.\egroup
  }{2021}]{chen2021ZGCNET}
Yuzhou Chen, Ignacio Segovia-Dominguez, and Yulia~R Gel.
\newblock Z-gcnets: Time zigzags at graph convolutional networks for time
  series forecasting.
\newblock In {\em ICML}, 2021.

\bibitem[\protect\citeauthoryear{Chen \bgroup \em et al.\egroup
  }{2022}]{tampsgcnets2022}
Yuzhou Chen, Ignacio Segovia-Dominguez, Baris Coskunuzer, and Yulia Gel.
\newblock {TAMP}-s2{GCN}ets: Coupling time-aware multipersistence knowledge
  representation with spatio-supra graph convolutional networks for time-series
  forecasting.
\newblock In {\em International Conference on Learning Representations}, 2022.

\bibitem[\protect\citeauthoryear{Cho \bgroup \em et al.\egroup
  }{2014}]{cho2014grued}
Kyunghyun Cho, B~{van Merrienboer}, Caglar Gulcehre, F~Bougares, H~Schwenk, and
  Yoshua Bengio.
\newblock Learning phrase representations using rnn encoder-decoder for
  statistical machine translation.
\newblock In {\em EMNLP}, 2014.

\bibitem[\protect\citeauthoryear{Choi \bgroup \em et al.\egroup
  }{2022}]{choi2022graph}
Jeongwhan Choi, Hwangyong Choi, Jeehyun Hwang, and Noseong Park.
\newblock Graph neural controlled differential equations for traffic
  forecasting.
\newblock In {\em Proceedings of the AAAI Conference on Artificial
  Intelligence}, volume~36, pages 6367--6374, 2022.

\bibitem[\protect\citeauthoryear{Dauphin \bgroup \em et al.\egroup
  }{2017}]{dauphin2017language}
Yann~N Dauphin, Angela Fan, Michael Auli, and David Grangier.
\newblock Language modeling with gated convolutional networks.
\newblock In {\em International conference on machine learning}, pages
  933--941. PMLR, 2017.

\bibitem[\protect\citeauthoryear{Deng and Hooi}{2021}]{deng2021graph}
Ailin Deng and Bryan Hooi.
\newblock Graph neural network-based anomaly detection in multivariate time
  series.
\newblock In {\em Proceedings of the AAAI Conference on Artificial
  Intelligence}, volume~35, pages 4027--4035, 2021.

\bibitem[\protect\citeauthoryear{Fang \bgroup \em et al.\egroup
  }{2021}]{fang2021STODE}
Zheng Fang, Qingqing Long, Guojie Song, and Kunqing Xie.
\newblock Spatial-temporal graph ode networks for traffic flow forecasting.
\newblock In {\em KDD}, 2021.

\bibitem[\protect\citeauthoryear{Fey}{2019}]{fey2019just}
Matthias Fey.
\newblock Just jump: Dynamic neighborhood aggregation in graph neural networks.
\newblock {\em arXiv preprint arXiv:1904.04849}, 2019.

\bibitem[\protect\citeauthoryear{Gao and Ji}{2019}]{gao2019graph}
Hongyang Gao and Shuiwang Ji.
\newblock Graph u-nets.
\newblock In {\em international conference on machine learning}, pages
  2083--2092. PMLR, 2019.

\bibitem[\protect\citeauthoryear{Gao and Ribeiro}{2022}]{gao2022equivalence}
Jianfei Gao and Bruno Ribeiro.
\newblock On the equivalence between temporal and static equivariant graph
  representations.
\newblock In {\em International Conference on Machine Learning}, pages
  7052--7076. PMLR, 2022.

\bibitem[\protect\citeauthoryear{Gilmer \bgroup \em et al.\egroup
  }{2017}]{gilmer2017neural}
Justin Gilmer, Samuel~S Schoenholz, Patrick~F Riley, Oriol Vinyals, and
  George~E Dahl.
\newblock Neural message passing for quantum chemistry.
\newblock In {\em International conference on machine learning}, pages
  1263--1272. PMLR, 2017.

\bibitem[\protect\citeauthoryear{Guo \bgroup \em et al.\egroup
  }{2019}]{guo2019astgcn}
Shengnan Guo, Youfang Lin, Ning Feng, Chao Song, and Huaiyu Wan.
\newblock Attention based spatial-temporal graph convolutional networks for
  traffic flow forecasting.
\newblock In {\em AAAI}, Jul. 2019.

\bibitem[\protect\citeauthoryear{Guo \bgroup \em et al.\egroup
  }{2020}]{guo2020optimized}
Kan Guo, Yongli Hu, Zhen Qian, Hao Liu, Ke~Zhang, Yanfeng Sun, Junbin Gao, and
  Baocai Yin.
\newblock Optimized graph convolution recurrent neural network for traffic
  prediction.
\newblock {\em IEEE Transactions on Intelligent Transportation Systems},
  22(2):1138--1149, 2020.

\bibitem[\protect\citeauthoryear{Hamilton}{2020}]{hamilton2020time}
James~Douglas Hamilton.
\newblock {\em Time series analysis}.
\newblock Princeton university press, 2020.

\bibitem[\protect\citeauthoryear{He \bgroup \em et al.\egroup
  }{2016}]{he2016deep}
Kaiming He, Xiangyu Zhang, Shaoqing Ren, and Jian Sun.
\newblock Deep residual learning for image recognition.
\newblock In {\em CVPR}, 2016.

\bibitem[\protect\citeauthoryear{Huang \bgroup \em et al.\egroup
  }{2019}]{Huang2019DSANet}
Siteng Huang, Donglin Wang, Xuehan Wu, and Ao~Tang.
\newblock Dsanet: Dual self-attention network for multivariate time series
  forecasting.
\newblock In {\em CIKM}, November 2019.

\bibitem[\protect\citeauthoryear{Huang \bgroup \em et al.\egroup
  }{2020}]{huang2020lsgcn}
Rongzhou Huang, Chuyin Huang, Yubao Liu, Genan Dai, and Weiyang Kong.
\newblock Lsgcn: Long short-term traffic prediction with graph convolutional
  networks.
\newblock In {\em IJCAI}, pages 2355--2361, 2020.

\bibitem[\protect\citeauthoryear{Hyndman \bgroup \em et al.\egroup
  }{2008}]{hyndman2008forecasting}
Rob Hyndman, Anne~B Koehler, J~Keith Ord, and Ralph~D Snyder.
\newblock {\em Forecasting with exponential smoothing: the state space
  approach}.
\newblock Springer Science \& Business Media, 2008.

\bibitem[\protect\citeauthoryear{Ishiguro \bgroup \em et al.\egroup
  }{2019}]{ishiguro2019graph}
Katsuhiko Ishiguro, Shin-ichi Maeda, and Masanori Koyama.
\newblock Graph warp module: an auxiliary module for boosting the power of
  graph neural networks.
\newblock {\em arXiv preprint arXiv:1902.01020}, 2019.

\bibitem[\protect\citeauthoryear{Jang \bgroup \em et al.\egroup
  }{2016}]{jang2016categorical}
Eric Jang, Shixiang Gu, and Ben Poole.
\newblock Categorical reparameterization with gumbel-softmax.
\newblock {\em arXiv preprint arXiv:1611.01144}, 2016.

\bibitem[\protect\citeauthoryear{Jiang \bgroup \em et al.\egroup
  }{2021}]{jiang2021dl}
Renhe Jiang, Du~Yin, Zhaonan Wang, Yizhuo Wang, Jiewen Deng, Hangchen Liu,
  Zekun Cai, Jinliang Deng, Xuan Song, and Ryosuke Shibasaki.
\newblock Dl-traff: Survey and benchmark of deep learning models for urban
  traffic prediction.
\newblock In {\em Proceedings of the 30th ACM international conference on
  information \& knowledge management}, pages 4515--4525, 2021.

\bibitem[\protect\citeauthoryear{Kipf and Welling}{2016}]{kipf2016semi}
Thomas~N Kipf and Max Welling.
\newblock Semi-supervised classification with graph convolutional networks.
\newblock {\em arXiv preprint arXiv:1609.02907}, 2016.

\bibitem[\protect\citeauthoryear{Kipf \bgroup \em et al.\egroup
  }{2018}]{kipf2018neural}
Thomas Kipf, Ethan Fetaya, Kuan-Chieh Wang, Max Welling, and Richard Zemel.
\newblock Neural relational inference for interacting systems.
\newblock In {\em International Conference on Machine Learning}, pages
  2688--2697. PMLR, 2018.

\bibitem[\protect\citeauthoryear{Lee \bgroup \em et al.\egroup
  }{2019}]{lee2019self}
Junhyun Lee, Inyeop Lee, and Jaewoo Kang.
\newblock Self-attention graph pooling.
\newblock In {\em International conference on machine learning}, pages
  3734--3743. PMLR, 2019.

\bibitem[\protect\citeauthoryear{Li and Zhu}{2021}]{li2021stfgnn}
Mengzhang Li and Zhanxing Zhu.
\newblock Spatial-temporal fusion graph neural networks for traffic flow
  forecasting.
\newblock In {\em AAAI}, May 2021.

\bibitem[\protect\citeauthoryear{Li \bgroup \em et al.\egroup
  }{2017}]{li2017diffusion}
Yaguang Li, Rose Yu, Cyrus Shahabi, and Yan Liu.
\newblock Diffusion convolutional recurrent neural network: Data-driven traffic
  forecasting.
\newblock {\em arXiv preprint arXiv:1707.01926}, 2017.

\bibitem[\protect\citeauthoryear{Li \bgroup \em et al.\egroup
  }{2018a}]{LiYS018}
Yaguang Li, Rose Yu, Cyrus Shahabi, and Yan Liu.
\newblock Diffusion convolutional recurrent neural network: Data-driven traffic
  forecasting.
\newblock In {\em {ICLR} (Poster)}, 2018.

\bibitem[\protect\citeauthoryear{Li \bgroup \em et al.\egroup
  }{2018b}]{li2018dcrnn_traffic}
Yaguang Li, Rose Yu, Cyrus Shahabi, and Yan Liu.
\newblock Diffusion convolutional recurrent neural network: Data-driven traffic
  forecasting.
\newblock In {\em ICLR}, 2018.

\bibitem[\protect\citeauthoryear{Makridakis and
  Hibon}{1997}]{makridakis1997arma}
Spyros Makridakis and Michele Hibon.
\newblock Arma models and the box--jenkins methodology.
\newblock {\em Journal of forecasting}, 16(3):147--163, 1997.

\bibitem[\protect\citeauthoryear{Nix and Weigend}{1994}]{nix1994estimating}
David~A Nix and Andreas~S Weigend.
\newblock Estimating the mean and variance of the target probability
  distribution.
\newblock In {\em Proceedings of 1994 ieee international conference on neural
  networks (ICNN'94)}, volume~1, pages 55--60. IEEE, 1994.

\bibitem[\protect\citeauthoryear{Pham \bgroup \em et al.\egroup
  }{2017}]{pham2017graph}
Trang Pham, Truyen Tran, Hoa Dam, and Svetha Venkatesh.
\newblock Graph classification via deep learning with virtual nodes.
\newblock {\em arXiv preprint arXiv:1708.04357}, 2017.

\bibitem[\protect\citeauthoryear{Ramp{\'a}{\v{s}}ek and
  Wolf}{2021}]{rampavsek2021hierarchical}
Ladislav Ramp{\'a}{\v{s}}ek and Guy Wolf.
\newblock Hierarchical graph neural nets can capture long-range interactions.
\newblock In {\em 2021 IEEE 31st International Workshop on Machine Learning for
  Signal Processing (MLSP)}, pages 1--6. IEEE, 2021.

\bibitem[\protect\citeauthoryear{Scheinerman and
  Ullman}{2011}]{scheinerman2011fractional}
Edward~R Scheinerman and Daniel~H Ullman.
\newblock {\em Fractional graph theory: a rational approach to the theory of
  graphs}.
\newblock Courier Corporation, 2011.

\bibitem[\protect\citeauthoryear{Shang \bgroup \em et al.\egroup
  }{2021}]{shang2021discrete}
Chao Shang, Jie Chen, and Jinbo Bi.
\newblock Discrete graph structure learning for forecasting multiple time
  series.
\newblock {\em arXiv preprint arXiv:2101.06861}, 2021.

\bibitem[\protect\citeauthoryear{Shih \bgroup \em et al.\egroup
  }{2019}]{shih2019temporal}
Shun-Yao Shih, Fan-Keng Sun, and Hung-yi Lee.
\newblock Temporal pattern attention for multivariate time series forecasting.
\newblock {\em Machine Learning}, 108(8):1421--1441, 2019.

\bibitem[\protect\citeauthoryear{Song \bgroup \em et al.\egroup
  }{2020}]{song2020stsgcn}
Chao Song, Youfang Lin, Shengnan Guo, and Huaiyu Wan.
\newblock Spatial-temporal synchronous graph convolutional networks: A new
  framework for spatial-temporal network data forecasting.
\newblock In {\em AAAI}, Apr. 2020.

\bibitem[\protect\citeauthoryear{Sutskever \bgroup \em et al.\egroup
  }{2014}]{sutskever2014sequence}
Ilya Sutskever, Oriol Vinyals, and Quoc~V Le.
\newblock Sequence to sequence learning with neural networks.
\newblock In {\em NeurIPS}, pages 3104--3112, 2014.

\bibitem[\protect\citeauthoryear{Vaswani \bgroup \em et al.\egroup
  }{2017}]{vaswani2017attention}
Ashish Vaswani, Noam Shazeer, Niki Parmar, Jakob Uszkoreit, Llion Jones,
  Aidan~N Gomez, {\L}ukasz Kaiser, and Illia Polosukhin.
\newblock Attention is all you need.
\newblock {\em Advances in neural information processing systems}, 30, 2017.

\bibitem[\protect\citeauthoryear{Watson}{1994}]{watson1994vector}
Mark~W Watson.
\newblock Vector autoregressions and cointegration.
\newblock {\em Handbook of econometrics}, 4:2843--2915, 1994.

\bibitem[\protect\citeauthoryear{Wu \bgroup \em et al.\egroup
  }{2019a}]{wu2019graphwavenet}
Zonghan Wu, Shirui Pan, Guodong Long, Jing Jiang, and Chengqi Zhang.
\newblock Graph wavenet for deep spatial-temporal graph modeling.
\newblock In {\em IJCAI}, pages 1907--1913, 7 2019.

\bibitem[\protect\citeauthoryear{Wu \bgroup \em et al.\egroup
  }{2019b}]{zonghanwu2019}
Zonghan Wu, Shirui Pan, Guodong Long, Jing Jiang, and Chengqi Zhang.
\newblock Graph wavenet for deep spatial-temporal graph modeling.
\newblock In {\em {IJCAI}}, pages 1907--1913, 2019.

\bibitem[\protect\citeauthoryear{Wu \bgroup \em et al.\egroup
  }{2020}]{wu2020connecting}
Zonghan Wu, Shirui Pan, Guodong Long, Jing Jiang, Xiaojun Chang, and Chengqi
  Zhang.
\newblock Connecting the dots: Multivariate time series forecasting with graph
  neural networks.
\newblock In {\em Proceedings of the 26th ACM SIGKDD international conference
  on knowledge discovery \& data mining}, pages 753--763, 2020.

\bibitem[\protect\citeauthoryear{Xu \bgroup \em et al.\egroup
  }{2018}]{xu2018representation}
Keyulu Xu, Chengtao Li, Yonglong Tian, Tomohiro Sonobe, Ken-ichi Kawarabayashi,
  and Stefanie Jegelka.
\newblock Representation learning on graphs with jumping knowledge networks.
\newblock In {\em International conference on machine learning}, pages
  5453--5462. PMLR, 2018.

\bibitem[\protect\citeauthoryear{Yu \bgroup \em et al.\egroup
  }{2017}]{yu2017spatio}
Bing Yu, Haoteng Yin, and Zhanxing Zhu.
\newblock Spatio-temporal graph convolutional networks: A deep learning
  framework for traffic forecasting.
\newblock {\em arXiv preprint arXiv:1709.04875}, 2017.

\bibitem[\protect\citeauthoryear{Yu \bgroup \em et al.\egroup
  }{2018a}]{bing2018stgcn}
Bing Yu, Haoteng Yin, and Zhanxing Zhu.
\newblock Spatio-temporal graph convolutional networks: A deep learning
  framework for traffic forecasting.
\newblock In {\em IJCAI}, 7 2018.

\bibitem[\protect\citeauthoryear{Yu \bgroup \em et al.\egroup }{2018b}]{YuYZ18}
Bing Yu, Haoteng Yin, and Zhanxing Zhu.
\newblock Spatio-temporal graph convolutional networks: {A} deep learning
  framework for traffic forecasting.
\newblock In {\em {IJCAI}}, pages 3634--3640, 2018.

\end{thebibliography}
